\documentclass[conference]{IEEEtran}
\IEEEoverridecommandlockouts
\usepackage{cite}
\usepackage{amsmath,amssymb,amsfonts}
\usepackage{algorithmic}
\usepackage{graphicx}
\usepackage{textcomp}
\usepackage{xcolor}
\def\BibTeX{{\rm B\kern-.05em{\sc i\kern-.025em b}\kern-.08em
    T\kern-.1667em\lower.7ex\hbox{E}\kern-.125emX}}
\newtheorem{theorem}{Theorem}
\newtheorem{corollary}{Corollary}
\usepackage{graphicx} 

\begin{document}

\title{Stochastic Variational Inference for Bayesian Sparse Gaussian Process Regression\thanks{This research is supported by the National Research Foundation, Prime Minister’s Office, Singapore under its Campus for Research Excellence and Technological Enterprise (CREATE) programme and the Singapore Ministry of Education Academic Research Fund Tier $2$, MOE$2016$-T$2$-$2$-$156$.}
}

\author{\IEEEauthorblockN{Haibin Yu} 
\IEEEauthorblockA{
\textit{National University of Singapore}\\
Republic of Singapore \\
haibin@u.nus.edu}
\and
\IEEEauthorblockN{Trong Nghia Hoang}
\IEEEauthorblockA{
\textit{MIT-IBM Watson AI Lab}\\
Cambridge, MA, USA\\
nghiaht@ibm.com}
\and
\IEEEauthorblockN{Bryan Kian Hsiang Low}
\IEEEauthorblockA{
\textit{National University of Singapore}\\
Republic of Singapore \\
lowkh@comp.nus.edu.sg}
\and
\IEEEauthorblockN{Patrick Jaillet}
\IEEEauthorblockA{
\textit{MIT}\\
Cambridge, MA, USA\\
jaillet@mit.edu}
}

\maketitle

\begin{abstract}
This paper presents a novel  variational inference framework for deriving a family of  Bayesian \emph{sparse Gaussian process regression} (SGPR) models whose approximations are variationally optimal with respect to the full-rank GPR model enriched with various corresponding correlation structures of the observation noises.
Our \emph{variational Bayesian SGPR} (VBSGPR) models jointly treat both the distributions of the inducing variables and hyperparameters as variational parameters, which enables the decomposability of the variational lower bound that in turn can be exploited for stochastic optimization.  
Such a stochastic optimization involves iteratively following the stochastic gradient of the variational lower bound to improve its estimates of the optimal variational distributions of the inducing variables and hyperparameters (and hence the predictive distribution) of our VBSGPR models and is guaranteed to achieve asymptotic convergence to them.
We show that the  stochastic gradient is an unbiased estimator of the exact gradient and can be computed in constant time per iteration, hence achieving scalability to big data.
We empirically evaluate the performance of our proposed framework on two real-world, massive datasets.
\end{abstract}


\section{Introduction}
\label{sect:intro}
A \emph{Gaussian process regression} (GPR) model is a rich class of Bayesian non-parametric models that can exploit correlation of the data/observations for performing probabilistic non-linear regression by providing a Gaussian predictive distribution with formal measures of predictive uncertainty.
	Such a \emph{full-rank GPR} (FGPR) model, though highly expressive, incurs cubic time in the data size to compute the predictive distribution and 
	learn the hyperparameters (i.e., defining its correlation structure) via maximum likelihood estimation, 
specifically, in each iteration of gradient ascent to refine the  hyperparameter estimates to improve the log-marginal likelihood.
	So, to learn the hyperparameters in reasonable time, only a very small subset of the data can be considered, which compromises the estimation accuracy:
	It is typically not representative of all the data in describing the underlying correlation structure due to its sparsity over the input space. 
	
	To improve its time efficiency, a number of \emph{sparse GPR} (SGPR) models exploiting low-rank covariance matrix approximations \cite{candela10,candela05} have been proposed,
	many of which impose a common structural assumption of conditional independence (but of varying degrees) on the FGPR model based on the notion of \emph{inducing variables} and can therefore be encompassed under a unifying view presented in~\cite{candela05}.
	As a result, they incur linear time in the data size that is still prohibitively expensive for training with big data (i.e., million-sized datasets).
	To scale up to big data, parallel \cite{LowUAI13,LowAAAI15,LowDyDESS15} and online \cite{Csato02,LowAAAI14} variants of several of these SGPR models have been developed for prediction (by assuming known hyperparameters) but not hyperparameter learning.
	
	The chief concern with the unifying view of~\cite{candela05} is that it does not rigorously quantify the approximation quality of a SGPR model \cite{Titsias09a}.
	To address this concern, the work of~\cite{Titsias09} has proposed a principled variational inference framework that involves minimizing the \emph{Kullback-Leibler} (KL) distance between distributions of some latent variables (including the inducing variables) induced by the variational SGPR approximation and
 the FGPR model given the data/observations or, equivalently, maximizing a lower bound of the log-marginal likelihood to yield the \emph{deterministic training conditional} (DTC) approximation \cite{Seeger03}. Hyperparameter learning is then achieved by maximizing this variational lower bound with respect to the hyperparameters via gradient ascent, which still incurs linear time in the data size per iteration but can be substantially reduced by means of parallelization \cite{Yarin14} or stochastic optimization \cite{Lawrence13,cheng2016incremental}. 
%
	Unifying frameworks of variational SGPR models and their stochastic and distributed variants are subsequently proposed in \cite{NghiaICML15,HoangICML16} to, respectively, perform stochastic and distributed variational inference 
for any SGPR model (including DTC) spanned by the unifying view of~\cite{candela05}.
The work of~\cite{bui2017streaming} has extended two SGPR models (i.e., DTC and \emph{fully independent training conditional} (FITC) approximation \cite{Snelson06}) to handle streaming data.

	However, all the above-mentioned variational SGPR models and their stochastic and distributed variants suffer from the following critical issues: 
	(a) The above equivalence only holds for the case of fixed hyperparameters;
	otherwise, since the log-marginal likelihood 
	also depends on the same hyperparameters that are optimized to maximize its variational lower bound, the resulting KL distance, which quantifies the gap between the log-marginal likelihood and its lower bound, may not be minimized;
	(b) similar to variational expectation-maximization \cite{Jordan08}, the log-marginal likelihood does not necessarily increase in each iteration of gradient ascent to refine the hyperparameter estimates to improve its variational lower bound; and
	(c) they all find point estimates of the hyperparameters, which risk overfitting, especially when the number of hyperparameters is all but small.
	
	To resolve these issues, the notable work of~\cite{Titsias13} has introduced a \emph{variational Bayesian DTC} (VBDTC) approximation (Section~\ref{Variational Inference of the Bayesian DTC}) capable of learning a variational distribution of the hyperparameters.
	This learned distribution of hyperparameters is particularly desirable in conveying the uncertainty/confidence of the hyperparameter estimates and for use in Bayesian GP regression (Section~\ref{predict}), active learning \cite{LowAAMAS13,NghiaICML14,LowAAMAS08,LowICAPS09,LowAAMAS11,LowAAMAS14,YehongAAAI16}, Bayesian optimization~\cite{Erik17,Ghahramani14,NghiaAAAI18,ling16}, among others.
Unfortunately, such a VBDTC approximation cannot handle big data (e.g., million-sized datasets) because it incurs linear time in the data size per iteration of gradient ascent. The  \emph{variational Bayesian sparse spectrum GPR} (VSSGPR) model~\cite{Gal2015Improving} overcomes this scalability issue by achieving constant time per iteration of stochastic gradient ascent. But, like VBDTC, VSSGPR imposes a highly restrictive assumption of conditional independence between the test outputs and the training data given the learned hyperparameters (i.e., in its test conditional in equation $4$ therein), thus compromising its predictive performance as shown in our experiments (Section~\ref{Experiments and Discussion}). 
This assumption is later relaxed in the work of~\cite{MinhAAAI17}.
It remains an open question whether more refined SGPR models as well as those others spanned by the unifying view of~\cite{candela05} (e.g., FITC, \emph{partially independent training conditional} (PITC), \emph{partially independent conditional} (PIC) \cite{Snelson07a} approximations) are amenable to the variational Bayesian treatment and achieve scalability through stochastic optimization.
		
To address this question, this paper presents a novel variational inference framework for deriving a family of Bayesian SGPR models (e.g., VBDTC, VBFITC, VBPIC) whose approximations are, interestingly, variationally optimal with respect to the FGPR model enriched with various corresponding correlation structures of the observation noises (Section~\ref{Variational Inference of the Bayesian DTC}).
Our framework introduces a novel reparameterization of the GP model (Section~\ref{Seperation of Gaussian Process}) for enabling a variational treatment of the distribution of hyperparameters.
	Unlike VBDTC, our framework does not need to assume independently distributed observation noises with constant variance and is thus more robust to different noise correlation structures, hence catering to more realistic applications of GP.
	Furthermore, instead of just considering the distribution of hyperparameters as variational parameters \cite{Titsias13,Gal2015Improving}, we jointly treat both the distributions of the inducing variables and hyperparameters as variational parameters, which  enables the decomposability of the variational lower bound that in turn can be exploited for stochastic optimization (Section~\ref{Stochastic Variational Inference for GPR}).  
	Such a stochastic optimization involves iteratively following the stochastic gradient of the variational lower bound to improve its estimates of the optimal variational distributions of the inducing variables and hyperparameters (and hence the predictive distribution (Section~\ref{predict})) of our \emph{variational Bayesian SGPR} (VBSGPR) models and is guaranteed to achieve asymptotic convergence to them.
	We show that the derived stochastic gradient is an unbiased estimator of the exact gradient and can be computed in constant time (i.e., independent of data size) per iteration, thus achieving scalability to big data.
	We empirically evaluate the performance of the stochastic variants of our VBSGPR models on two real-world  datasets (Section~\ref{Experiments and Discussion}).
%
%
\section{Background and Notations}
\subsection{Full-Rank GP Regression (FGPR) with Correlated  Noises}
\label{full}
Let $\mathcal{X}$ denote a $d$-dimensional input feature space such that each input vector $\mathbf{x}\in\mathcal{X}$ is associated with a latent output variable $f_{\mathbf{x}}$.
		Let $\{f_{\mathbf{x}}\}_{\mathbf{x}\in\mathcal{X}}$ denote a \emph{Gaussian process} (GP), that is, every finite subset of $\{f_{\mathbf{x}}\}_{\mathbf{x}\in\mathcal{X}}$ follows a multivariate Gaussian distribution. Then, the GP is fully specified by its \emph{prior} mean $\mathbb{E}[f_\mathbf{x}]$  (i.e., assumed to be zero to ease notations) and covariance $k_{\mathbf{x}\mathbf{x}'}\triangleq\mathrm{cov}[f_\mathbf{x}, f_{\mathbf{x}'}]$ for all $\mathbf{x}, \mathbf{x}' \in \mathcal{X}$, the latter of which can be defined, for example, by the widely-used squared exponential covariance function $k_{\mathbf{x}\mathbf{x}'} \triangleq 
\sigma_f^2\exp(-0.5\|\mathbf{\Lambda}\mathbf{x} - \mathbf{\Lambda}\mathbf{x}'\|^2_2)$ where 
$\mathbf{\Lambda} = \mathrm{diag}[\lambda_1,\ldots,\lambda_d]$ and $\sigma_f^2$ are its \emph{inverted} length-scale and signal variance hyperparameters, respectively. 
Suppose that a column vector $\mathbf{y}_\mathcal{D}\triangleq (y_{\mathbf{x}})^\top_{\mathbf{x}\in\mathcal{D}}$ of noisy observed outputs $y_{\mathbf{x}}\triangleq f_{\mathbf{x}}+\varepsilon_{\mathbf{x}}$ (i.e., corrupted by an additive noise $\varepsilon_{\mathbf{x}}$) is available for some set $\mathcal{D}\subset\mathcal{X}$ of training inputs such that $\boldsymbol{\varepsilon}_{\mathcal{D}}\triangleq(\varepsilon_{\mathbf{x}})^\top_{\mathbf{x} \in \mathcal{D}}$ follows a multivariate Gaussian distribution $p(\boldsymbol{\varepsilon}_{\mathcal{D}}) \triangleq \mathcal{N}(\mathbf{0},\mathbf{C}_{\mathcal{DD}})$ where $\mathbf{C}_{\mathcal{DD}}$   
 is a covariance matrix representing the correlation of observation noises $\boldsymbol{\varepsilon}_\mathcal{D}$. 	It follows that $p(\mathbf{y}_\mathcal{D}|\mathbf{f}_\mathcal{D})=\mathcal{N}(\mathbf{f}_\mathcal{D},\mathbf{C}_{\mathcal{DD}})$ where $\mathbf{f}_\mathcal{D}\triangleq (f_{\mathbf{x}})^\top_{\mathbf{x}\in\mathcal{D}}$.
Then, a FGPR model with correlated observation noises 
can perform probabilistic regression by providing a GP \emph{posterior}/predictive distribution $p(f_{\mathbf{x}^*} | \mathbf{y}_\mathcal{D}) = \mathcal{N}(\mathbf{K}_{\mathbf{x}^*\mathcal{D}}(\mathbf{K}_{\mathcal{D}\mathcal{D}} + \mathbf{C}_{\mathcal{D}\mathcal{D}})^{-1}\mathbf{y}_\mathcal{D}, k_{\mathbf{x}^*\mathbf{x}^*} - \mathbf{K}_{\mathbf{x}^*\mathcal{D}}(\mathbf{K}_{\mathcal{D}\mathcal{D}} + \mathbf{C}_{\mathcal{D}\mathcal{D}})^{-1}\mathbf{K}_{\mathcal{D}\mathbf{x}^*})$ of the latent output $f_{\mathbf{x}^*}$ for any test input $\mathbf{x}^*\in\mathcal{X}$ where $\mathbf{K}_{\mathbf{x}^*\mathcal{D}} \triangleq (k_{\mathbf{x}^*\mathbf{x}})_{\mathbf{x}\in \mathcal{D}}$, $\mathbf{K}_\mathcal{DD} \triangleq (k_{\mathbf{x}\mathbf{x}'})_{\mathbf{x},\mathbf{x}' \in \mathcal{D}}$, and $\mathbf{K}_{\mathcal{D}\mathbf{x}^*} \triangleq \mathbf{K}^\top_{\mathbf{x}^*\mathcal{D}}$.
Computing the GP predictive distribution incurs $\mathcal{O}(|\mathcal{D}|^3)$ time due to inversion of $\mathbf{K}_\mathcal{DD} + \mathbf{C}_{\mathcal{D}\mathcal{D}}$. 
The FGPR hyperparameters \textcolor{black}{$\boldsymbol{\theta}\triangleq (\lambda_1,\ldots,\lambda_d, \sigma_f)^\top$} 
can be learned using \emph{maximum likelihood estimation} (MLE) by maximizing the log-marginal likelihood $\log p(\mathbf{y}_\mathcal{D}) = \log \mathcal{N}(\mathbf{0}, \mathbf{K}_\mathcal{DD}\hspace{-0mm} +\hspace{-0mm}  \mathbf{C}_{\mathcal{D}\mathcal{D}})$ with respect to \textcolor{black}{$\boldsymbol{\theta}$} 
via gradient ascent, which incurs $\mathcal{O}(|\mathcal{D}|^3)$ time per iteration. 	
So, the FGPR model with correlated noises scales poorly in data size $|\mathcal{D}|$. 
To improve its scalability, our key idea is to impose different sparsity structures on $\mathbf{C}_{\mathcal{D}\mathcal{D}}$ to yield a family of VBSGPR models, as shown  in Section~\ref{Variational Inference of the Bayesian DTC}.
	\subsection{Sparse Gaussian Process Regression (SGPR)}
	\label{sgpr}
	To reduce the cubic time cost of the FGPR model, 		
	the SGPR models spanned by the unifying view of~\cite{candela05} exploit a vector $\mathbf{f}_\mathcal{U}\triangleq (f_{\mathbf{x}})_{\mathbf{x}\in\mathcal{U}}^\top$ of inducing output variables for some small set $\mathcal{U}\subset\mathcal{X}$ of inducing inputs (i.e., $|\mathcal{U}|\ll|\mathcal{D}|$) for approximating the GP predictive distribution $p(f_{\mathbf{x}^*}|\mathbf{y}_\mathcal{D})$. In particular, they utilize a common structural assumption \cite{Snelson07a} that the joint distribution of $f_{\mathbf{x}^*}$ and $\mathbf{f}_\mathcal{D} \triangleq (f_{\mathbf{x}})_{\mathbf{x}\in\mathcal{D}}^\top$ given $\mathbf{f}_\mathcal{U}$ factorizes over a pre-defined partition of the input space $\mathcal{X}$ into $B$ disjoint subsets $\mathcal{X}_1,\ldots,\mathcal{X}_{B}$ (i.e., $\mathcal{X} = \mathcal{X}_1 \cup \mathcal{X}_2 \cup \ldots \cup \mathcal{X}_{B}$): 
Formally, without loss of generality, supposing $\mathbf{x}^* \in \mathcal{X}_B$, 
then $p(f_{\mathbf{x}^*},\mathbf{f}_\mathcal{D}|\mathbf{f}_\mathcal{U}) = p(f_{\mathbf{x}^*}|\mathbf{f}_{\mathcal{D}_B},\mathbf{f}_\mathcal{U}) \prod_{i=1}^{B}p(\mathbf{f}_{\mathcal{D}_i}|\mathbf{f}_\mathcal{U})$
	where $\mathbf{f}_{\mathcal{D}_i} \triangleq (f_{\mathbf{x}})_{\mathbf{x}\in\mathcal{D}_i}^\top$ is a column vector of latent outputs for the disjoint subset $\mathcal{D}_i \triangleq (\mathcal{X}_i \cap \mathcal{D}) \subset \mathcal{D}$ for $i = 1, 2, \ldots, B$. Using this factorization,   
$p(f_{\mathbf{x}^*}|\mathbf{y}_\mathcal{D}) = \int p(f_{\mathbf{x}^*}|\mathbf{y}_{\mathcal{D}_B},\mathbf{f}_\mathcal{U})\ p(\mathbf{f}_\mathcal{U}|\mathbf{y}_\mathcal{D})\ \mathrm{d}\mathbf{f}_\mathcal{U} \simeq \int q(f_{\mathbf{x}^*}|\mathbf{y}_{\mathcal{D}_B},\mathbf{f}_\mathcal{U})\ q(\mathbf{f}_\mathcal{U})\ \mathrm{d}\mathbf{f}_\mathcal{U}$	
	where $\mathbf{y}_{\mathcal{D}_B} \triangleq (y_{\mathbf{x}})^\top_{\mathbf{x}\in\mathcal{D}_B}$ is a vector of noisy observed outputs for the subset $\mathcal{D}_B$ of training inputs, the equality is derived in Appendix C.$1$ of \cite{NghiaICML15}, and
	$p(f_{\mathbf{x}^*}|\mathbf{y}_{\mathcal{D}_B},\mathbf{f}_\mathcal{U})$ and $p(\mathbf{f}_\mathcal{U}|\mathbf{y}_\mathcal{D})$ are, respectively, approximated by $q(f_{\mathbf{x}^*}|\mathbf{y}_{\mathcal{D}_B},\mathbf{f}_\mathcal{U})$ and $q(\mathbf{f}_\mathcal{U})$ 
that can be appropriately defined to reproduce the predictive distribution of any SGPR model \cite{NghiaICML15} spanned by the unifying view of \cite{candela05}, which can be computed in $\mathcal{O}(|\mathcal{D}||\mathcal{U}|^2)$ time.   
	The SGPR hyperparameters can be learned using MLE by maximizing its corresponding log-marginal likelihood via gradient ascent, which incurs $\mathcal{O}(|\mathcal{D}||\mathcal{U}|^2)$ time per iteration.
	To scale up to big data, these linear time complexities can be significantly reduced  using parallelization or stochastic optimization (Section~\ref{sect:intro}).
	\subsection{Bayesian SGPR Models}	
	\label{Bayesian SGP}\vspace{1mm}
	For the FGPR and SGPR models described above, 
	point estimates of their hyperparameters are learned, which is vulnerable to overfitting, especially when the number of hyperparameters is all but small (Section~\ref{sect:intro}). 
	To mitigate this issue of overfitting, a Bayesian approach to sparse GP regression can be employed by introducing priors $p(\boldsymbol{\theta}) \triangleq p(\mathbf{\Lambda})\ p(\sigma_f)$ over  hyperparameters $\boldsymbol{\theta}$,	
thus yielding the predictive distribution:\vspace{-0.3mm}
	\begin{equation}
		\hspace{0mm}
		\begin{array}{rcl}
			p(f_{\mathbf{x}^*}|\mathbf{y}_\mathcal{D}) &\hspace{-2.4mm}=&\hspace{-2.4mm} \displaystyle\int\hspace{-0mm} p(f_{\mathbf{x}^*}|\mathbf{y}_{\mathcal{D}_B},\mathbf{f}_\mathcal{U},\boldsymbol{\theta})\   p(\mathbf{f}_\mathcal{U},\boldsymbol{\theta}|\mathbf{y}_\mathcal{D}) \ \mathrm{d}\mathbf{f}_\mathcal{U}\ \mathrm{d}\boldsymbol{\theta} \\
			&\hspace{-2.4mm}\simeq&\hspace{-2.4mm}\displaystyle\int\hspace{-0mm} q(f_{\mathbf{x}^*}|\mathbf{y}_{\mathcal{D}_B},\mathbf{f}_\mathcal{U},\boldsymbol{\theta})\ q(\mathbf{f}_\mathcal{U},\boldsymbol{\theta})\ \mathrm{d}\mathbf{f}_\mathcal{U}\ \mathrm{d}\boldsymbol{\theta} \vspace{-5.5mm}
		\end{array}
		\label{E7}\vspace{3mm}
	\end{equation}
	where 
	$p(\mathbf{f}_\mathcal{U}, \boldsymbol{\theta} | \mathbf{y}_\mathcal{D})$ is approximated by $q(\mathbf{f}_\mathcal{U},\boldsymbol{\theta})$ which generalizes $q(\mathbf{f}_\mathcal{U})$ above
by additionally and jointly considering the hyperparameters $\boldsymbol{\theta}$ as variational variables.
	Though~\eqref{E7}, in principle, allows a Bayesian treatment of $\boldsymbol{\theta}$ to be incorporated into the existing SGPR models, computing the resulting predictive distribution is intractable because
	it involves integrating, over $\boldsymbol{\Lambda}$,
	probability terms in~\eqref{E7} containing the inverse of $\mathbf{K}_\mathcal{UU} \triangleq (k_{\mathbf{x}\mathbf{x}'})_{\mathbf{x},\mathbf{x}'\in\mathcal{U}}$ that depends on $\boldsymbol{\Lambda}$ but without an analytical form with respect to $\boldsymbol{\Lambda}$. 
%
	To resolve this, we introduce a reparameterization trick	
to make the prior distribution of inducing outputs independent of the hyperparameters $\boldsymbol{\theta}$, as discussed next. 

	%
\section{Reparameterizing Bayesian SGPR Models}
\label{Seperation of Gaussian Process}
Let $\phi: \mathbb{R}^d \rightarrow \mathcal{H}$ denote a non-linear feature map from the input space $\mathbb{R}^d$ into a \emph{reproducing kernel Hilbert space} (RKHS) $\mathcal{H}$ whose inner product is defined as $\langle\phi(\mathbf{x}),\phi(\mathbf{x}')\rangle_\mathcal{H} \triangleq \exp(-0.5\|\mathbf{x} - \mathbf{x}'\|_2^2)$. Given $\phi$, the GP covariance/kernel function can be interpreted as $k_{\mathbf{xx'}} \triangleq \left\langle\sigma(\mathbf{x})\phi(\boldsymbol{\Lambda}\mathbf{x}),\sigma(\mathbf{x}')\phi(\mathbf{\Lambda}\mathbf{x}')\right\rangle_\mathcal{H} = \sigma(\mathbf{x})\sigma(\mathbf{x}')\exp(-0.5\|\mathbf{\Lambda x} - \mathbf{\Lambda x}'\|_2^2)$ where $\sigma$ is an arbitrary function mapping from $\mathbb{R}^d$ to $\mathbb{R}$. This implies $k_\mathbf{xx} = \sigma^2(\mathbf{x})$ which allows $\sigma^2(\mathbf{x})$ to be interpreted as the prior variance $k_\mathbf{xx}$ of $f_\mathbf{x}$ (Section~\ref{full}). 

We will now describe the reparameterization trick: 
Let $\mathcal{I}\triangleq\{\mathbf{\Lambda}\mathbf{x}\}_{\mathbf{x} \in \mathcal{U}}$.
Intuitively, $\mathcal{I}$ can be interpreted as a set of \emph{rotated inducing inputs} with the diagonal matrix $\mathbf{\Lambda}$ of inverted length-scales being the rotation matrix. 
%
%
Let each rotated inducing input $\mathbf{z} \in \mathcal{I}$ be associated with a latent output variable $s_{\mathbf{z}}$. 
Then, for all $\mathbf{z},\mathbf{z}' \in \mathcal{I}$,  
$\text{cov}[s_{\mathbf{z}}, s_{\mathbf{z}'}] \triangleq \langle \sigma(\mathbf{z})\phi(\mathbf{z}), \sigma(\mathbf{z}^\prime)\phi(\mathbf{z}^\prime) \rangle_{\mathcal{H}} = \sigma(\mathbf{z})\sigma(\mathbf{z}')\exp(-0.5\|\mathbf{z} - \mathbf{z}'\|_2^2)$, by definition of RKHS. 
By assuming that the prior variances of $s_\mathbf{z}$ for all $\mathbf{z}\in\mathcal{I}$ are identical 
and equal to some constant $\zeta^2$ (i.e., $\sigma(\mathbf{z}) = \zeta > 0$), \textcolor{black}{$\text{cov}[s_{\mathbf{z}}, s_{\mathbf{z}'}] = \zeta^2\exp(-0.5\|\mathbf{z} - \mathbf{z}'\|_2^2)$} which is independent of $\boldsymbol{\theta}$. Consequently, the prior covariance matrix $\mathbf{\Sigma}_\mathcal{II} \triangleq (\text{cov}[s_{\mathbf{z}}, s_{\mathbf{z}'}])_{\mathbf{z},\mathbf{z}' \in \mathcal{I}}$ of the inducing output variables $\mathbf{s}_\mathcal{I} \triangleq (s_{\mathbf{z}})^\top_{\mathbf{z}\in\mathcal{I}}$ is independent of $\boldsymbol{\theta}$. 
%
%
Furthermore, the cross-covariance matrix $\mathbf{K}_{\mathcal{D}\mathcal{I}} \triangleq (\text{cov}[f_{\mathbf{x}}, s_{\mathbf{z}}])_{\mathbf{x} \in \mathcal{D},\mathbf{z} \in \mathcal{I}}$ between the latent outputs $\mathbf{f}_\mathcal{D}$ for some set $\mathcal{D}$ of training inputs and the inducing outputs $\mathbf{s}_\mathcal{I}$ can be computed analytically using 
the definition of RKHS: $\text{cov}[f_{\mathbf{x}}, s_{\mathbf{z}}] = \langle \sigma(\mathbf{x})\phi(\mathbf{\Lambda}\mathbf{x}), \sigma(\mathbf{z})\phi(\mathbf{z})\rangle_{\mathcal{H}} = \zeta\sigma(\mathbf{x})\exp(-0.5\|\mathbf{\Lambda}\mathbf{x}-\mathbf{z}\|^2_2)$. 
Like many existing GP models, the prior variances of 
$f_\mathbf{x}$ for all $\mathbf{x}\in\mathcal{X}$
are assumed to be identical and equal to a signal variance hyperparameter $\sigma_f^2$ (i.e., $\sigma(\mathbf{x}) = \sigma_f$) for tractable learning, hence circumventing the need to learn 
an infinite number of prior variance hyperparameters.
The resulting representation of the GP model from the reparameterization trick will allow the optimal variational distributions of inducing outputs $\mathbf{s}_\mathcal{I}$ and hyperparameters $\boldsymbol{\theta}$ (hence the predictive distribution) to be tractably derived for a family of VBSGPR models, as discussed in Section~\ref{Variational Inference of the Bayesian DTC}.\vspace{1mm}

\emph{Remark 1:} The definition of $\mathcal{I}$ seems to suggest its construction by first selecting the inducing inputs $\mathcal{U}$ and then rotating them via $\mathbf{\Lambda}$, which is not possible since $\mathbf{\Lambda}$ is not known \emph{a priori}.
However, as shall be discussed in Section~\ref{Variational Inference of the Bayesian DTC},
it is possible to first select $\mathcal{I}$ and then optimize the variational distribution of $\mathbf{\Lambda}$, which has an effect of optimizing the distribution of inducing inputs $\mathcal{U}$ in original input space $\mathcal{X}$.\vspace{1mm}



\emph{Remark 2:} Let $\mathcal{Z}\triangleq\{\mathbf{\Lambda}\mathbf{x}\}_{\mathbf{x} \in \mathcal{X}}$.
By setting the (identical) prior variances of $s_\mathbf{z}$ for all $\mathbf{z}\in\mathcal{Z}$ to unity
(i.e., $\zeta = 1$), $\{s_\mathbf{z}\}_{\mathbf{z}\in\mathcal{Z}}$ denote a \emph{standard} GP with unit signal variance and length-scales~\cite{Titsias13},
which is a special case of our representation of the GP model here.
Then, $f_\mathbf{x} = \sigma_f s_{\mathbf{\Lambda x}}$ for all $\mathbf{x} \in \mathcal{X}$.
%
%
%
\section{Variational Bayesian SGPR Models}
\label{Variational Inference of the Bayesian DTC}
Using our representation
of the GP model defined above (Section~\ref{Seperation of Gaussian Process}), the predictive distribution~\eqref{E7} of a Bayesian SGPR model can be slightly modified to $p(f_{\mathbf{x}^*}|\mathbf{y}_\mathcal{D})=\int p(f_{\mathbf{x}^*}|\mathbf{y}_{\mathcal{D}_B},\mathbf{s}_\mathcal{I},\boldsymbol{\theta})\ p(\mathbf{s}_\mathcal{I},\boldsymbol{\theta}|\mathbf{y}_\mathcal{D})\ \mathrm{d}\mathbf{s}_\mathcal{I}\ \mathrm{d}\boldsymbol{\theta}$ such that deriving the posterior $p(\mathbf{s}_\mathcal{I},\boldsymbol{\theta}|\mathbf{y}_\mathcal{D})=p(\mathbf{y}_\mathcal{D},\mathbf{s}_\mathcal{I},\boldsymbol{\theta})/p(\mathbf{y}_\mathcal{D})$ requires computing the likelihood:\vspace{-0.6mm}
\begin{equation}	
\hspace{0mm}
\begin{array}{c}
p(\mathbf{y}_\mathcal{D}) \hspace{-0mm}=\hspace{-0mm} \displaystyle\mathbb{E}_{\boldsymbol{\theta}}\hspace{-0mm}\left[\int \hspace{-0mm}p(\mathbf{y}_\mathcal{D}|\mathbf{f}_\mathcal{D}) p(\mathbf{f}_\mathcal{D}|\mathbf{s}_\mathcal{I},\boldsymbol{\theta}) p(\mathbf{s}_\mathcal{I})\mathrm{d}\mathbf{f}_\mathcal{D}\mathrm{d}\mathbf{s}_\mathcal{I}\right]\vspace{-0mm}
\end{array}
\label{E13}\vspace{-0.6mm}
\end{equation} 	
\textcolor{black}{where $p(\boldsymbol{\theta}) \triangleq \mathcal{N}(\mathbf{1}, \text{diag}[0.1])$, $p(\mathbf{s}_\mathcal{I})=\mathcal{N}(\mathbf{0},\mathbf{\Sigma}_\mathcal{II})$, $p(\mathbf{y}_\mathcal{D}|\mathbf{f}_\mathcal{D})=\mathcal{N}(\mathbf{f}_\mathcal{D},\mathbf{C}_{\mathcal{DD}})$}, and\vspace{-0.0mm}
\begin{equation}
\hspace{0mm}p(\mathbf{f}_\mathcal{D}|\mathbf{s}_\mathcal{I},\boldsymbol{\theta})=\displaystyle\mathcal{N}(\mathbf{K}_\mathcal{DI}\mathbf{\Sigma}_\mathcal{II}^{-1}\mathbf{s}_\mathcal{I},\mathbf{K}_\mathcal{DD} -\mathbf{K}_\mathcal{DI}\mathbf{\Sigma}_\mathcal{II}^{-1}\mathbf{K}_\mathcal{ID})
\label{Modified Gaussian Process Model}\vspace{-0.0mm}
\end{equation}
such that $\mathbf{K}_\mathcal{DI}$ is previously defined in Section~\ref{Seperation of Gaussian Process} and
$\mathbf{K}_\mathcal{ID} = \mathbf{K}^{\top}_\mathcal{DI}$. However, the integration in~\eqref{E13} (and hence $p(\mathbf{s}_\mathcal{I},\boldsymbol{\theta}|\mathbf{y}_\mathcal{D})$) cannot be evaluated in closed form. To resolve this, instead of using exact inference, we adopt variational inference to approximate the posterior distribution $p(\mathbf{f}_\mathcal{D},\mathbf{s}_\mathcal{I},\boldsymbol{\theta}|\mathbf{y}_\mathcal{D})=p(\mathbf{f}_\mathcal{D}|\mathbf{s}_\mathcal{I},\boldsymbol{\theta})\ p(\mathbf{s}_\mathcal{I},\boldsymbol{\theta}|\mathbf{y}_\mathcal{D})$ with a factorized variational distribution 
		$q(\mathbf{f}_\mathcal{D},\mathbf{s}_\mathcal{I},\boldsymbol{\theta})\triangleq p(\mathbf{f}_\mathcal{D}|\mathbf{s}_\mathcal{I},\boldsymbol{\theta})\ q(\mathbf{s}_\mathcal{I})\ q(\boldsymbol{\theta})$ 
where $p(\mathbf{f}_\mathcal{D}|\mathbf{s}_\mathcal{I},\boldsymbol{\theta})$ is the exact training conditional~\eqref{Modified Gaussian Process Model}, $	q(\mathbf{s}_\mathcal{I})\triangleq\mathcal{N}(\mathbf{m},\mathbf{S})$, 
$q(\boldsymbol{\theta}) \triangleq q(\mathbf{\Lambda})\ q(\sigma_f)$, 
$q(\mathbf{\Lambda})\triangleq\prod_{i=1}^{d}\mathcal{N}(\lambda_i|\nu_i,\xi_i)$ with $\boldsymbol{\nu} \triangleq (\nu_1,\ldots,\nu_d)^\top$ and $\mathbf{\Xi} \triangleq \mathrm{diag}[\xi_1,\ldots,\xi_d]$, and $q(\sigma_f)\triangleq\mathcal{N}(\alpha, \beta)$. Then, the log-marginal likelihood $\log p(\mathbf{y}_\mathcal{D})$ can be decomposed into a sum of its variational lower bound $\mathcal{L}(q)$ and the nonnegative KL distance between the variational distribution $q(\mathbf{f}_\mathcal{D},\mathbf{s}_\mathcal{I},\boldsymbol{\theta})$ and the posterior distribution $p(\mathbf{f}_\mathcal{D},\mathbf{s}_\mathcal{I},\boldsymbol{\theta}|\mathbf{y}_\mathcal{D})$, the latter of which quantifies the gap between $\log p(\mathbf{y}_\mathcal{D})$ and $\mathcal{L}(q)$, that is, 
\begin{equation}
\hspace{0mm}\log p(\mathbf{y}_\mathcal{D})\hspace{-0mm}=\hspace{-0mm}\mathcal{L}(q)+\mathrm{KL}(q(\mathbf{f}_\mathcal{D},\mathbf{s}_\mathcal{I},\boldsymbol{\theta})|| p(\mathbf{f}_\mathcal{D},\mathbf{s}_\mathcal{I},\boldsymbol{\theta}|\mathbf{y}_\mathcal{D})), \label{log likelihood}\vspace{-0mm}
\end{equation}
as derived in Appendix~\ref{Derivation of Equation (15)} of~\cite{HaibinAPP} where\vspace{-0.6mm} 
\begin{equation}
\hspace{-0mm}\mathcal{L}(q) \triangleq\hspace{-0mm} \int \hspace{-0mm}q(\mathbf{f}_\mathcal{D},\mathbf{s}_\mathcal{I},\boldsymbol{\theta})\log\frac{p(\mathbf{y}_\mathcal{D},\mathbf{f}_\mathcal{D},\mathbf{s}_\mathcal{I},\boldsymbol{\theta})}{q(\mathbf{f}_\mathcal{D},\mathbf{s}_\mathcal{I},\boldsymbol{\theta})}\ \mathrm{d}\mathbf{f}_\mathcal{D}\ \mathrm{d}\mathbf{s}_\mathcal{I}\ \mathrm{d}\boldsymbol{\theta}.\vspace{-0.6mm}
\end{equation}

\emph{Remark 3:} The likelihood term $p(\mathbf{y}_{\mathcal{D}})$~\eqref{E13} in~\eqref{log likelihood} is a constant with respect to $q(\mathbf{s}_\mathcal{I})$ and $q(\boldsymbol{\theta})$ (specifically, their parameters $\mathbf{m},\mathbf{S},\boldsymbol{\nu},\mathbf{\Xi},\alpha, \beta$). Consequently, maximizing $\mathcal{L}(q)$ with respect to $q(\mathbf{f}_\mathcal{D},\mathbf{s}_\mathcal{I},\boldsymbol{\theta})$ is equivalent to minimizing $\mathrm{KL}(q(\mathbf{f}_\mathcal{D},\mathbf{s}_\mathcal{I},\boldsymbol{\theta})|| p(\mathbf{f}_\mathcal{D},\mathbf{s}_\mathcal{I},\boldsymbol{\theta}|\mathbf{y}_\mathcal{D}))$. This equivalence, however, does not hold for existing variational SGPR models and their stochastic and distributed variants optimizing point estimates of all hyperparameters, as discussed in Section~\ref{sect:intro}.\vspace{1mm}

The variational inference framework of \cite{Titsias13} is similar in spirit to the above. However, the framework of \cite{Titsias13} assumes i.i.d. observation noises (i.e., $\mathbf{C}_{\mathcal{DD}} = \sigma_n^2\mathbf{I}$ and $\zeta = 1$) and ignores their correlation, 
which consequently yields the VBDTC approximation (see Remark $4$ later).
The challenge remains in investigating whether the other more refined SGPR models spanned by the unifying view of~\cite{candela05} (e.g., FITC, PITC, PIC) are amenable to such a variational Bayesian treatment since they have been empirically demonstrated~\cite{NghiaICML15,HoangICML16} to give better predictive performance than DTC.
%
%

To address this challenge, our key idea is to relax the strong assumption of i.i.d. observation noises with constant variance $\sigma^2_n$ imposed by VBDTC and allow observation noises to be correlated with some structure across the input space, hence being robust to different noise correlation structures and in turn catering to more realistic applications of GP. 
Interestingly, this results in a noise-robust family of \emph{variational Bayesian SGPR} (VBSGPR) models (e.g., VBDTC, VBFITC, VBPIC), which we will describe below.

Let $\mathbf{C}_{\mathcal{DD}}\triangleq\mathrm{blkdiag}[\mathbf{K}^{\varepsilon}_{\mathcal{D}\mathcal{D}}-\mathbf{K}^{\varepsilon}_{\mathcal{D}\mathcal{U}}\mathbf{K}^{\varepsilon^{-1}}_{\mathcal{U}\mathcal{U}}\mathbf{K}^{\varepsilon}_{\mathcal{U}\mathcal{D}}]+\sigma^2_n\mathbf{I}$ be a block-diagonal noise covariance matrix constructed from the $B$ diagonal blocks of $\mathbf{K}^{\varepsilon}_{\mathcal{D}\mathcal{D}}-\mathbf{K}^{\varepsilon}_{\mathcal{D}\mathcal{U}}\mathbf{K}^{\varepsilon^{-1}}_{\mathcal{U}\mathcal{U}}\mathbf{K}^{\varepsilon}_{\mathcal{U}\mathcal{D}}+\sigma^2_n\mathbf{I}$, each of which is a matrix $\mathbf{C}_{\mathcal{D}_i\mathcal{D}_i}\triangleq\mathbf{K}^{\varepsilon}_{\mathcal{D}_i\mathcal{D}_i}-\mathbf{K}^{\varepsilon}_{\mathcal{D}_i\mathcal{U}}\mathbf{K}^{\varepsilon^{-1}}_{\mathcal{U}\mathcal{U}}\mathbf{K}^{\varepsilon}_{\mathcal{U}\mathcal{D}_i}+\sigma^2_n\mathbf{I}$ for $i=1,\ldots,B$, and $\mathbf{K}^{\varepsilon}_{\mathcal{D}\mathcal{D}} \triangleq (k^{\varepsilon}_{\mathbf{x}\mathbf{x}'})_{\mathbf{x},\mathbf{x}' \in \mathcal{D}}$, $\mathbf{K}^{\varepsilon}_{\mathcal{D}\mathcal{U}} \triangleq (k^{\varepsilon}_{\mathbf{x}\mathbf{x}'})_{\mathbf{x} \in \mathcal{D},\mathbf{x}' \in \mathcal{U}}$, $\mathbf{K}^{\varepsilon}_{\mathcal{U}\mathcal{U}}\triangleq (k^{\varepsilon}_{\mathbf{x}\mathbf{x}'})_{\mathbf{x},\mathbf{x}' \in \mathcal{U}}$, and $\mathbf{K}^{\varepsilon}_{\mathcal{U}\mathcal{D}} \triangleq \mathbf{K}^{\varepsilon^\top}_{\mathcal{D}\mathcal{U}}$ are matrices with components $k^{\varepsilon}_{\mathbf{x}\mathbf{x}'}$ defined by a covariance function similar to that used for $k_{\mathbf{x}\mathbf{x}'}$ (Section~\ref{full}) but with different hyperparameter values\footnote{We do not assign any prior over the hyperparameters of $k^{\varepsilon}_{\mathbf{x}\mathbf{x}'}$ and the noise variance $\sigma^2_n$. 
Instead, they are treated as parameters optimized to maximize $\mathcal{L}(q)$ via stochastic gradient ascent \cite{Lawrence13}. 
In our experiments, we observe that even if we set the hyperparameters of $k^{\varepsilon}_{\mathbf{x}\mathbf{x}'}$ by hand, the predictive performance does not vary much and our VBPIC approximation can significantly outperform the state-of-the-art variational SGPR models and their stochastic and distributed variants. A Bayesian treatment of these hyperparameters is highly non-trivial due to a complication similar to that discussed in Section~\ref{Bayesian SGP} 
and will be investigated in our future work.}. Our first major result ensues:\vspace{1mm}
\begin{theorem}
$\mathcal{L}(q)$ in~\eqref{log likelihood} can be analytically evaluated as \vspace{-0mm} 
\begin{equation}
\hspace{-0mm}
\begin{array}{rcl}
\mathcal{L}(q) &\hspace{-2.4mm}= &\hspace{-2.4mm}\displaystyle\frac{1}{2}\Big(2\mathbf{m}^{\hspace{-0.5mm}\top}\hspace{-0.5mm}\mathbf{\Sigma}_\mathcal{II}^{-1}\mathbf{\Omega}_\mathcal{ID}\mathbf{C}_\mathcal{DD}^{-1}\mathbf{y}_\mathcal{D}\hspace{-0.5mm}-\hspace{-0.5mm}\mathbf{m}^{\hspace{-0.5mm}\top}\hspace{-0.5mm}\mathbf{Q}\mathbf{m} -\mathrm{Tr}[\mathbf{S}\mathbf{Q}]\vspace{0.5mm}
\\
&&\hspace{-2.4mm}-\hspace{0mm}\mathrm{Tr}[\mathbf{C}_\mathcal{DD}^{-1}\mathbf{\Upsilon}_\mathcal{DD}]\hspace{-0.7mm}+\hspace{-0.7mm}\mathrm{Tr}[\mathbf{\Sigma}_\mathcal{II}^{-1}\mathbf{\Psi}_\mathcal{II}] +\hspace{-0.5mm}\log|\mathbf{S}|\vspace{0.5mm}
\\
&&\hspace{-2.4mm}-\boldsymbol{\nu}^\top\boldsymbol{\nu}\hspace{-0.5mm}-\hspace{-0.5mm}\mathrm{Tr}[\mathbf{\Xi}]\hspace{-0.5mm}+\hspace{-0.5mm}\log|\mathbf{\Xi}|\hspace{-0.5mm}-\hspace{-0.5mm}\alpha^2\hspace{-0.5mm} -\hspace{-0.5mm} \beta\hspace{-0.5mm} +\hspace{-0.5mm} \log\beta\Big)\hspace{-1mm}+\hspace{-0.5mm}\mathrm{const}\vspace{0mm} 
\vspace{0mm}
\end{array}\hspace{-4.8mm}
\label{L(q)}
\end{equation} 
where $\mathbf{Q} \triangleq \mathbf{\Sigma}_\mathcal{II}^{-1}\mathbf{\Psi}_\mathcal{II}\mathbf{\Sigma}_\mathcal{II}^{-1}\hspace{-0mm}+\hspace{-0mm}\mathbf{\Sigma}_\mathcal{II}^{-1}$. More interestingly, using the above expression, it can be shown that $\mathcal{L}(q)$ is maximized at $q^*(\mathbf{s}_\mathcal{I})=\mathcal{N}(\mathbf{m}^*,\mathbf{S}^*)$ where
\begin{equation}
\begin{array}{rcl}
\mathbf{m}^*&\hspace{-2.4mm}\triangleq&\hspace{-2.4mm}\displaystyle\mathbf{\Sigma}_\mathcal{II}(\mathbf{\Sigma}_\mathcal{II}+\mathbf{\Psi}_\mathcal{II})^{-1}\mathbf{\Omega}_\mathcal{ID}\mathbf{C}^{-1}_\mathcal{DD}\mathbf{y}_\mathcal{D}\ ,\vspace{0.5mm}\\
\mathbf{S}^*&\hspace{-2.4mm}\triangleq&\hspace{-2.4mm}\displaystyle\mathbf{\Sigma}_\mathcal{II}(\mathbf{\Sigma}_\mathcal{II}+\mathbf{\Psi}_\mathcal{II})^{-1}\mathbf{\Sigma}_\mathcal{II}
\end{array}
\label{q(u)}\vspace{-0mm}
\end{equation} 
such that $\mathbf{\Omega}_\mathcal{ID}\triangleq\mathbb{E}_{q(\boldsymbol{\theta})}[\mathbf{K}_{\mathcal{ID}}]$, $\mathbf{\Upsilon}_\mathcal{DD}\triangleq\mathbb{E}_{q(\boldsymbol{\theta})}[\mathbf{K}_{\mathcal{DD}}]$, $\mathbf{\Psi}_\mathcal{II}\triangleq\mathbb{E}_{q(\boldsymbol{\theta})}[\mathbf{K}_{\mathcal{ID}}\mathbf{C}^{-1}_\mathcal{DD}\mathbf{K}_{\mathcal{DI}}]$, 
and $\mathrm{const}$ absorbs all terms indep. of $\mathbf{m}$, $\mathbf{S}$, $\boldsymbol{\nu}$, $\mathbf{\Xi}, \alpha, \beta$.\vspace{1mm}
\label{thm1}
\end{theorem}
Its proof is in Appendix~\ref{Derivation of the Lower Bound} of~\cite{HaibinAPP}. Appendix~\ref{Derivation of Omega, Psi and Upsilon} of~\cite{HaibinAPP} gives the closed-form expressions of $\mathbf{\Omega}_\mathcal{ID}$, $\mathbf{\Upsilon}_\mathcal{DD}$, and $\mathbf{\Psi}_\mathcal{II}$.\vspace{1mm}

\emph{Remark 4:} Note that $q^*(\mathbf{s}_\mathcal{I})$ in Theorem \ref{thm1} closely resembles that of PIC and PITC (see eqs.~$64$ and~$65$ in Appendix D.$1$.$1$ of~\cite{NghiaICML15}) except for the expectation over hyperparameters $\boldsymbol{\theta}$ due to the variational Bayesian treatment. So, we call them VBPIC and VBPITC, respectively. By setting $B = |\mathcal{D}|$, $\mathbf{C}_{\mathcal{DD}}$ becomes a diagonal matrix to give VBFIC and VBFITC. When $\mathbf{C}_{\mathcal{DD}}=\sigma^2_n\mathbf{I}$, $q^*(\mathbf{s}_\mathcal{I})$~\eqref{q(u)} resembles that of DTC (see eqs.~$68$ and~$69$ in Appendix D.$1$.$3$ of \cite{NghiaICML15}) except for the expectation over $\boldsymbol{\theta}$ due to the variational Bayesian treatment and coincides with that in Appendix B.$1$ of \cite{Titsias13}. So, we refer to it as VBDTC.\vspace{1mm} 

\emph{Remark 5:} In the non-Bayesian setting of the hyperparameters, it has been previously established that the predictive distribution of FITC can be reproduced as a direct result of applying either variational inference \cite{Titsias09a} with $\mathbf{C}_{\mathcal{DD}}=\mathrm{diag}[\mathbf{K}_{\mathcal{D}\mathcal{D}}-\mathbf{K}_{\mathcal{D}\mathcal{U}}\mathbf{K}^{-1}_{\mathcal{U}\mathcal{U}}\mathbf{K}_{\mathcal{U}\mathcal{D}}]+\sigma^2_n\mathbf{I}$ or expectation propagation \cite{Matthias2016Understanding} on the FGPR model. Our derivation of VBFITC is in fact similar in spirit to that of \cite{Titsias09a} except for our variational Bayesian treatment of its hyperparameters. On the other hand, it is unclear whether FITC's equivalent EP derivation in \cite{Matthias2016Understanding} can be easily extended to incorporate a Bayesian treatment of its hyperparameters.
%
%
\section{Stochastic Optimization}	
	\label{Stochastic Variational Inference for GPR}
	The VBDTC approximation~\cite{Titsias13} has explicitly plugged the optimal $q^*(\mathbf{s}_\mathcal{I})$ (see Theorem~\ref{thm1}) into $\mathcal{L}(q)$~\eqref{L(q)} and reduced it to $\mathcal{L}(q)$~\eqref{eq:TitsiasL(q)} in Appendix~\ref{Derivation of the Lower Bound} of~\cite{HaibinAPP}. Given $\mathcal{L}(q)$~\eqref{eq:TitsiasL(q)}, the parameters $\boldsymbol{\nu}$ and $\mathbf{\Xi}$ of $q(\mathbf{\Lambda})$ and $\alpha$ and $\beta$ of $q(\sigma_f)$ can be optimized via gradient ascent. However, evaluating the exact gradients
	$\partial\mathcal{L}/\partial\boldsymbol{\nu}$, $\partial\mathcal{L}/\partial\mathbf{\Xi}$, $\partial\mathcal{L}/\partial \alpha$ and $\partial\mathcal{L}/\partial \beta$  
	incur $\mathcal{O}(|\mathcal{D}||\mathcal{I}|^2)$ time,
	which scales poorly in the data size $|\mathcal{D}|$. To overcome the above issue of scalability, we utilize stochastic gradient ascent updates instead of exact ones, which requires the stochastic gradients to be unbiased estimators of the exact gradients to guarantee convergence.
	The key idea is to iteratively compute the stochastic gradients by randomly sampling a single or few mini-batches of data from the dataset (i.e., comprising $B$ disjoint mini-batches) whose incurred time per iteration is independent of data size $|\mathcal{D}|$.
	To achieve this, an important requirement is the decomposability of $\mathcal{L}(q)$~\eqref{eq:TitsiasL(q)} into a summation of $B$ terms, each of which is associated with a mini-batch $(\mathcal{D}_i, \mathbf{y}_{\mathcal{D}_i})$ of data of size $|\mathcal{D}_i|=\mathcal{O}(|\mathcal{I}|)$	
that can be exploited for computing the stochastic gradients.
	Unfortunately, $\mathcal{L}(q)$~\eqref{eq:TitsiasL(q)} is not decomposable due to its $(\mathbf{\Sigma}_\mathcal{II}+\mathbf{\Psi}_\mathcal{II})^{-1}$ term.			
	To remedy this, we do not  plug $q^*(\mathbf{s}_\mathcal{I})$~\eqref{q(u)} into $\mathcal{L}(q)$~\eqref{L(q)} to yield~\eqref{eq:TitsiasL(q)} but
	instead jointly treat $q(\mathbf{s}_\mathcal{I})$, $q(\mathbf{\Lambda})$, and $q(\sigma_f)$ as variational parameters, which enables the decomposability of $\mathcal{L}(q)$~\eqref{L(q)}:\vspace{1mm}
\begin{corollary} $\mathcal{L}(q)$~\eqref{L(q)} (Theorem~\ref{thm1}) can be decomposed into\vspace{-0mm}	
	\begin{equation*}
		\hspace{-1.7mm}
		\begin{array}{l}
			\mathcal{L}(q)=\displaystyle\sum_{i=1}^B\mathcal{L}_i(q) \hspace{-0.5mm}+\hspace{-0.5mm} \frac{1}{2}\Big( \hspace{-1mm}-\hspace{-0.5mm}\mathbf{m}^\top\mathbf{\Sigma}_\mathcal{II}^{-1}\mathbf{m}\hspace{-0.5mm}- \hspace{-0.5mm}\mathrm{Tr}[\mathbf{S}\mathbf{\Sigma}_\mathcal{II}^{-1}]\hspace{-0.5mm}+\hspace{-0.5mm}\log|\mathbf{S}| \vspace{-0.5mm}\\
			 \quad\qquad\ \displaystyle-\boldsymbol{\nu}^\top\boldsymbol{\nu}\hspace{-0.5mm} -\hspace{-0.5mm}\mathrm{Tr}[\mathbf{\Xi}]\hspace{-0.5mm}+\hspace{-0.5mm}\log|\mathbf{\Xi}|\hspace{-0.5mm}-\hspace{-0.5mm}\alpha^2 \hspace{-0.5mm}-\hspace{-0.5mm} \beta \hspace{-0.5mm}+\hspace{-0.5mm} \log \beta\Big)\hspace{-1mm} +\hspace{-0.5mm}\mathrm{const}\ ,\\
			\mathcal{L}_i(q)\hspace{-0.5mm}\triangleq\hspace{-0.5mm}\displaystyle \frac{1}{2}\Big(2\mathbf{m}^{\hspace{-0.5mm}\top}\mathbf{\Sigma}_\mathcal{II}^{-1}\mathbf{\Omega}_{\mathcal{I}\mathcal{D}_i}\mathbf{C}_{\mathcal{D}_i\mathcal{D}_i}^{-1}\mathbf{y}_{\mathcal{D}_i}\hspace{-1mm}-\hspace{-0.5mm}\mathbf{m}^{\hspace{-0.5mm}\top}\mathbf{\Sigma}_\mathcal{II}^{-1}\mathbf{\Psi}_\mathcal{II}^{i}\mathbf{\Sigma}_\mathcal{II}^{-1}\mathbf{m}\\
			\quad\ \displaystyle -\mathrm{Tr}[\mathbf{S}\mathbf{\Sigma}_\mathcal{II}^{-1}\mathbf{\Psi}_\mathcal{II}^{i}\mathbf{\Sigma}_\mathcal{II}^{-1}]\hspace{-0.5mm}-\hspace{-0.5mm}\mathrm{Tr}[\mathbf{C}_{\mathcal{D}_i\mathcal{D}_i}^{-1}\mathbf{\Upsilon}_{\mathcal{D}_i\mathcal{D}_i}] \hspace{-0.5mm}+\hspace{-0.5mm}\mathrm{Tr}[\mathbf{\Sigma}_\mathcal{II}^{-1}\mathbf{\Psi}_\mathcal{II}^{i}]\Big)
		\end{array}
		\label{Lower Bound}		
	\end{equation*}
	where 
	$\mathbf{\Psi}_\mathcal{II}^{i}\triangleq\mathbb{E}_{q(\boldsymbol{\theta})}[\mathbf{K}_{\mathcal{I}\mathcal{D}_i}\mathbf{C}^{-1}_{\mathcal{D}_i\mathcal{D}_i}\mathbf{K}_{\mathcal{D}_i\mathcal{I}}]$.\vspace{1mm}
	\label{decompo}
\end{corollary}	 
	
	Our main result below exploits the decomposability of $\mathcal{L}(q)$ in Corollary~\ref{decompo} to derive  stochastic gradients $\partial\widetilde{\mathcal{L}}/\partial\mathbf{m}$, $\partial\widetilde{\mathcal{L}}/\partial\mathbf{S}$, $\partial\widetilde{\mathcal{L}}/\partial\boldsymbol{\nu}$, $\partial\widetilde{\mathcal{L}}/\partial\mathbf{\Xi}$, $\partial\widetilde{\mathcal{L}}/\partial \alpha$, and $\partial\widetilde{\mathcal{L}}/\partial \beta$ that are unbiased estimators of their respective exact gradients,
	which is the key contribution of our work in this paper:\vspace{1mm}
	\begin{theorem}
		Let $\mathcal{S}$ be a set of i.i.d. samples drawn from a uniform distribution over $\{1,2,\dots,B\}$. 
		Construct the stochastic gradients $\partial\widetilde{\mathcal{L}}/\partial\mathbf{m}$, $\partial\widetilde{\mathcal{L}}/\partial\mathbf{S}$, $\partial\widetilde{\mathcal{L}}/\partial\boldsymbol{\nu}$, $\partial\widetilde{\mathcal{L}}/\partial\mathbf{\Xi}$, $\partial\widetilde{\mathcal{L}}/\partial \alpha$, and $\partial\widetilde{\mathcal{L}}/\partial\beta$ 
using the mini-batches $(\mathcal{D}_s, \mathbf{y}_{\mathcal{D}_s})$ for $s\in\mathcal{S}$ and current estimates of $(\mathbf{m}, \mathbf{S}, \boldsymbol{\nu}, \mathbf{\Xi}, \alpha, \beta)$ according to~\eqref{zoo} in Appendix~\ref{A.3} of~\cite{HaibinAPP}. 
		Then, $\mathbb{E}[\partial\widetilde{\mathcal{L}}/\partial\mathbf{m}]=\partial{\mathcal{L}}/\partial\mathbf{m}$, 
		$\mathbb{E}[\partial\widetilde{\mathcal{L}}/\partial\mathbf{S}]=\partial{\mathcal{L}}/\partial\mathbf{S}$, 
		$\mathbb{E}[\partial\widetilde{\mathcal{L}}/\partial\boldsymbol{\nu}]=\partial{\mathcal{L}}/\partial\boldsymbol{\nu}$,
		$\mathbb{E}[\partial\widetilde{\mathcal{L}}/\partial\mathbf{\Xi}]=\partial{\mathcal{L}}/\partial\mathbf{\Xi}$, $\mathbb{E}[\partial\widetilde{\mathcal{L}}/\partial \alpha]=\partial{\mathcal{L}}/\partial \alpha$, and \ $\mathbb{E}[\partial\widetilde{\mathcal{L}}/\partial \beta]=\partial{\mathcal{L}}/\partial \beta$.
		\label{thm2}\vspace{1mm}
	\end{theorem}	
Its proof is in Appendix~\ref{A.3} of~\cite{HaibinAPP}.\vspace{1mm} 

\emph{Remark 6:} The stochastic gradients (Theorem~\ref{thm2}) can be computed in closed form in $\mathcal{O}(|\mathcal{S}||\mathcal{I}|^3)$ time per iteration that reduces to $\mathcal{O}(|\mathcal{I}|^3)$ time by setting $|\mathcal{S}|=1$ in our experiments.
So, if the number of iterations of stochastic gradient ascent needed for convergence is  much smaller than $t \min(|\mathcal{D}|/|\mathcal{I}|,B)$ where $t$ is the required number of iterations of exact gradient ascent, then our stochastic variants achieve a huge speedup over the corresponding VBSGPR models.\vspace{0mm}
\section{Bayesian Prediction with VBSGPR Models}
\label{predict}
Recall that the predictive distribution $p(f_{\mathbf{x}^*}|\mathbf{y}_\mathcal{D})$ is computationally intractable. We thus approximate it by $q(f_{\mathbf{x}^*}|\mathbf{y}_\mathcal{D})=\int q(f_{\mathbf{x}^*}|\mathbf{y}_{\mathcal{D}_B},\mathbf{s}_\mathcal{I},\mathbf{\Lambda},\sigma_f)\ q^+(\mathbf{s}_\mathcal{I})\ q^+(\mathbf{\Lambda})\ q^+(\sigma_f)\ \mathrm{d}\mathbf{s}_\mathcal{I}\ \mathrm{d}\mathbf{\Lambda}\ \mathrm{d}\mathbf{\sigma_f}$ where $q^+(\mathbf{s}_\mathcal{I})\triangleq\mathcal{N}(\mathbf{m}^+,\mathbf{S}^+)$,  $q^+(\mathbf{\Lambda})\triangleq\prod_{i=1}^{d}\mathcal{N}(\nu_i^+,\xi_i^+)$ with $\boldsymbol{\nu}^+ \triangleq (\nu_1^+,\ldots,\nu_d^+)^\top$ and $\mathbf{\Xi}^+ \triangleq \mathrm{diag}[\xi_1^+,\ldots,\xi_d^+]$, and $q(\sigma_f)\triangleq\mathcal{N}(\alpha^+,\beta^+)$ are obtained from the stochastic gradient ascent updates (Section~\ref{Stochastic Variational Inference for GPR}). Note that $q(f_{\mathbf{x}^*}|\mathbf{y}_{\mathcal{D}_B}, \mathbf{s}_\mathcal{I},\mathbf{\Lambda},\sigma_f)$ is set to $p(f_{\mathbf{x}^*}|\mathbf{s}_\mathcal{I},\mathbf{\Lambda},\sigma_f)$ for the VBPITC, VBFIC, VBFITC, and VBDTC models and to  $p(f_{\mathbf{x}^*}|\mathbf{y}_{\mathcal{D}_B}, \mathbf{s}_\mathcal{I},\mathbf{\Lambda},\sigma_f)$ for the VBPIC model. Although the predictive distribution $q(f_{\mathbf{x}^*}|\mathbf{y}_\mathcal{D})$ is not Gaussian, its predictive mean $\mu_{\mathbf{x}^*|\mathcal{D}}$ and variance $\sigma^2_{\mathbf{x}^*|\mathcal{D}}$ can be computed analytically for VBPITC, VBFIC, VBFITC, and VBDTC and via sampling for VBPIC, as derived in Appendix~\ref{q(y^*)} of~\cite{HaibinAPP}.
Their respective predictive means $\mu_{\mathbf{x}^*|\mathcal{D}}$ closely resemble that of PITC, FIC, FITC, DTC, and PIC (see eqs.~$84$ and~$86$ in Appendix D.$4$ of~\cite{NghiaICML15}) except for the expectations over $\mathbf{\Lambda}$ and $\sigma_f$ due to the variational Bayesian treatment. Their predictive variances $\sigma^2_{\mathbf{x}^*|\mathcal{D}}$ are also similar except for the expectations over $\mathbf{\Lambda}$ and $\sigma_f$ and an additional positive term arising from the uncertainty of $\mathbf{\Lambda}$ and $\sigma_f$.
\section{Experiments and Discussion}
	\label{Experiments and Discussion}
	This section empirically evaluates the predictive performance and time efficiency of the stochastic variants, denoted by VBDTC$+$, VBFITC$+$, and VBPIC$+$, of our VBSGPR models (respectively, VBDTC, VBFITC, and VBPIC).
We will first use the small AIMPEAK dataset \cite{LowUAI13} on traffic speeds of size $41850$ to  evaluate the convergence of the variational distributions $q^+(\mathbf{s}_\mathcal{I})$ and $q^+(\mathbf{\Lambda},\sigma_f)$ induced by our stochastic variants VBDTC$+$, VBFITC$+$, and VBPIC$+$ to, respectively, $q(\mathbf{s}_\mathcal{I})$ and $q(\mathbf{\Lambda},\sigma_f)$ induced by VBDTC \cite{Titsias13}, VBFITC, and VBPIC performing exact gradient ascent updates via \emph{scaled conjugate gradient} (SCG).
To do this, we use the KL distance metric to measure the distance between the variational distributions obtained from the stochastic vs. exact gradient ascent.	

Then, using the real-world TWITTER dataset 
on buzz events of size $583250$ and AIRLINE dataset~\cite{Lawrence13} on flight delays of size $2055733$, we will compare the performance of the stochastic variants of our VBSGPR models with that of the state-of-the-art GP models such as the stochastic variants of variational DTC (SVIGP) \cite{Lawrence13} and variational PIC (PIC$+$) \cite{NghiaICML15}, distributed variational DTC (Dist-VGP) \cite{Yarin14}, and rBCM \cite{deisenroth2015distributed}, all of which find point estimates of hyperparameters.
Such a comparison will demonstrate the benefits of adopting a variational Bayesian treatment of the hyperparameters by our VBSGPR models.
We will also compare the performance of our stochastic VBSGPR models with that of the stochastic variant of \emph{variational Bayesian sparse spectrum GPR} (VSSGPR) model \cite{Gal2015Improving}. 
To evaluate their predictive performance, we use the \emph{root mean square error} (RMSE) metric: $\sqrt{\sum_{\mathbf{x}^*\in\mathcal{T}}(y_{\mathbf{x}^*}-\mu_{\mathbf{x}^*|\mathcal{D}})^2/|\mathcal{T}|}$ and 
the \emph{mean negative log probability} (MNLP) metric: $0.5\sum_{\mathbf{x}^*\in\mathcal{T}}\{(y_{\mathbf{x}^*}-\mu_{\mathbf{x}^*|\mathcal{D}})^2 / \sigma^2_{x^*|\mathcal{D}} + \log(2\pi\sigma^2_{x^*|\mathcal{D}})\} / |\mathcal{T}|$ 
where $\mathcal{T}$ denotes a set of test inputs.

All datasets are modeled using GPs whose prior covariance is defined by the squared exponential covariance function defined in Section~\ref{full}. 
All experiments are run on a Linux system with Intel$\circledR$ Xeon$\circledR$ E$5$-$2683$ CPU at $2.1$GHz with $256$GB memory.

\subsection{Empirical Convergence of Stochastic VBSGPR Models}
	\label{app:AIMPEAK}
The AIMPEAK dataset \cite{LowUAI13} of size $41850$ comprises traffic speeds (km/h) along $775$ road segments of an urban road network during morning peak hours on April $20$, $2011$. Each input (i.e., road segment) denotes a $5$D feature vector of length, number of lanes, speed limit, direction, and time, the last of which comprises $54$ five-minute time slots. The output corresponds to  traffic speed. We randomly select training data of size $1000$, \textcolor{black}{which is partitioned into $B=10$ mini-batches}, and $50$ inducing inputs from the inputs of the training data.	
		\begin{figure}
			\begin{tabular}{ccc}
				\hspace{-2.5mm}\includegraphics[height=2.5cm]{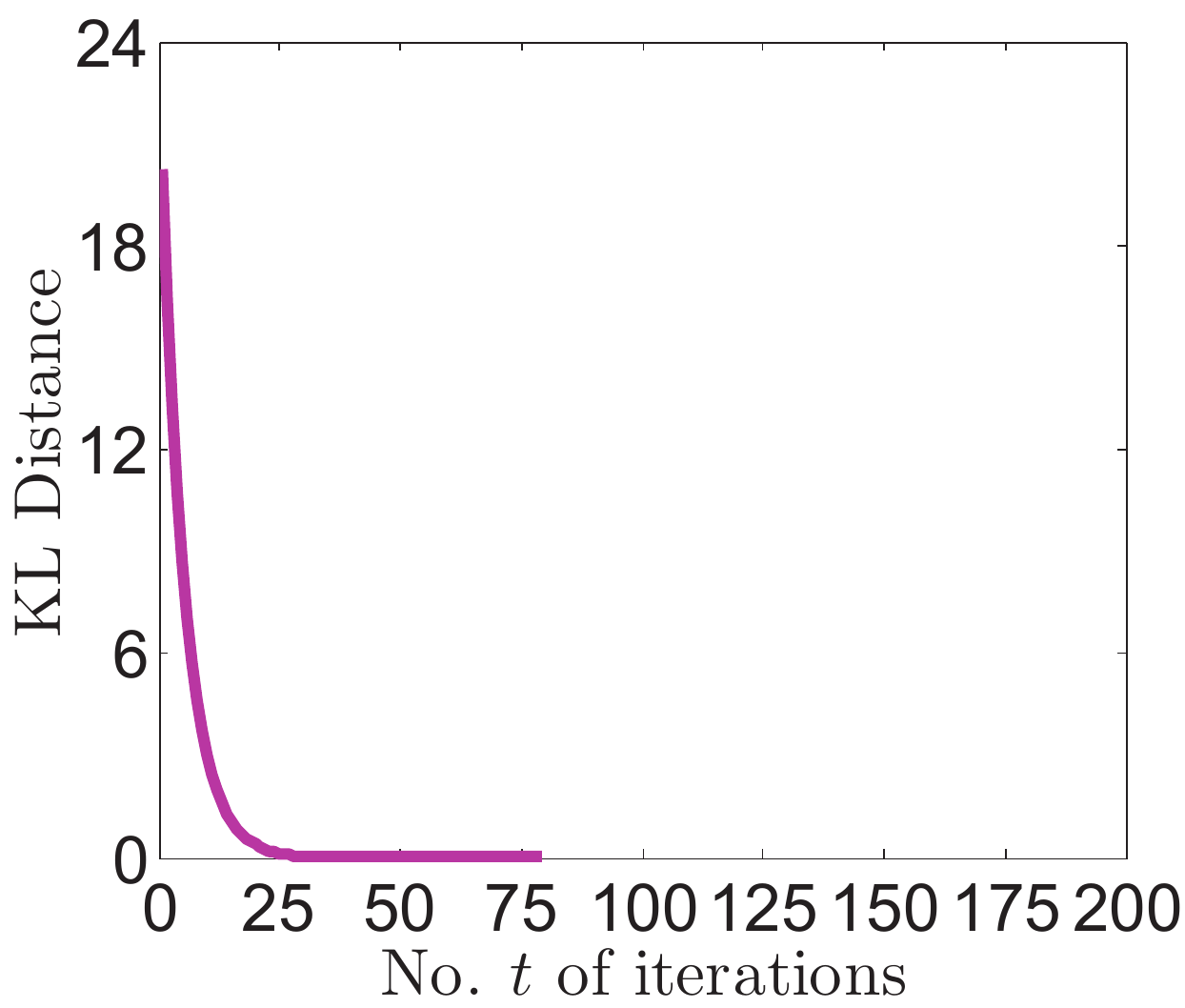} & \hspace{-4mm}\includegraphics[height=2.55cm]{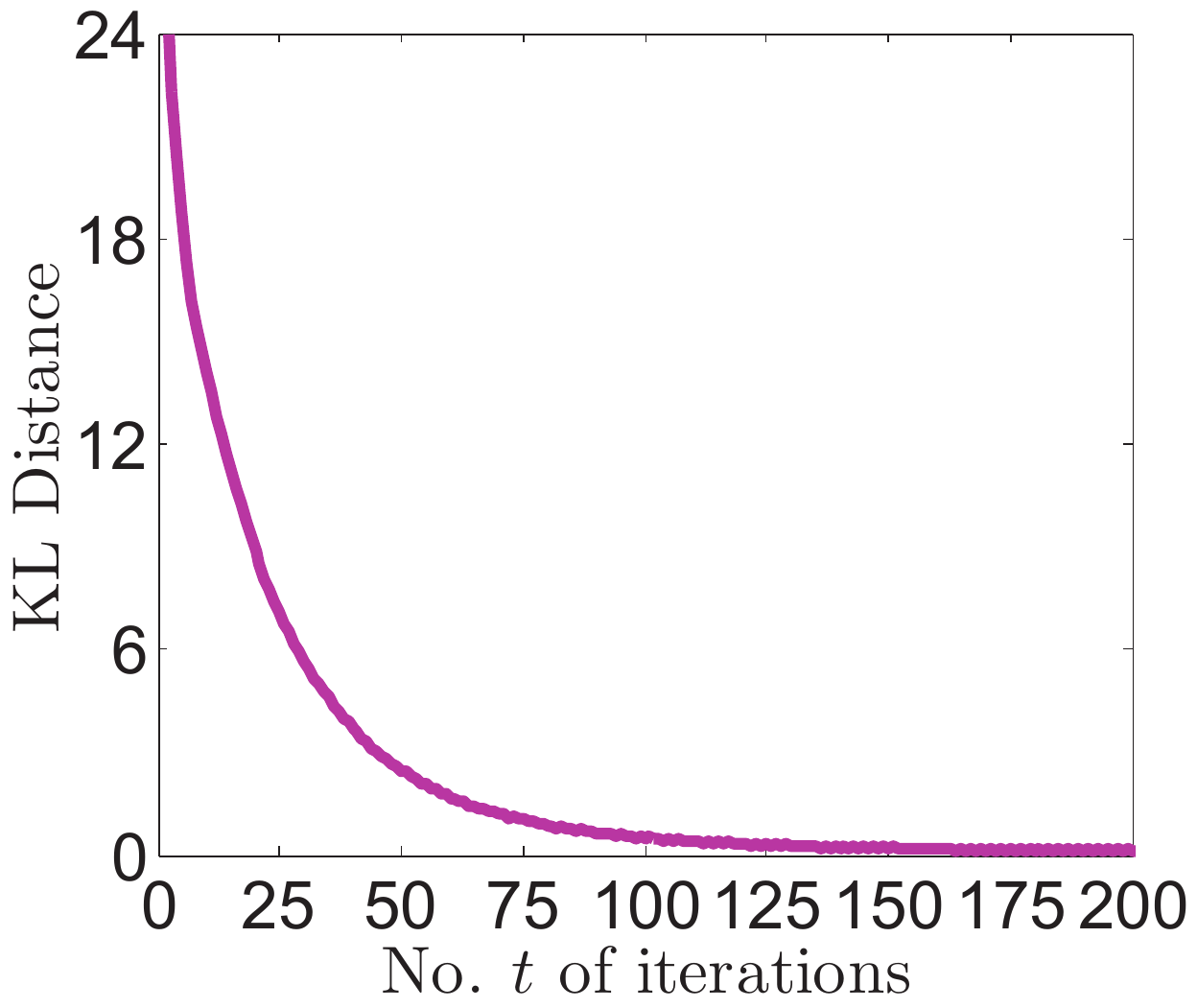} & \hspace{-4mm}\includegraphics[height=2.5cm]{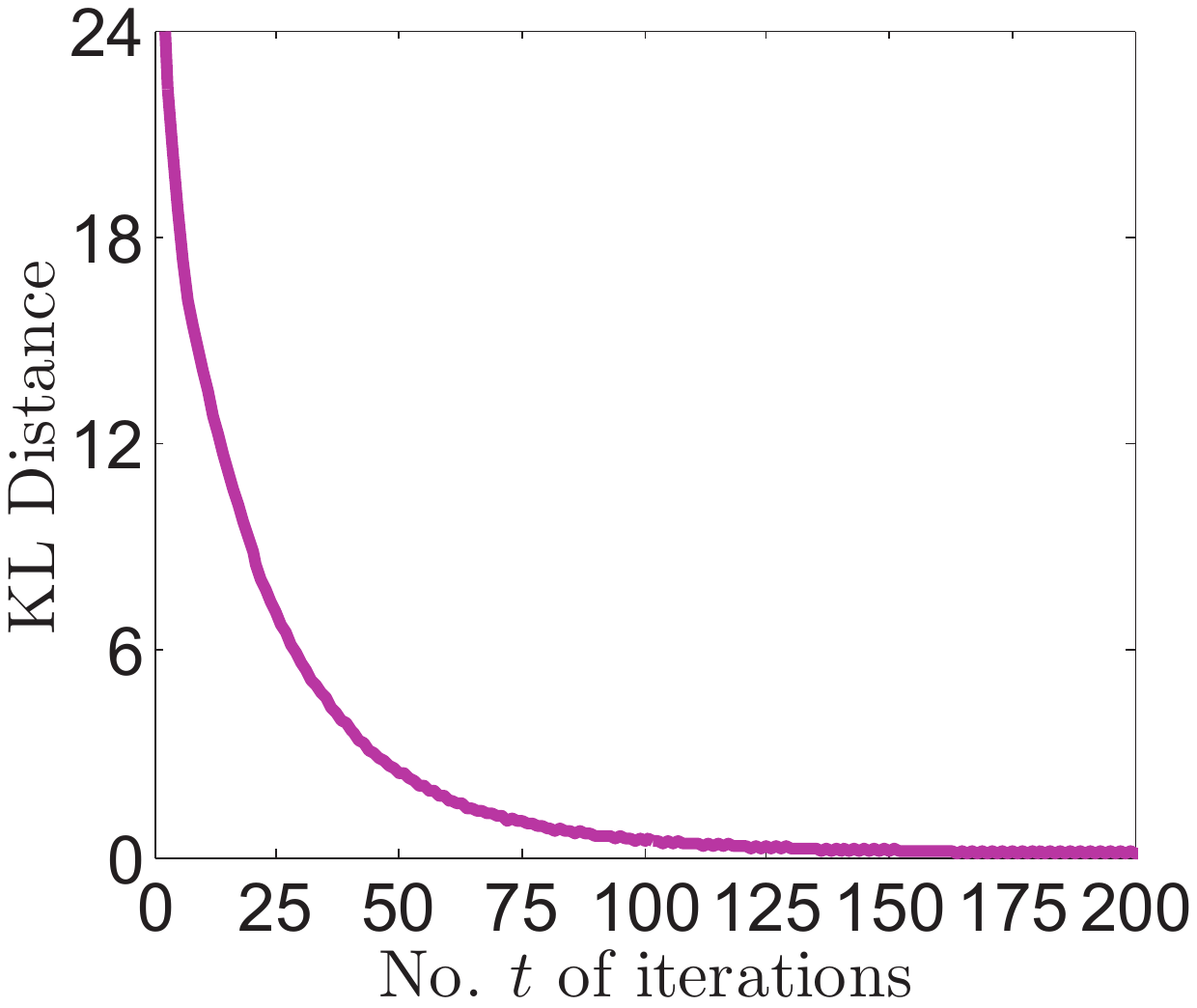}\vspace{-1mm}\\
				\hspace{-2.5mm}(a) & \hspace{-4mm}(b) & \hspace{-4mm}(c) \vspace{0mm}\\
				\hspace{-2.5mm}\includegraphics[height=2.5cm]{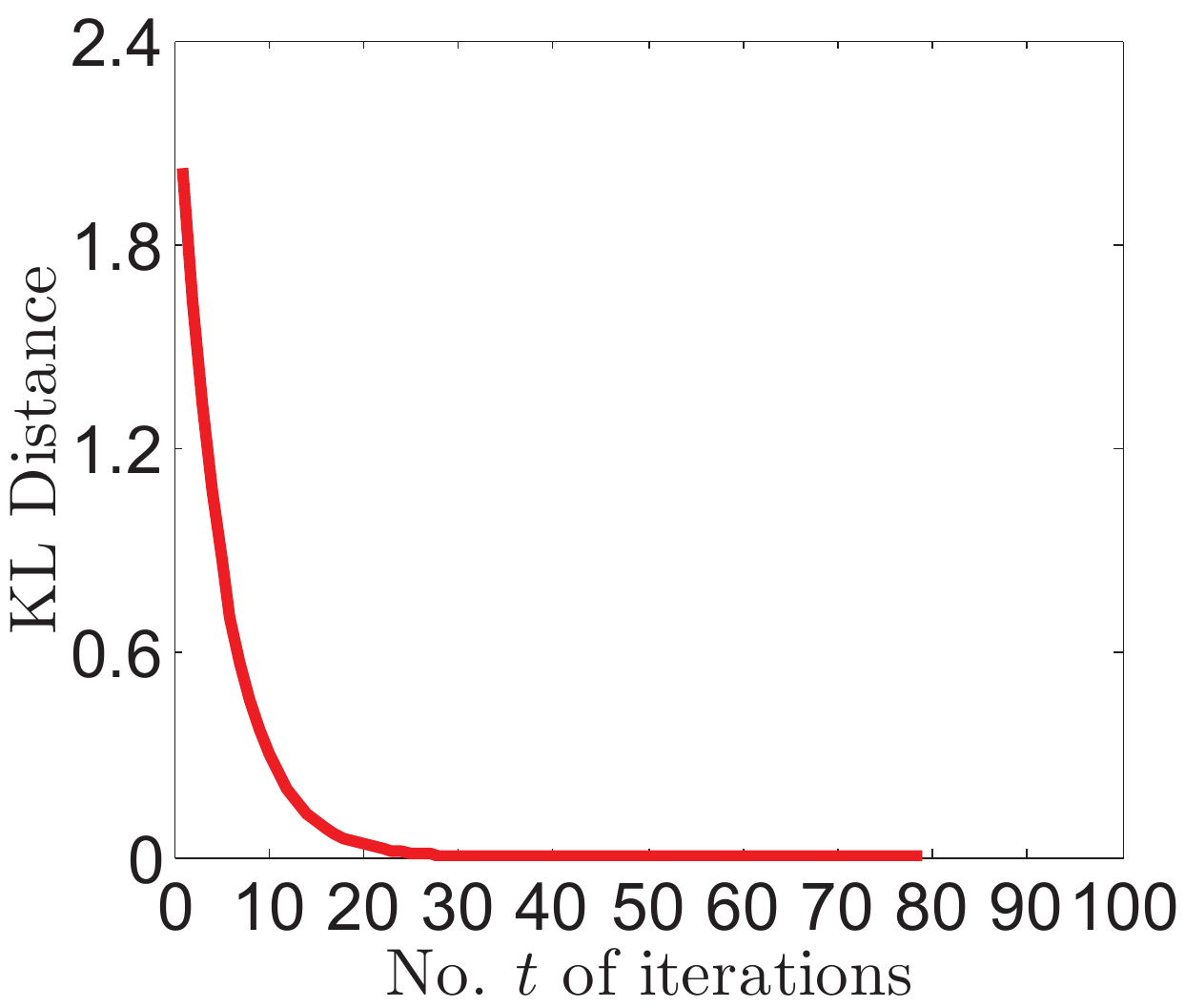} & \hspace{-4mm}\includegraphics[height=2.52cm]{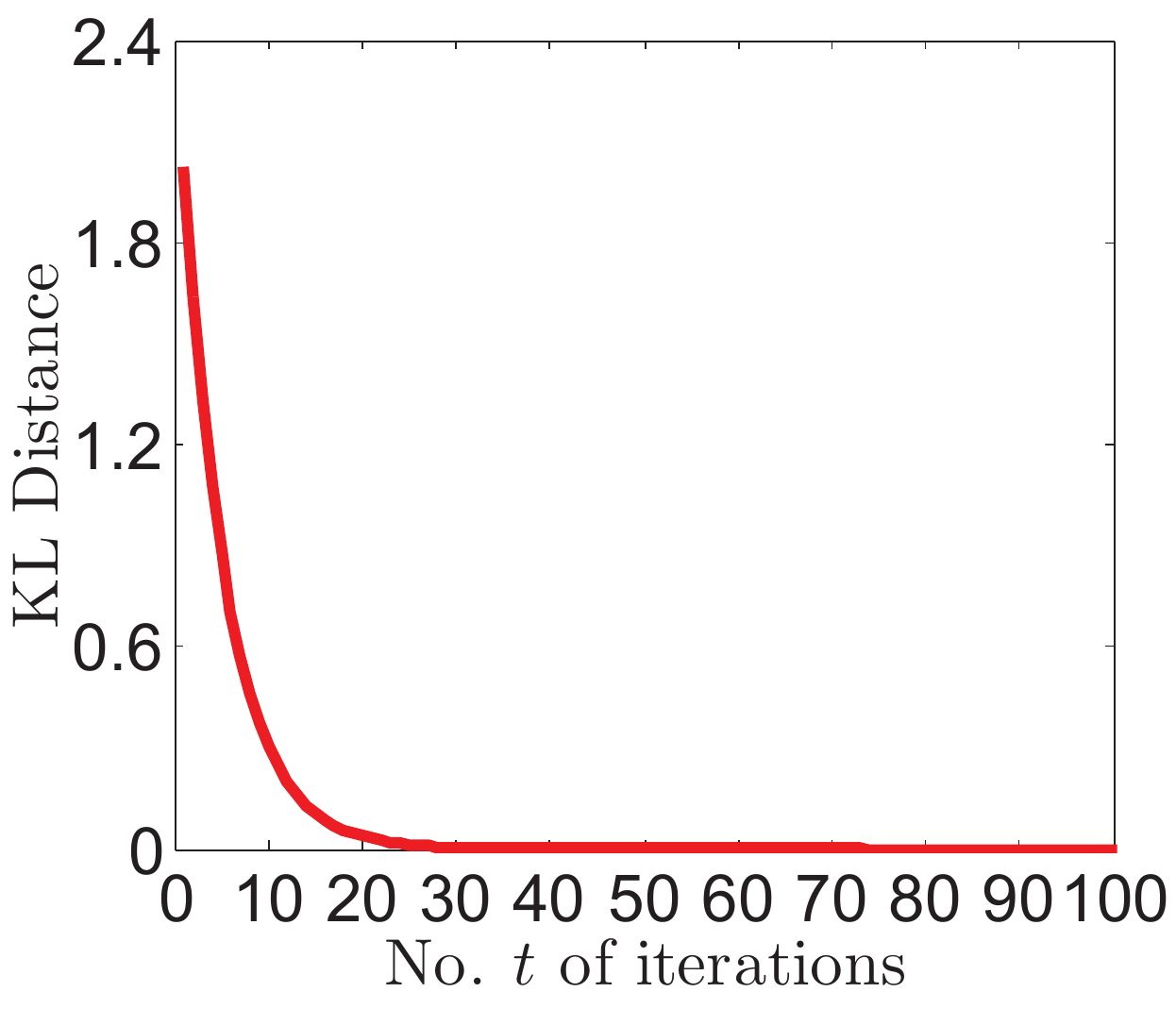} & \hspace{-4mm}\includegraphics[height=2.5cm]{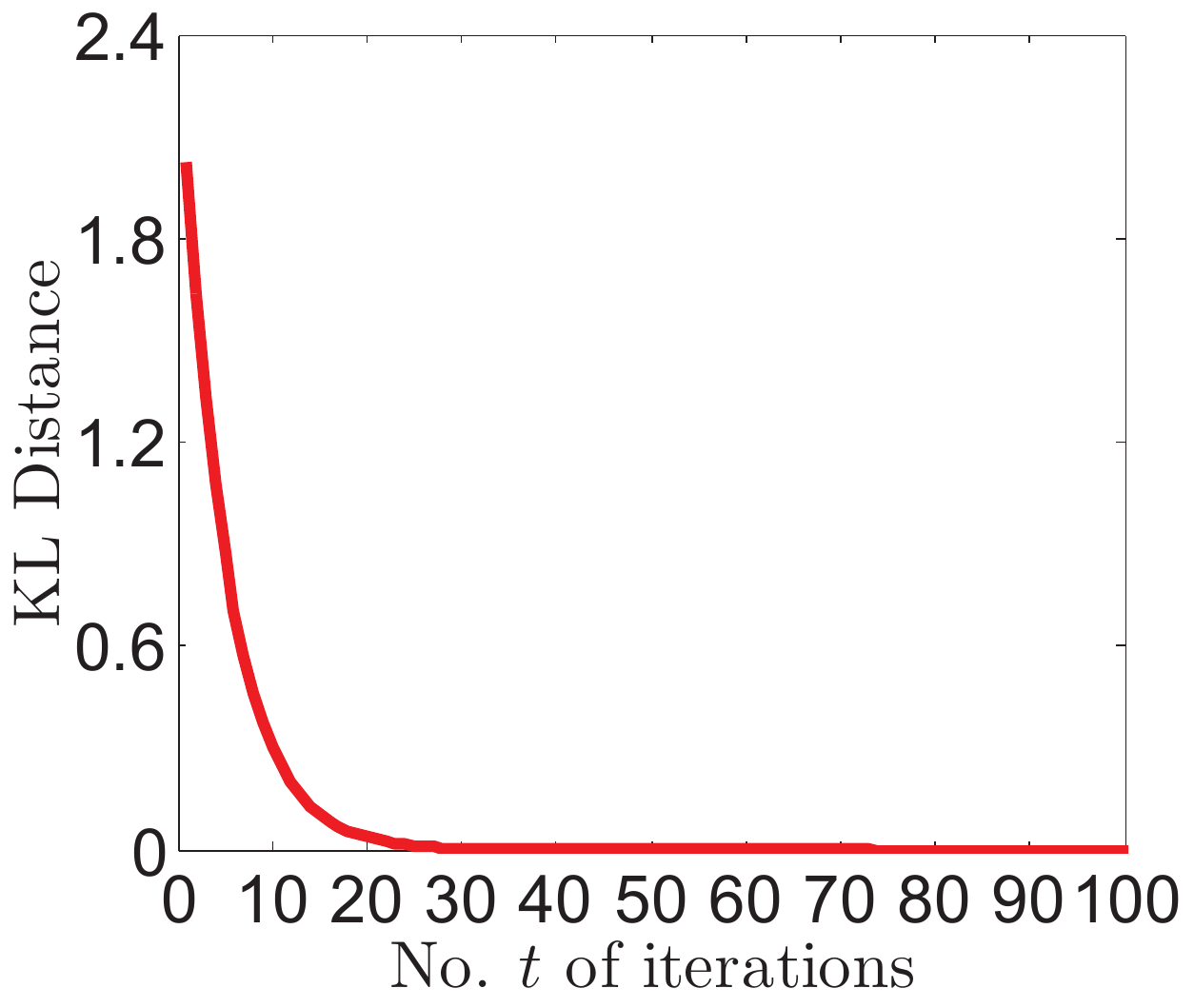} \vspace{-1mm}\\
				\hspace{-2.5mm}(d) & \hspace{-4mm}(e) & \hspace{-4mm}(f)
			\end{tabular}
			\caption{Graphs of KL distance $\mathrm{KL}(q(\mathbf{s}_\mathcal{I})||q^+(\mathbf{s}_\mathcal{I}))$ of (a) VBDTC$+$ to VBDTC, (b) VBFITC$+$ to VBFITC, (c) VBPIC$+$ to VBPIC, and $\mathrm{KL}((q(\mathbf{\Lambda},\sigma_f)||q^+(\mathbf{\Lambda},\sigma_f))$ of (d) VBDTC$+$ to VBDTC, (e) VBFITC$+$ to VBFITC, (f) VBPIC$+$ to VBPIC vs. no. $t$ of iterations for AIMPEAK dataset.}
			\label{fig1}
		\end{figure}
				
		Figs.~\ref{fig1}a-\ref{fig1}c (Figs.~\ref{fig1}d-\ref{fig1}f) shows results of the KL distance $\mathrm{KL}(q(\mathbf{s}_\mathcal{I})||q^+(\mathbf{s}_\mathcal{I}))$ ($\mathrm{KL}(q(\mathbf{\Lambda},\sigma_f)||q^+(\mathbf{\Lambda},\sigma_f))$) of $q^+(\mathbf{s}_\mathcal{I})$ to $q(\mathbf{s}_\mathcal{I})$ ($q^+(\mathbf{\Lambda},\sigma_f)$ to $q(\mathbf{\Lambda},\sigma_f)$) averaged over $5$ random selections of training data and mini-batch sequences with an increasing number $t$ of iterations. It can be observed that the variational distributions $q^+(\mathbf{s}_\mathcal{I})$ and $q^+(\mathbf{\Lambda},\sigma_f)$ induced by VBDTC$+$, VBFITC$+$, and VBPIC$+$ converge rapidly to, respectively, $q(\mathbf{s}_\mathcal{I})$ and $q(\mathbf{\Lambda},\sigma_f)$ induced by VBDTC, VBFITC, and VBPIC, thus corroborating our theoretical results in Section~\ref{Stochastic Variational Inference for GPR}. From Figs.~\ref{fig1}a-\ref{fig1}c, it can also be observed that $q^+(\mathbf{s}_\mathcal{I})$ induced by  VBDTC$+$ converges faster to $q(\mathbf{s}_\mathcal{I})$ than that by VBFITC$+$ and VBPIC$+$, which can be explained by its much simpler noise structure by assuming i.i.d. observation noises with constant variance $ \sigma_n^2 $.
\subsection{Empirical Evaluation on AIRLINE and TWITTER Datasets}
The TWITTER dataset 
contains $583250$ instances of buzz events on Twitter.  The input denotes a relatively large $77$D feature vector described at 
http://ama.liglab.fr/datasets/buzz/, which makes this dataset suitable for evaluating  robustness to overfitting.
The output is the popularity of the instance's topic. 		
%
The massive benchmark AIRLINE dataset~\cite{Lawrence13} contains $2055733$ records of information about every commercial flight in the USA from January to April $2008$. The input denotes an $8$D feature vector of age of the aircraft (no. of years in service), travel distance (km), airtime, departure and arrival time (min.) as well as day of the week, day of month, and month. The output is the delay time (min.) of the flight. 
For each dataset, $5\%$ is randomly selected and set aside as test data. The remaining data is used as training data and \textcolor{black}{partitioned into $B=1000$ mini-batches} using $k$-means (i.e., $k = B$). We randomly select $100$ inducing inputs from the inputs of the training data. 

\begin{figure}
		\begin{tabular}{cc}
			\hspace{-2mm}\includegraphics[scale=0.16]{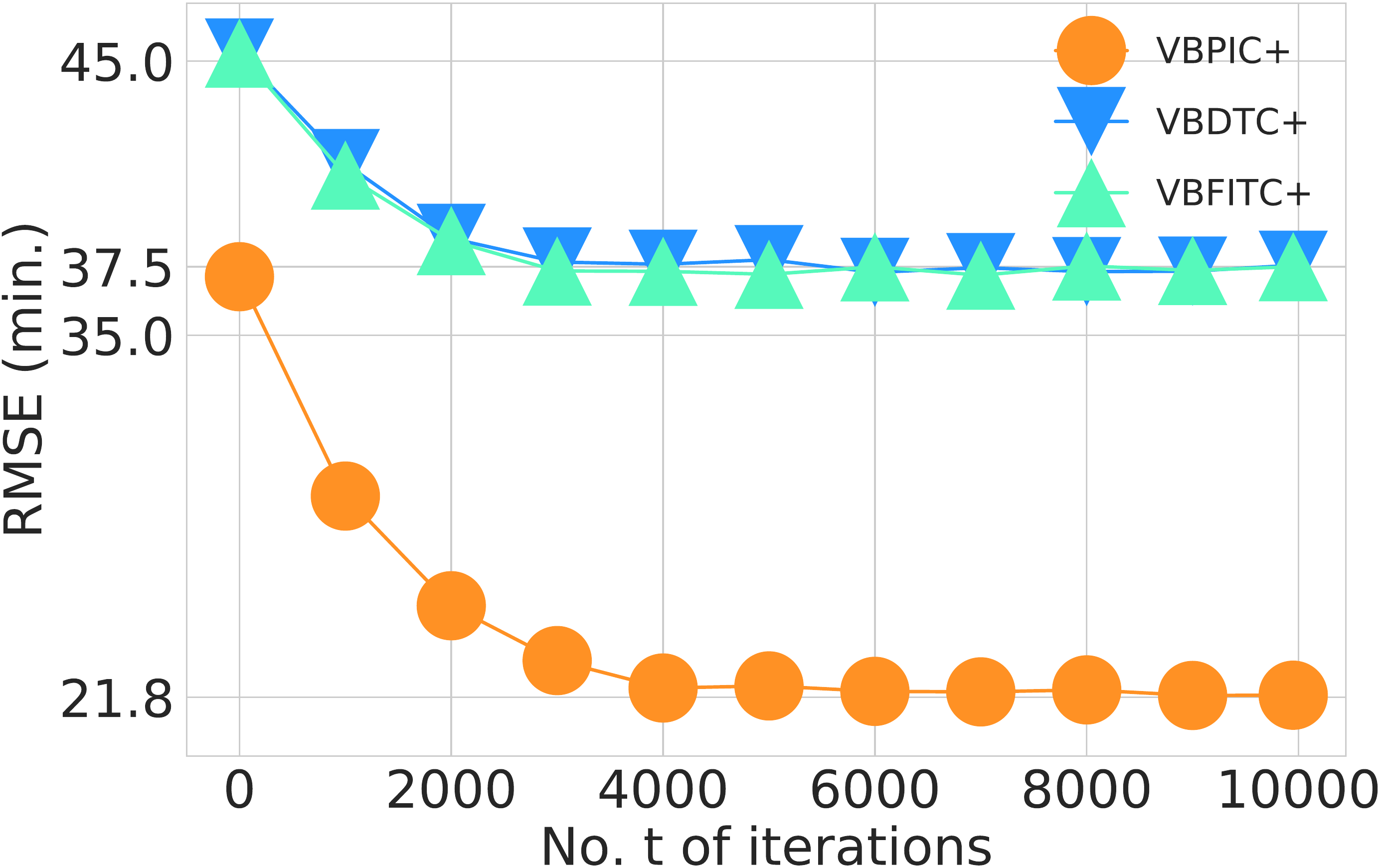} &
			\hspace{-4mm}\includegraphics[scale=0.16]{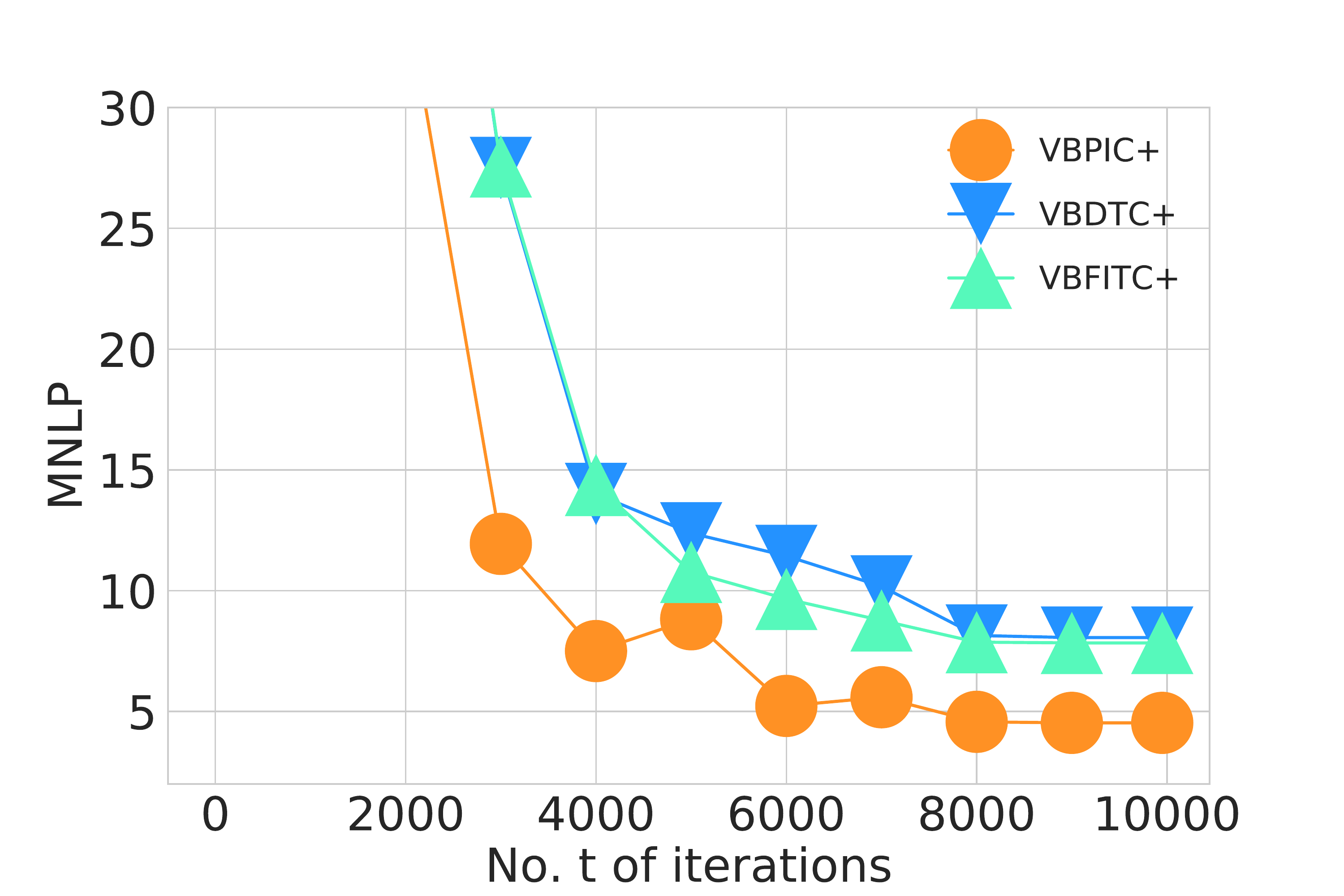} 
			\vspace{-1mm}\\
			\hspace{-2mm}{(a)} & \hspace{-4mm}{(b)}
			\\
			\hspace{-2mm}\includegraphics[scale=0.16]{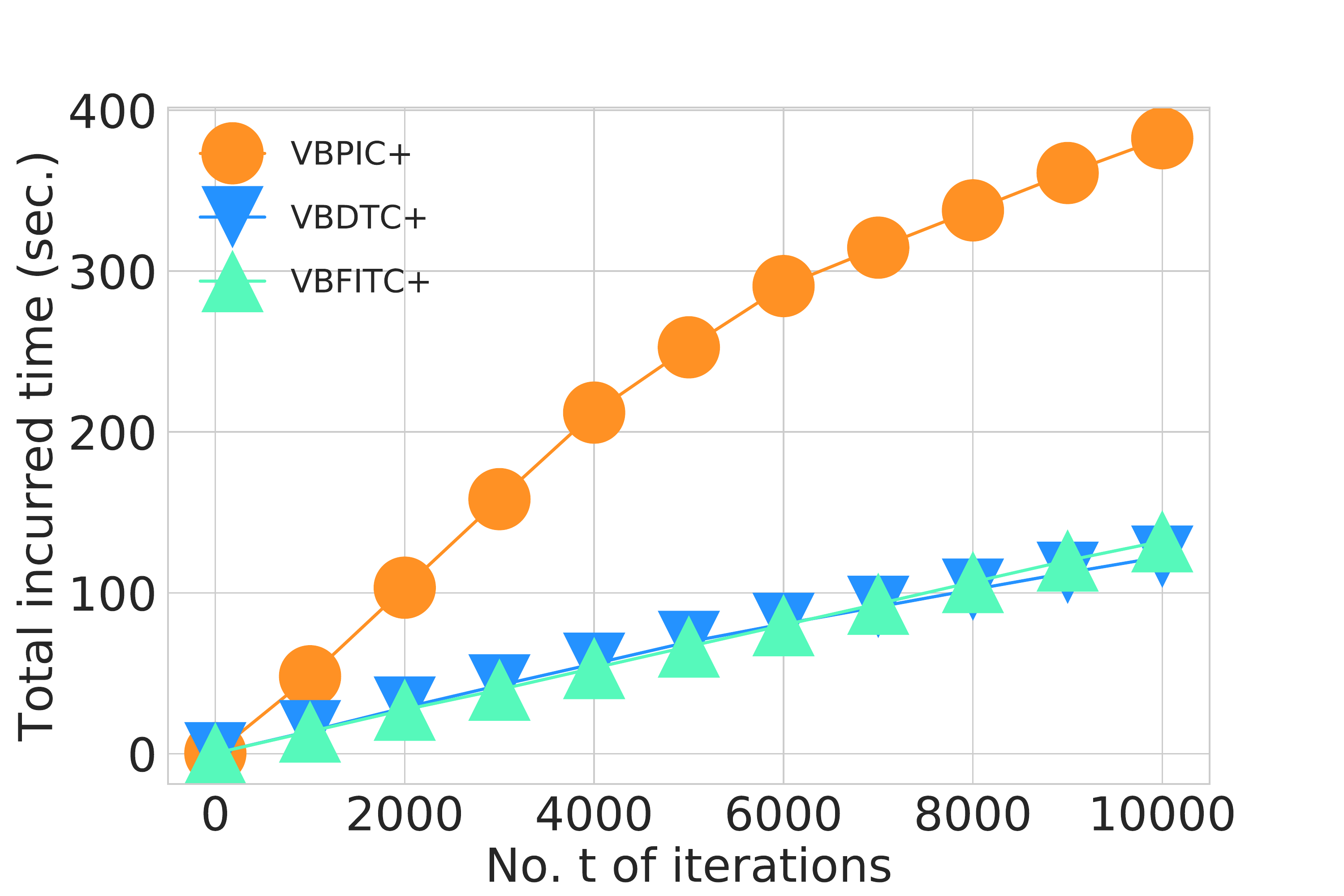} &
			\hspace{-4mm}\includegraphics[scale=0.16]{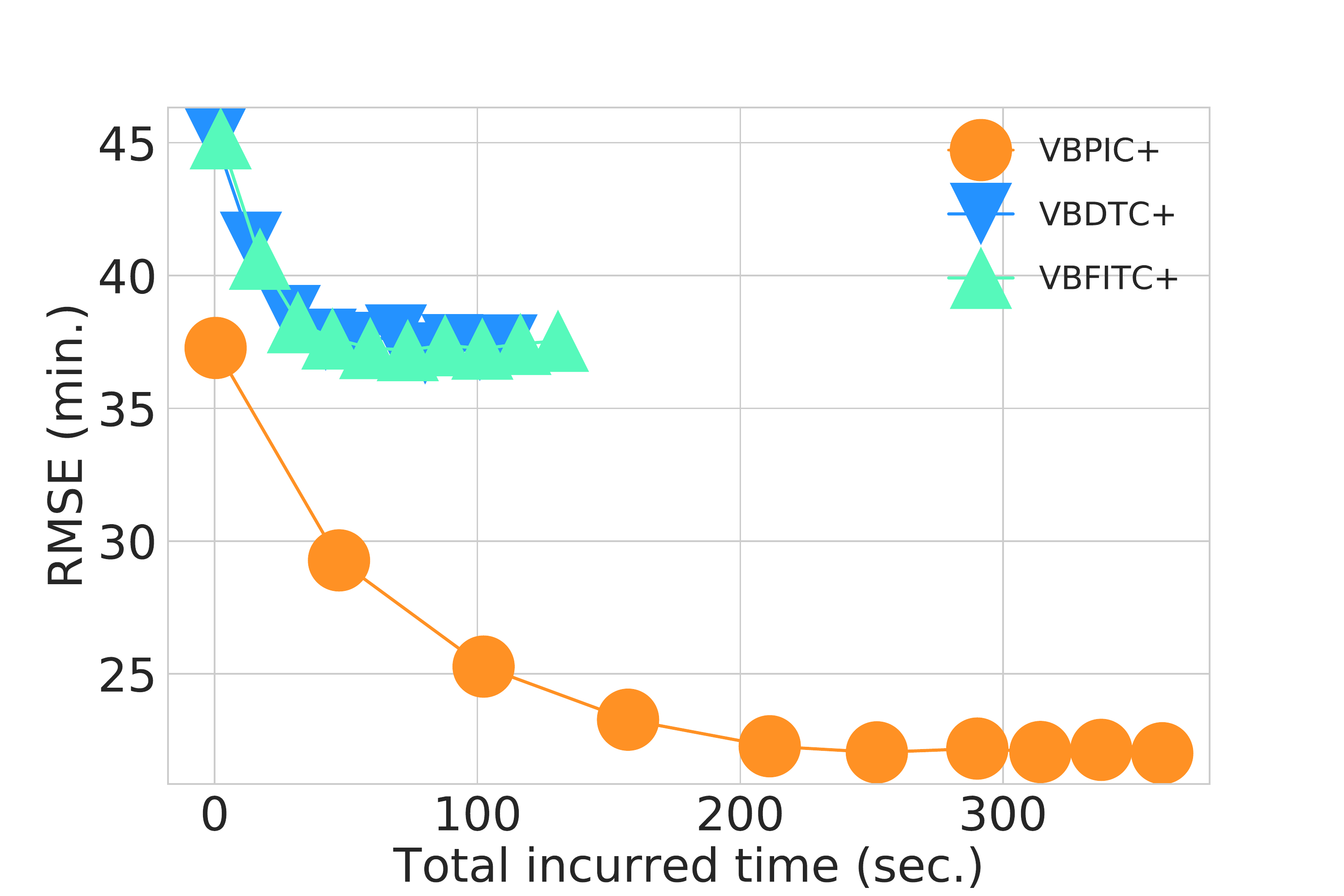}
			\vspace{-1mm}\\
			\hspace{-2mm}{(c)} & \hspace{-4mm}{(d)}\vspace{-0.7mm}
		\end{tabular}
		\caption{Graphs of (a) RMSE, (b) MNLP, and (c) total incurred time vs. number $t$ of iterations, and (d) graphs of RMSE vs. total incurred time of VBDTC$+$, VBFITC$+$, and VBPIC$+$ for the AIRLINE dataset.}
		\label{fig4}
	\end{figure}

	Figs.~\ref{fig4}a and~\ref{fig4}b show results of RMSE and MNLP achieved by the stochastic variants of our VBSGPR models averaged over $5$ random selections of $5\%$ test data and mini-batch sequences with an increasing number $t$ of iterations for the AIRLINE dataset.
It can be observed that VBPIC$+$ (RMSE of $21.87$~min. and MNLP of $4.53$) achieves considerably better predictive performance than VBFITC$+$ (RMSE of $37.05$~min. and \textcolor{black}{MNLP of $7.84$}) and VBDTC$+$ (RMSE of $37.55$~min. and \textcolor{black}{MNLP of $8.06$}). 
		To explain this, VBFITC$+$ and VBDTC$+$ have both imposed a strong assumption of independently distributed observation noises. 
		In contrast, VBPIC$+$ caters to correlation of observation noises within each mini-batch of data (Sections~\ref{Variational Inference of the Bayesian DTC} and~\ref{Stochastic Variational Inference for GPR}), hence modeling and predicting real-world datasets with correlated noises better.
		Furthermore, unlike VBFITC$+$ and VBDTC$+$, VBPIC$+$ does not assume conditional independence between the training and test outputs given the inducing outputs in its test conditional.

Fig.~\ref{fig4}c exhibits a near-linear increase in total incurred time with an increasing number $t$ of iterations for VBDTC$+$, VBFITC$+$, and VBPIC$+$. Our experiments reveal that VBDTC$+$, VBFITC$+$, and VBPIC$+$ incur, respectively, an average of $0.0122$, $0.0132$, and $0.038$ seconds per iteration of stochastic gradient ascent update. 
Fig.~\ref{fig4}d shows that VBPIC$+$ can achieve a more superior trade-off between predictive performance vs. time efficiency than VBDTC$+$ and VBFITC$+$. 
%

\begin{figure}
		\begin{tabular}{cc}
			\hspace{-2mm}\includegraphics[scale=0.16]{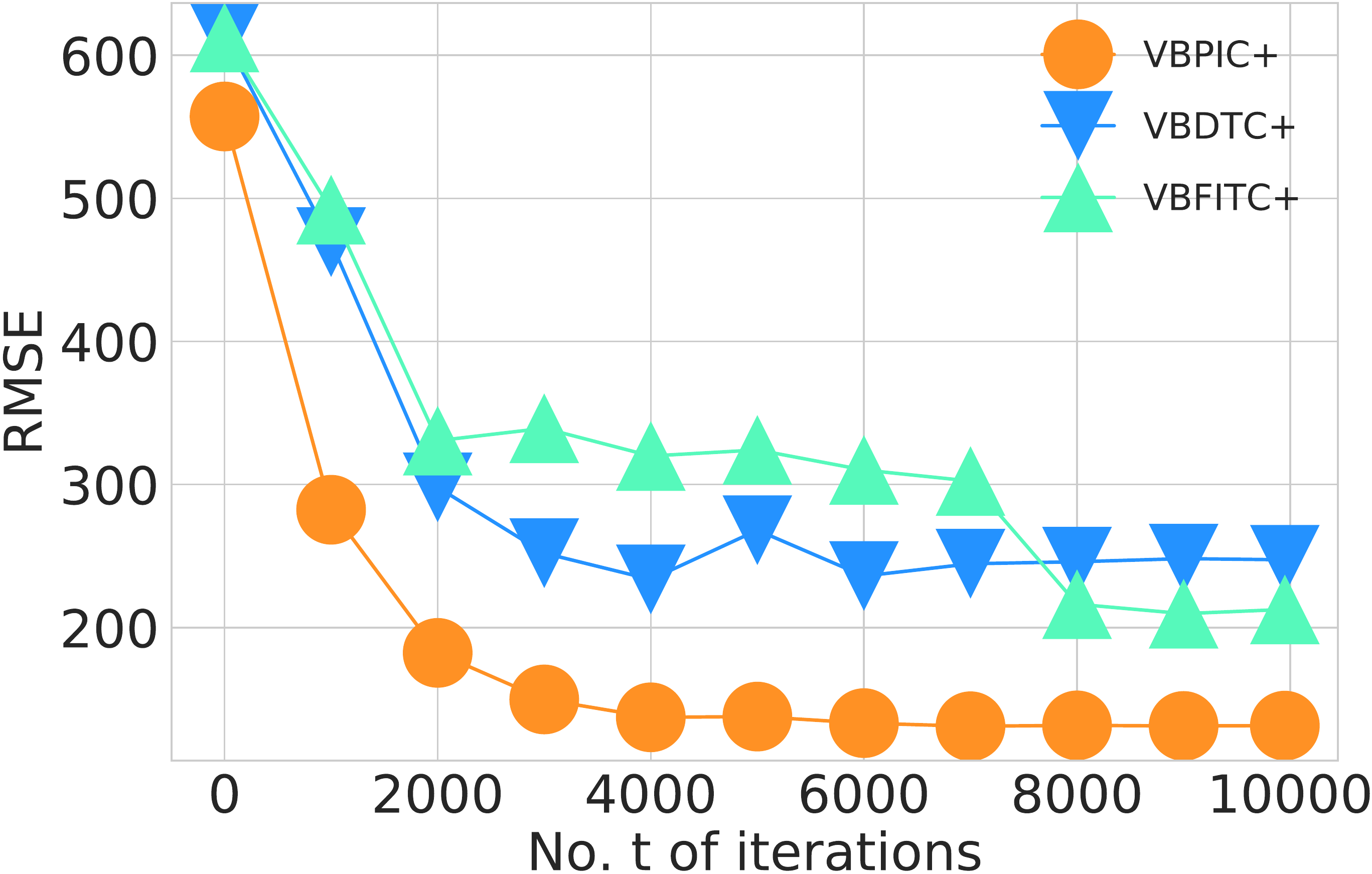} &
			\hspace{-4mm}\includegraphics[scale=0.16]{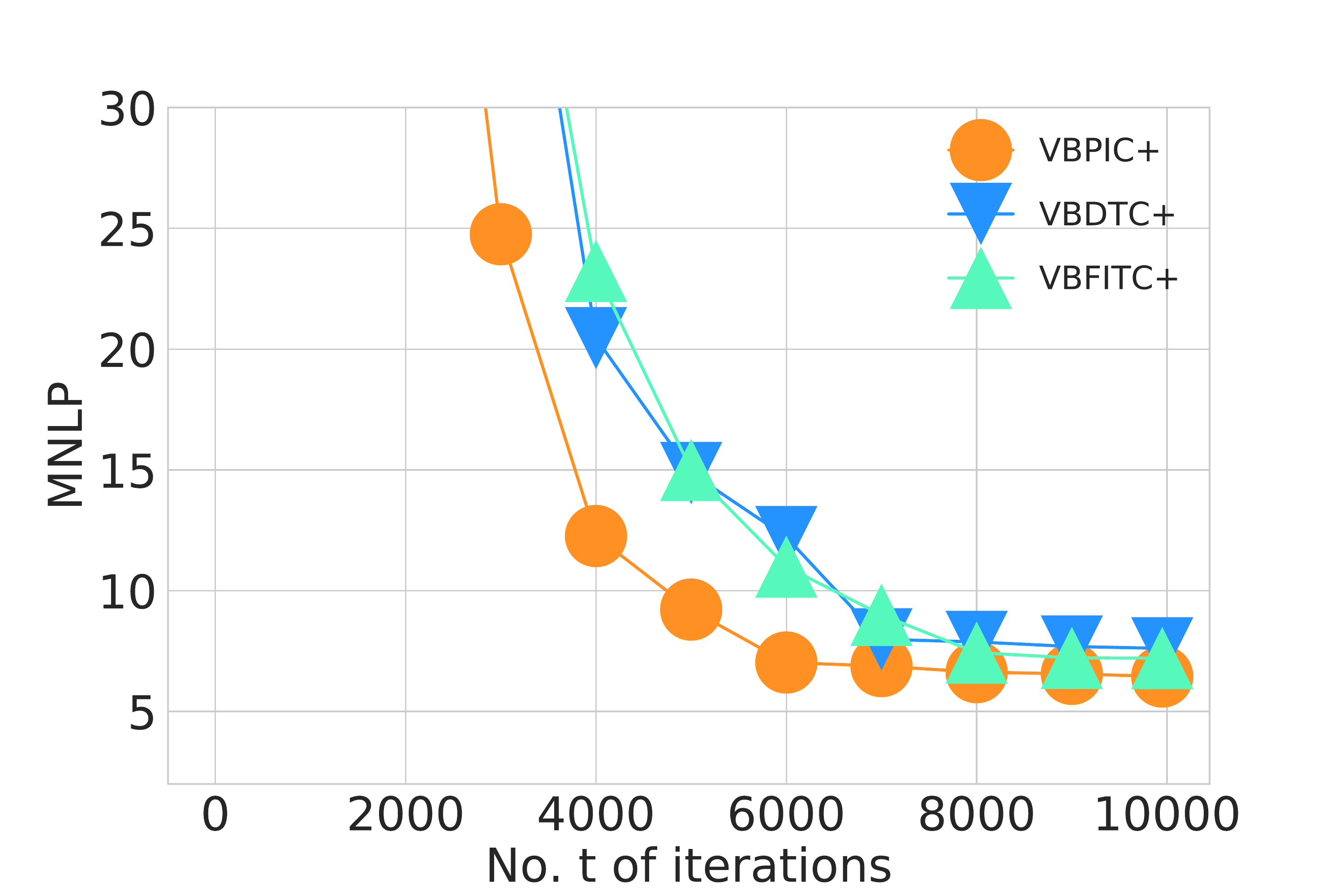} 
			\vspace{-1mm}\\
			\hspace{-2mm}{(a)} & \hspace{-4mm}{(b)}
			\\
			\hspace{-2mm}\includegraphics[scale=0.16]{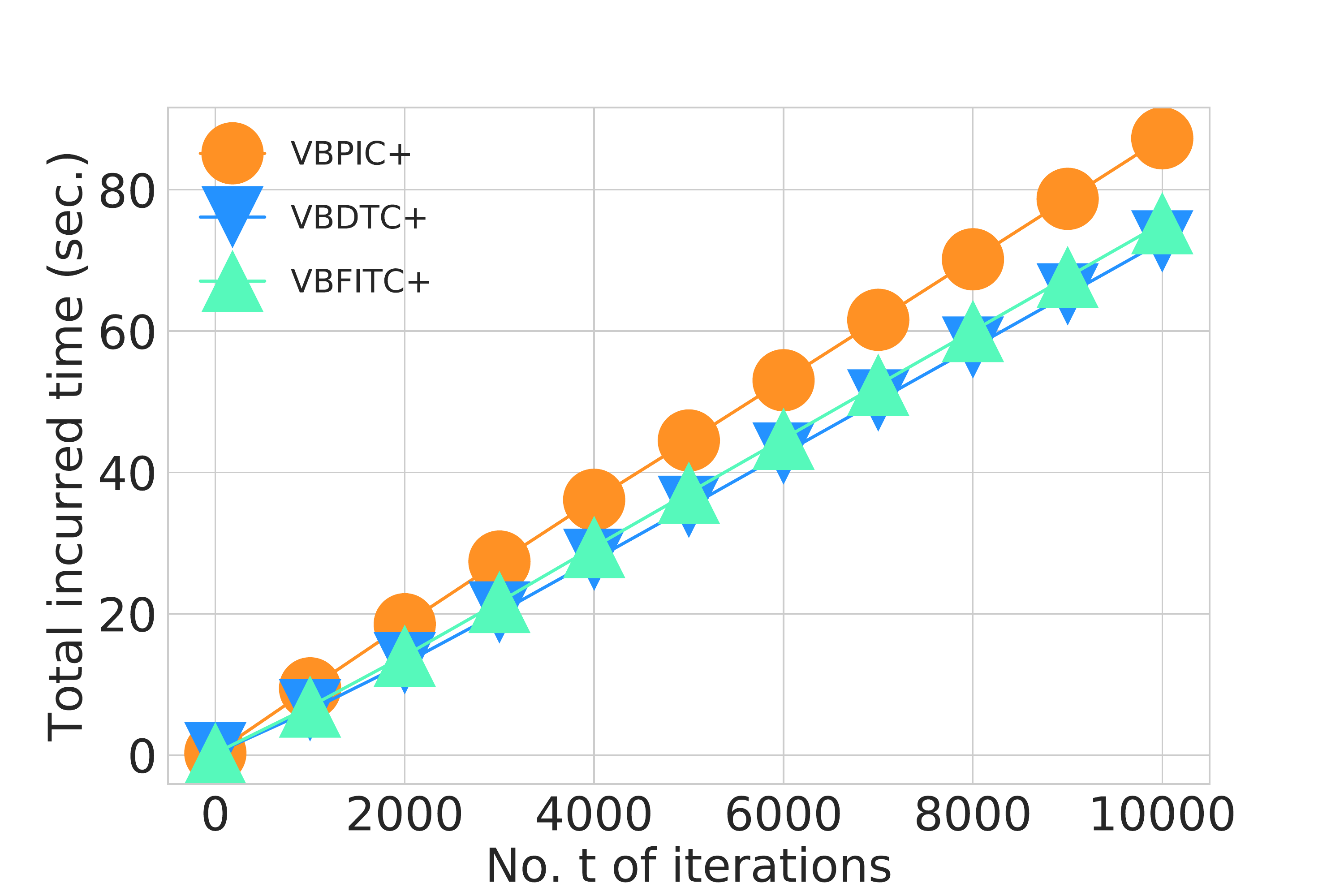} &
			\hspace{-4mm}\includegraphics[scale=0.16]{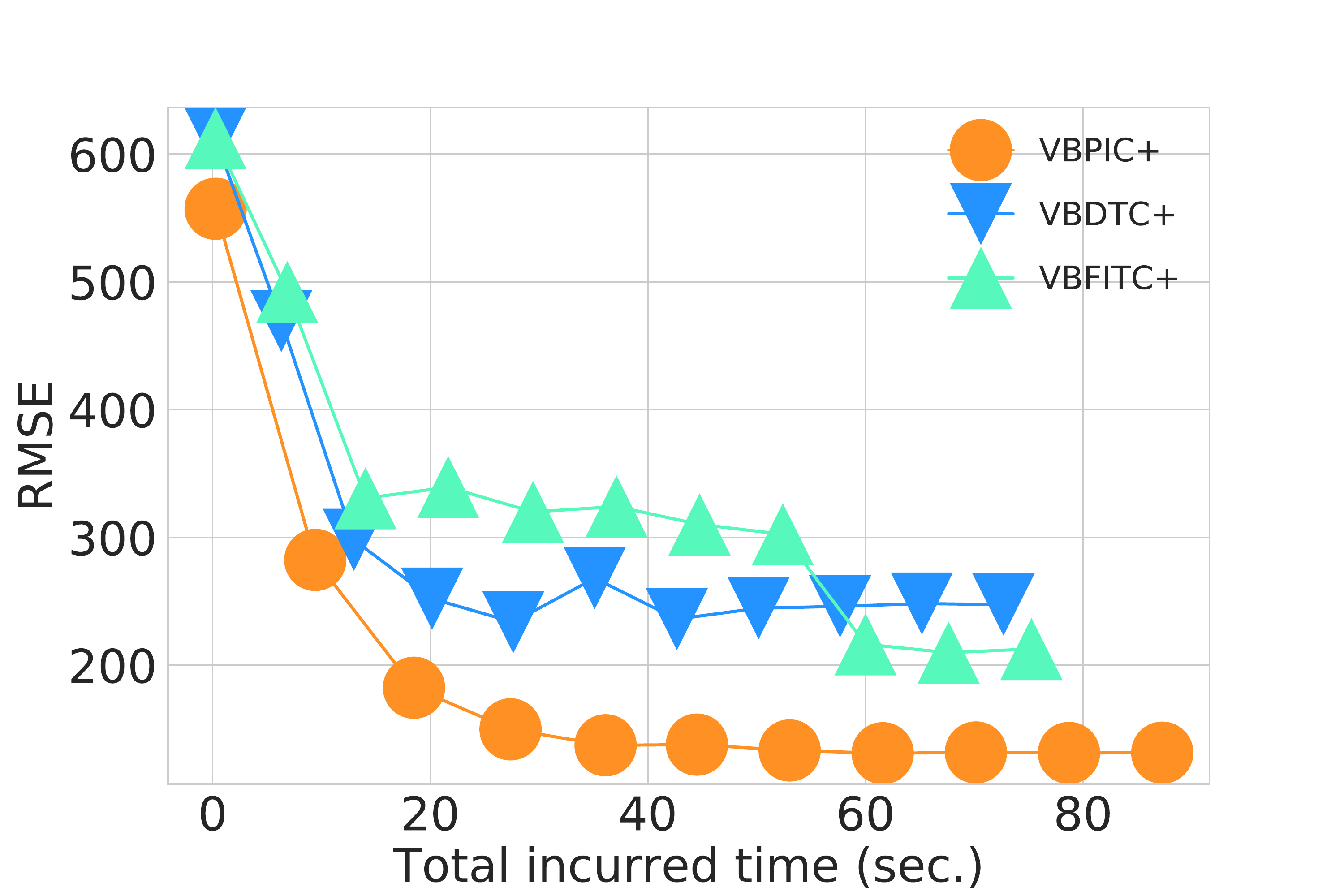}
			\vspace{-1mm}\\
			\hspace{-2mm}{(c)} & \hspace{-4mm}{(d)}\vspace{-1mm}
		\end{tabular}
		\caption{Graphs of (a) RMSE, (b) MNLP, and (c) total incurred time vs. number $t$ of iterations, and (d) graphs of RMSE vs. total incurred time of VBDTC$+$, VBFITC$+$, and VBPIC$+$ for the TWITTER dataset.}
		\label{fig3}
	\end{figure}

		Figs.~\ref{fig3}a and~\ref{fig3}b show results of RMSE and MNLP achieved by the stochastic variants of our VBSGPR models averaged over $5$ random selections of $5\%$ test data and mini-batch sequences with an increasing number $t$ of iterations for the TWITTER dataset. 
		The observations are similar to that for the AIRLINE dataset:
		It can be observed that \textcolor{black}{VBPIC$+$ (RMSE of $131.46$ and MNLP of $6.45$) achieves significantly better predictive performance than VBFITC$+$ (RMSE of $212.67$ and MNLP of $7.21$) and VBDTC$+$ (RMSE of $247.38$ and MNLP of $7.69$)}; 		
this can be explained by the same reasons as that discussed previously for the AIRLINE dataset.
		
		Fig.~\ref{fig3}c also exhibits a linear increase in total incurred time with an increasing number $t$ of iterations for VBDTC$+$, VBFITC$+$, and VBPIC$+$. Our experiments reveal that VBDTC$+$, VBFITC$+$, and VBPIC$+$ incur, respectively, \textcolor{black}{an average of $0.0073$, $0.0075$, and $0.0087$ seconds per iteration of stochastic gradient ascent update, which are shorter than that for the AIRLINE dataset} 
due to a smaller mini-batch size.		
Fig.~\ref{fig3}d reveals that VBPIC$+$ can similarly achieve the best trade-off between predictive performance vs. time efficiency.
		
Table~\ref{tab:exp} compares the predictive performance (RMSEs) achieved by state-of-the-art GP models for the AIRLINE and TWITTER datasets. 
It can be observed that our VBPIC$+$ significantly outperforms state-of-the-art SVIGP, Dist-VGP, rBCM, and PIC$+$, which find point estimates of hyperparameters, and VSSGPR due to its restrictive assumption, as discussed in Section~\ref{sect:intro}. In contrast, our VBPIC$+$ assumes a variational Bayesian treatment of its hyperparameters, thus achieving robustness to overfitting due to Bayesian model selection, as demonstrated later. Unlike VSSGPR, VBPIC$+$ does not assume conditional independence between the training and test outputs in its test conditional.
\begin{table}
\begin{small}	
\begin{tabular}{l|cccccc}
\hline
\hspace{-2.7mm} Dataset & \hspace{-2mm} SVIGP & \hspace{-3mm} Dist-VGP & \hspace{-3mm} rBCM & \hspace{-3mm} PIC$+$ & \hspace{-3.5mm} VBPIC$+$ & \hspace{-3.5mm} VSSGPR \hspace{-2.7mm} \\ 
\hline
\hspace{-2mm}AIRLINE & \hspace{-2mm} $39.53$ & \hspace{-3mm} $35.30$ & \hspace{-3.5mm} $34.40$ & \hspace{-3.5mm} $24.9$ & \hspace{-4mm} {\bf 21.87} & \hspace{-3.5mm} $38.95$\\ 
\hspace{-2mm}TWITTER\hspace{-2mm} & \hspace{-2mm} $-$ & \hspace{-3mm} $-$ & \hspace{-3.5mm} $-$ & \hspace{-3.5mm} $190.2$ & \hspace{-4mm} {\bf 131.4} & \hspace{-3.5mm} $585.9$\\
\hline
\end{tabular}
\end{small}
\caption{RMSE achieved by VBPIC$+$ and state-of-the-art GP models for AIRLINE and TWITTER datasets. The results of PIC$+$ and VSSGPR are obtained using their GitHub codes. The results of Dist-VGP and rBCM are taken from their respective papers and that of SVIGP is reported in~\cite{NghiaICML15}. They are all based on the same settings of training/test data sizes $= 2$M/$100$K ($554$K/$29$K) for the AIRLINE (TWITTER) dataset.}
\label{tab:exp}\vspace{-1mm}
\end{table}
		\begin{figure}
			\begin{tabular}{cc}
				\hspace{-2.5mm}\includegraphics[scale=0.16]{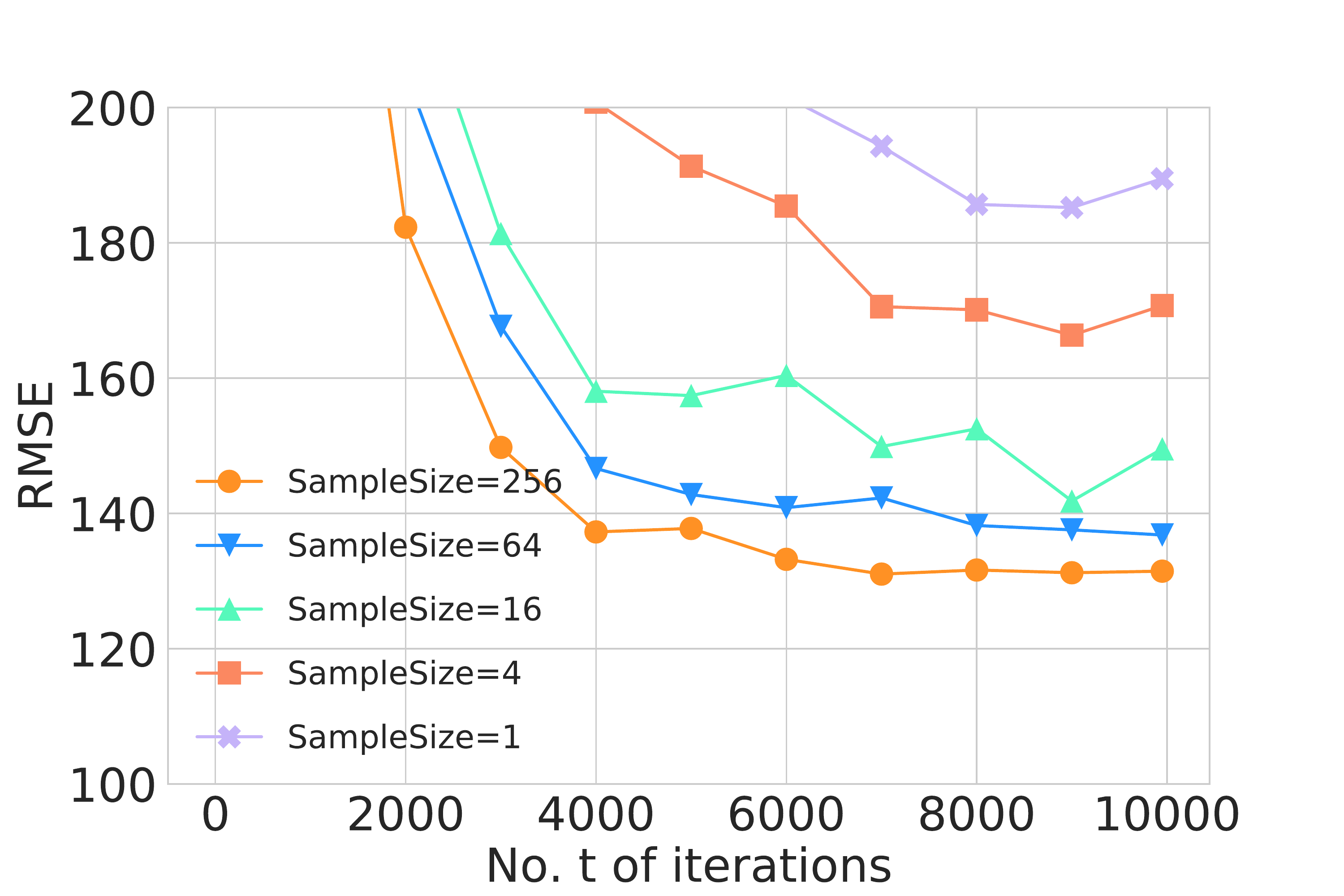} &
				\hspace{-4mm}\includegraphics[scale=0.16]{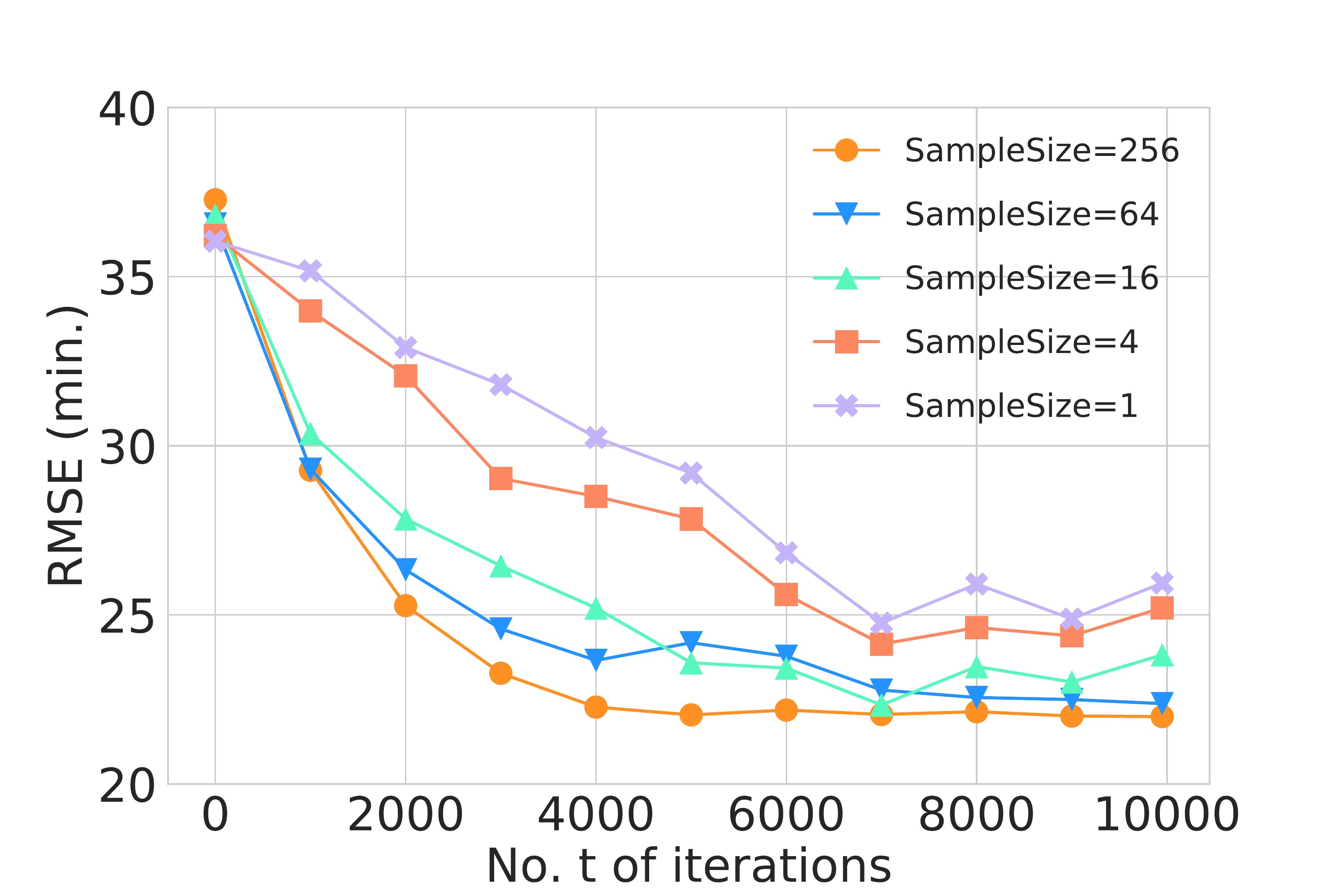}\vspace{-1mm}\\
				\hspace{-2.5mm}{(a)} & \hspace{-4mm}{(b)}\vspace{-1mm}
			\end{tabular}
			\caption{Graphs of RMSEs of VBPIC$+$ vs. number $t$ of iterations with varying sampling sizes for computing its predictive mean for the (a) TWITTER and (b) AIRLINE datasets.}
			\label{fig5}
		\end{figure}

Fig.~\ref{fig5} shows results of RMSEs achieved by our VBPIC$+$ with an increasing number $t$ of iterations and varying sample sizes for computing its predictive mean (Section~\ref{predict}).
Note that a sample size of $1$ reduces VBPIC to PIC that treats its sampled hyperparameters as a point estimate. By increasing the sample size, it can be observed that VBPIC$+$ converges faster to a lower RMSE using less iterations
due to its Bayesian model selection/averaging, thus demonstrating its increasing robustness to overfitting.

\begin{figure}
			\begin{tabular}{cc}
				\hspace{-2mm}\includegraphics[scale=0.16]{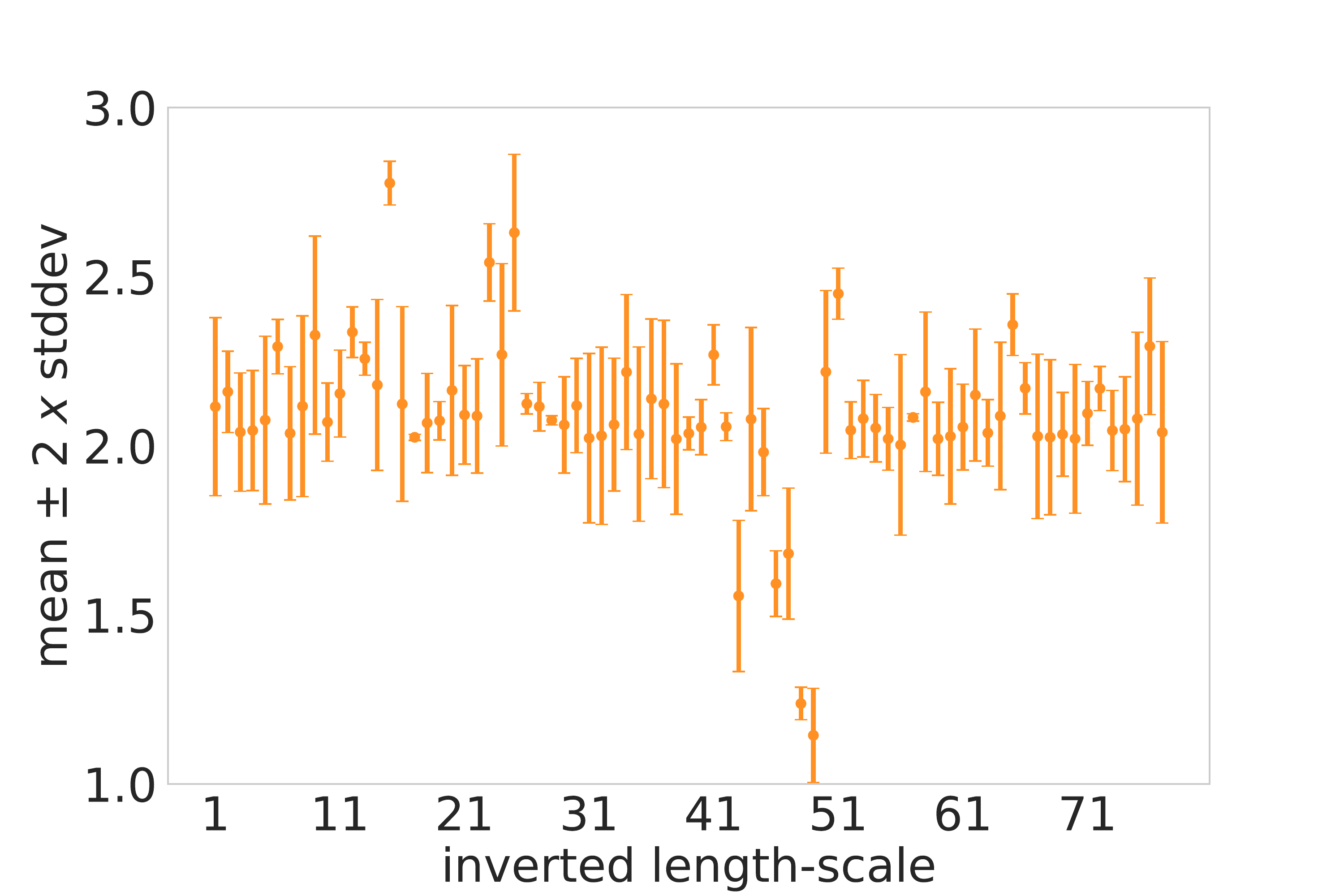} &
				\hspace{-3mm}\includegraphics[scale=0.16]{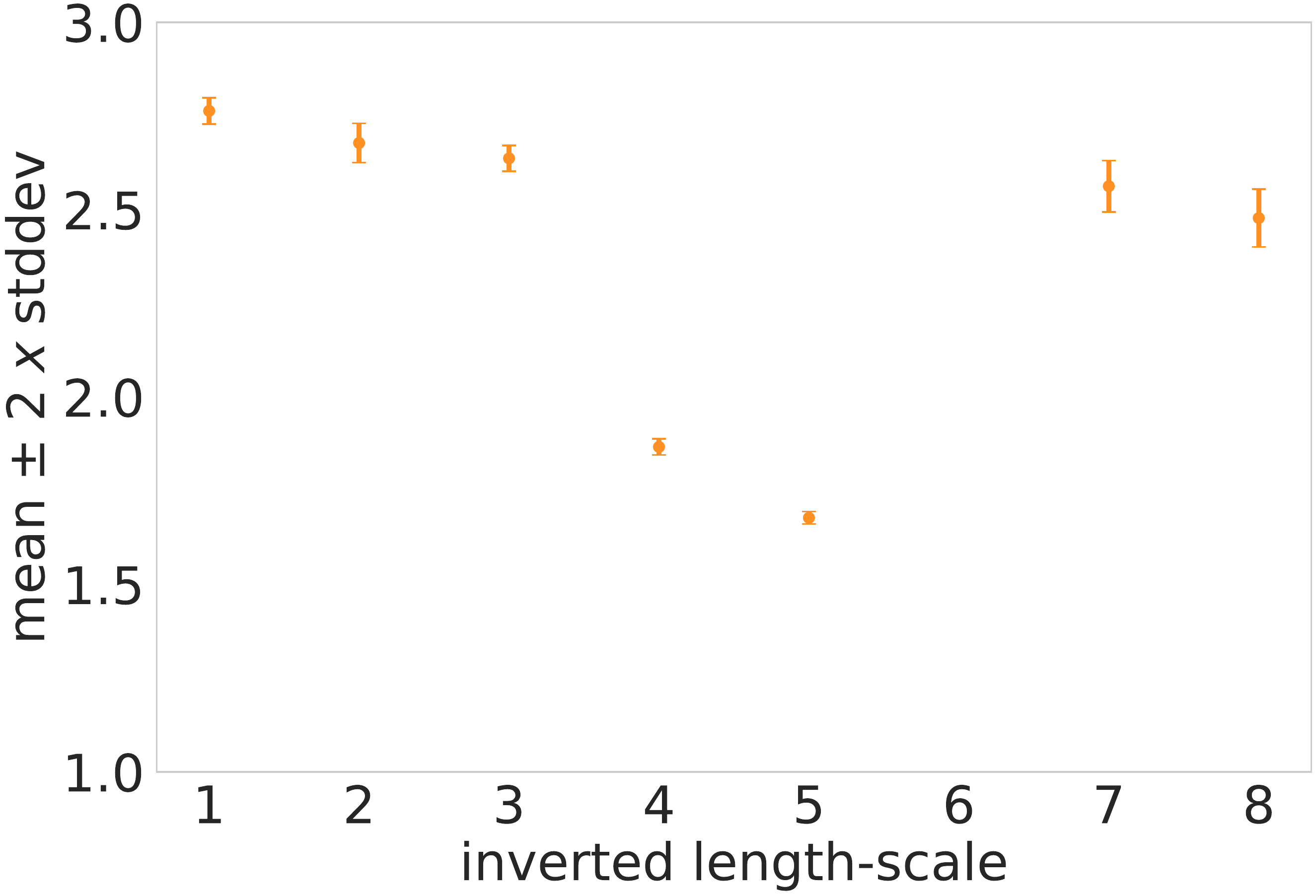}\vspace{-1mm}\\
				\hspace{-2mm}{(a)} & \hspace{-3mm}{(b)}\vspace{-2mm}
			\end{tabular}
			\caption{$95\%$ confidence intervals (mean $\nu^+_i$ $\pm$ 2$\ \times\ $standard deviation $(\xi^+_i)^{1/2}$) for \textcolor{black}{inverted length-scale} hyperparameters $\lambda_i$ for $i=1,\ldots,d$ \textcolor{black}{after $t=10000$ iterations} for the (a) TWITTER ($d=77$ normalized input dimensions) and (b) AIRLINE ($d=8$ normalized input dimensions) datasets.}
			\label{fig6}
		\end{figure}

Fig.~\ref{fig6} displays the $95\%$ confidence intervals (mean $\nu^+_i$ $\pm$ 2$\ \times\ $standard deviation $(\xi^+_i)^{1/2}$) for \textcolor{black}{inverted length-scale} hyperparameters $\lambda_i$ for $i=1,\ldots,d$ \textcolor{black}{after $t=10000$ iterations} for the TWITTER ($d=77$ normalized input dimensions) and AIRLINE ($d=8$ normalized input dimensions) datasets. It can be observed that the confidence intervals  are generally wider (i.e., larger uncertainty of $\lambda_1,\ldots,\lambda_d$) for the TWITTER dataset than for the AIRLINE dataset. To confirm this, we measure the \emph{mean log variance} (MLV) $\sum^d_{i=1}\log\xi^+_i/d$ of $\lambda_1,\ldots,\lambda_d$ and notice that the TWITTER dataset gives a higher MLV of $-4.09$ than that for the AIRLINE dataset (i.e., MLV $= -6.55$).
So, with a larger uncertainty of $\lambda_1,\ldots,\lambda_d$, their point estimates have a greater tendency to overfit and hence yield a poorer predictive performance, as observed in Fig.~\ref{fig5} (compare the performance gap between sample sizes of $1$ vs. $256$).
	%
	%
	%
	\section{Conclusion}
	This paper describes a novel variational inference framework for a family of VBSGPR models (e.g., VBDTC, VBFITC, VBPIC) whose approximations are variationally optimal with respect to the FGPR model enriched with various corresponding correlation structures of the observation noises.
	Our variational Bayesian treatment of hyperparameters enables our VBSGPR models to mitigate critical issues (e.g., overfitting) which plague existing variational SGPR models that optimize point estimates of hyperparameters (Section~\ref{sect:intro}).	
The stochastic variants of our VBSGPR models can yield good predictive performance fast and improve their predictive performance over time, thus achieving scalability to big data.
	Empirical evaluation on two real-world datasets reveals that the stochastic variant of our VBPIC can significantly outperform existing state-of-the-art GP models, thus demonstrating its robustness to overfitting due to Bayesian model selection while preserving scalability to big data through stochastic optimization.	
	For our future work, we plan to integrate our proposed framework with that of decentralized/distributed data/model fusion~\cite{chen2013gaussian,LowTASE15,LowUAI12,NghiaAAAI19,Ruofei18} for collective online learning of a massive number of VBSGPR models. 
\bibliographystyle{IEEEtran}
\bibliography{Reference}

\begin{thebibliography}{10}
\providecommand{\url}[1]{#1}
\csname url@samestyle\endcsname
\providecommand{\newblock}{\relax}
\providecommand{\bibinfo}[2]{#2}
\providecommand{\BIBentrySTDinterwordspacing}{\spaceskip=0pt\relax}
\providecommand{\BIBentryALTinterwordstretchfactor}{4}
\providecommand{\BIBentryALTinterwordspacing}{\spaceskip=\fontdimen2\font plus
\BIBentryALTinterwordstretchfactor\fontdimen3\font minus
  \fontdimen4\font\relax}
\providecommand{\BIBforeignlanguage}[2]{{%
\expandafter\ifx\csname l@#1\endcsname\relax
\typeout{** WARNING: IEEEtran.bst: No hyphenation pattern has been}%
\typeout{** loaded for the language `#1'. Using the pattern for}%
\typeout{** the default language instead.}%
\else
\language=\csname l@#1\endcsname
\fi
#2}}
\providecommand{\BIBdecl}{\relax}
\BIBdecl

\bibitem{candela10}
M.~{L{\'{a}}zaro-Gredilla}, J.~{Qui{\~{n}}onero-Candela}, C.~E. Rasmussen, and
  A.~R. {Figueiras-Vidal}, ``Sparse spectrum {Gaussian} process regression,''
  \emph{JMLR}, vol.~11, pp. 1865--1881, 2010.

\bibitem{candela05}
J.~{Qui{\~{n}}onero-Candela} and C.~E. Rasmussen, ``A unifying view of sparse
  approximate {Gaussian} process regression,'' \emph{JMLR}, vol.~6, pp.
  1939--1959, 2005.

\bibitem{LowUAI13}
J.~Chen, N.~Cao, K.~H. Low, R.~Ouyang, C.~K.-Y. Tan, and P.~Jaillet, ``Parallel
  {Gaussian} process regression with low-rank covariance matrix
  approximations,'' in \emph{Proc. {UAI}}, 2013, pp. 152--161.

\bibitem{LowAAAI15}
K.~H. Low, J.~Yu, J.~Chen, and P.~Jaillet, ``Parallel {Gaussian} process
  regression for big data: Low-rank representation meets {M}arkov
  approximation,'' in \emph{Proc. {AAAI}}, 2015, pp. 2821--2827.

\bibitem{LowDyDESS15}
K.~H. Low, J.~Chen, T.~N. Hoang, N.~Xu, and P.~Jaillet, ``Recent advances in
  scaling up {Gaussian} process predictive models for large spatiotemporal
  data,'' in \emph{Proc. {DyDESS}}, S.~Ravela and A.~Sandu, Eds.\hskip 1em plus
  0.5em minus 0.4em\relax LNCS 8964, Springer, 2015, pp. 167--181.

\bibitem{Csato02}
L.~Csat\'{o} and M.~Opper, ``Sparse online {Gaussian} processes,'' \emph{Neural
  Comput.}, vol.~14, pp. 641--669, 2002.

\bibitem{LowAAAI14}
N.~Xu, K.~H. Low, J.~Chen, K.~K. Lim, and E.~B. {\"{O}zg\"{u}l},
  ``{GP-Localize}: Persistent mobile robot localization using online sparse
  {Gaussian} process observation model,'' in \emph{Proc. {AAAI}}, 2014, pp.
  2585--2592.

\bibitem{Titsias09a}
M.~K. Titsias, ``Variational model selection for sparse {Gaussian} process
  regression,'' School of Computer Science, University of Manchester, Tech.
  Rep., 2009.

\bibitem{Titsias09}
------, ``Variational learning of inducing variables in sparse {Gaussian}
  processes,'' in \emph{Proc. {AISTATS}}, 2009, pp. 567--574.

\bibitem{Seeger03}
M.~Seeger, C.~Williams, and N.~D. Lawrence, ``Fast forward selection to speed
  up sparse {G}aussian process regression,'' in \emph{Proc. {AISTATS}}, 2003.

\bibitem{Yarin14}
Y.~Gal, M.~{van der Wilk}, and C.~E. Rasmussen, ``Distributed variational
  inference in sparse {Gaussian} process regression and latent variable
  models,'' in \emph{Proc. NIPS}, 2014, pp. 3257--3265.

\bibitem{Lawrence13}
J.~Hensman, N.~Fusi, and N.~Lawrence, ``{Gaussian} processes for big data,'' in
  \emph{Proc. {UAI}}, 2013, pp. 282--290.

\bibitem{cheng2016incremental}
C.-A. Cheng and B.~Boots, ``Incremental variational sparse {Gaussian} process
  regression,'' in \emph{Proc. NIPS}, 2016, pp. 4410--4418.

\bibitem{NghiaICML15}
T.~N. Hoang, Q.~M. Hoang, and K.~H. Low, ``A unifying framework of anytime
  sparse {Gaussian} process regression models with stochastic variational
  inference for big data,'' in \emph{Proc. {ICML}}, 2015, pp. 569--578.

\bibitem{HoangICML16}
------, ``A distributed variational inference framework for unifying parallel
  sparse {Gaussian} process regression models,'' in \emph{Proc. {ICML}}, 2016.

\bibitem{bui2017streaming}
T.~D. Bui, C.~Nguyen, and R.~E. Turner, ``Streaming sparse {Gaussian} process
  approximations,'' in \emph{Proc. NIPS}, 2017, pp. 3301--3309.

\bibitem{Snelson06}
E.~L. Snelson and Z.~Gharahmani, ``Sparse {Gaussian} processes using
  pseudo-inputs,'' in \emph{Proc. NIPS}, 2005, pp. 1257--1264.

\bibitem{Jordan08}
M.~J. Wainwright and M.~I. Jordan, ``Graphical models, exponential families,
  and variational inference,'' \emph{Foundations and Trends$^{\textregistered}$
  in Machine Learning}, vol.~1, no. 1--2, pp. 1--305, 2008.

\bibitem{Titsias13}
M.~K. Titsias and M.~{L{\'{a}}zaro-Gredilla}, ``Variational inference for
  {Mahalanobis} distance metrics in {Gaussian} process regression,'' in
  \emph{Proc. NIPS}, 2013, pp. 279--287.

\bibitem{LowAAMAS13}
N.~Cao, K.~H. Low, and J.~M. Dolan, ``Multi-robot informative path planning for
  active sensing of environmental phenomena: A tale of two algorithms,'' in
  \emph{Proc. {AAMAS}}, 2013, pp. 7--14.

\bibitem{NghiaICML14}
T.~N. Hoang, K.~H. Low, P.~Jaillet, and M.~Kankanhalli, ``Nonmyopic
  $\epsilon$-{B}ayes-optimal active learning of {Gaussian} processes,'' in
  \emph{Proc. {ICML}}, 2014, pp. 739--747.

\bibitem{LowAAMAS08}
K.~H. Low, J.~M. Dolan, and P.~Khosla, ``Adaptive multi-robot wide-area
  exploration and mapping,'' in \emph{Proc. {AAMAS}}, 2008, pp. 23--30.

\bibitem{LowICAPS09}
------, ``Information-theoretic approach to efficient adaptive path planning
  for mobile robotic environmental sensing,'' in \emph{Proc. {ICAPS}}, 2009,
  pp. 233--240.

\bibitem{LowAAMAS11}
------, ``Active {Markov} information-theoretic path planning for robotic
  environmental sensing,'' in \emph{Proc. {AAMAS}}, 2011, pp. 753--760.

\bibitem{LowAAMAS14}
R.~Ouyang, K.~H. Low, J.~Chen, and P.~Jaillet, ``Multi-robot active sensing of
  non-stationary {Gaussian} process-based environmental phenomena,'' in
  \emph{Proc. {AAMAS}}, 2014, pp. 573--580.

\bibitem{YehongAAAI16}
Y.~Zhang, T.~N. Hoang, K.~H. Low, and M.~Kankanhalli, ``Near-optimal active
  learning of multi-output {G}aussian processes,'' in \emph{Proc. {AAAI}},
  2016, pp. 2351--2357.

\bibitem{Erik17}
E.~Daxberger and K.~H. Low, ``Distributed batch {Gaussian} process
  optimization,'' in \emph{Proc. {ICML}}, 2017, pp. 951--960.

\bibitem{Ghahramani14}
J.~M. Hern{\'{a}}ndez-Lobato, M.~W. Hoffman, and Z.~Ghahramani, ``Predictive
  entropy search for efficient global optimization of black-box functions,'' in
  \emph{Proc. NIPS}, 2014, pp. 918--926.

\bibitem{NghiaAAAI18}
T.~N. Hoang, Q.~M. Hoang, and K.~H. Low, ``Decentralized high-dimensional
  {Bayesian} optimization with factor graphs,'' in \emph{Proc. {AAAI}}, 2018,
  pp. 3231--3238.

\bibitem{ling16}
C.~K. Ling, K.~H. Low, and P.~Jaillet, ``{Gaussian} process planning with
  {Lipschitz} continuous reward functions: Towards unifying {Bayesian}
  optimization, active learning, and beyond,'' in \emph{Proc. {AAAI}}, 2016,
  pp. 1860--1866.

\bibitem{Gal2015Improving}
Y.~Gal and R.~Turner, ``Improving the {G}aussian process sparse spectrum
  approximation by representing uncertainty in frequency inputs,'' in
  \emph{Proc. {ICML}}, 2015.

\bibitem{MinhAAAI17}
Q.~M. Hoang, T.~N. Hoang, and K.~H. Low, ``A generalized stochastic variational
  {Bayesian} hyperparameter learning framework for sparse spectrum {Gaussian}
  process regression,'' in \emph{Proc. {AAAI}}, 2017, pp. 2007--2014.

\bibitem{Snelson07a}
E.~L. Snelson and Z.~Ghahramani, ``Local and global sparse {G}aussian process
  approximations,'' in \emph{Proc. AISTATS}, 2007.

\bibitem{HaibinAPP}
H.~Yu, T.~N. Hoang, K.~H. Low, and P.~Jaillet, ``Stochastic variational
  inference for {Bayesian} sparse {Gaussian} process regression,''
  {arXiv:1711.00221}, 2017.

\bibitem{Matthias2016Understanding}
M.~Bauer, M.~van~der Wilk, and C.~E. Rasmussen, ``Understanding probabilistic
  sparse {Gaussian} process approximations,'' in \emph{Proc. NIPS}, 2016, pp.
  1533--1541.

\bibitem{deisenroth2015distributed}
M.~P. Deisenroth and J.~W. Ng, ``Distributed {Gaussian} processes,'' in
  \emph{Proc. ICML}, 2015, pp. 1481--1490.

\bibitem{chen2013gaussian}
J.~Chen, K.~H. Low, and C.~K.~Y. Tan, ``Gaussian process-based decentralized
  data fusion and active sensing for mobility-on-demand system,'' in
  \emph{Proc. {RSS}}, 2013.

\bibitem{LowTASE15}
J.~Chen, K.~H. Low, P.~Jaillet, and Y.~Yao, ``Gaussian process decentralized
  data fusion and active sensing for spatiotemporal traffic modeling and
  prediction in mobility-on-demand systems,'' \emph{{IEEE} Transactions on
  Automation Science and Engineering}, vol.~12, no.~3, pp. 901--921, 2015.

\bibitem{LowUAI12}
J.~Chen, K.~H. Low, C.~K.-Y. Tan, A.~Oran, P.~Jaillet, J.~M. Dolan, and G.~S.
  Sukhatme, ``Decentralized data fusion and active sensing with mobile sensors
  for modeling and predicting spatiotemporal traffic phenomena,'' in
  \emph{Proc. {UAI}}, 2012, pp. 163--173.

\bibitem{NghiaAAAI19}
T.~N. Hoang, Q.~M. Hoang, K.~H. Low, and J.~P. How, ``Collective online
  learning of {Gaussian} processes in massive multi-agent systems,'' in
  \emph{Proc. {AAAI}}, 2019.

\bibitem{Ruofei18}
R.~Ouyang and K.~H. Low, ``Gaussian process decentralized data fusion meets
  transfer learning in large-scale distributed cooperative perception,'' in
  \emph{Proc. AAAI}, 2018, pp. 3876--3883.

\bibitem{IMM2012-03274}
K.~B. Petersen and M.~S. Pedersen, ``The matrix cookbook,'' 2012.

\end{thebibliography}

\clearpage
\appendix


	\subsection{Derivation of~\eqref{log likelihood}}
	\label{Derivation of Equation (15)}
	For all $\mathbf{f}_\mathcal{D}$, $\mathbf{s}_\mathcal{I}$, $\mathbf{\Lambda}$, and $\sigma_f$,\vspace{-1.5mm}
	$$
	\vspace{-4.5mm}
	p(\mathbf{y}_\mathcal{D}) =\frac{p(\mathbf{y}_\mathcal{D},\mathbf{f}_\mathcal{D},\mathbf{s}_\mathcal{I},\mathbf{\Lambda},\sigma_f)}{p(\mathbf{f}_\mathcal{D},\mathbf{s}_\mathcal{I},\mathbf{\Lambda},\sigma_f|\mathbf{y}_\mathcal{D})}\ .
	\vspace{-0mm}
	$$
	So,
	$$	
	\vspace{-0mm}	
	\log p(\mathbf{y}_\mathcal{D})=\log \frac{p(\mathbf{y}_\mathcal{D},\mathbf{f}_\mathcal{D},\mathbf{s}_\mathcal{I},\mathbf{\Lambda},\sigma_f)}{p(\mathbf{f}_\mathcal{D},\mathbf{s}_\mathcal{I},\mathbf{\Lambda},\sigma_f|\mathbf{y}_\mathcal{D})}\ .
	\vspace{1mm}
	$$
	Let $q(\mathbf{f}_\mathcal{D},\mathbf{s}_\mathcal{I},\mathbf{\Lambda},\sigma_f)$ be an arbitrary probability density function that is independent of $\mathbf{y}_\mathcal{D}$. Integrating both sides of the above equation with respect to $q(\mathbf{f}_\mathcal{D},\mathbf{s}_\mathcal{I},\mathbf{\Lambda},\sigma_f)$ yields\vspace{-1mm}
	\begin{equation}\vspace{-3mm}
	\hspace{-1.7mm}
	\begin{array}{l}
	\log p(\mathbf{y}_\mathcal{D})\\
	\displaystyle=\hspace{-2mm}\int\hspace{-1mm} q(\mathbf{f}_\mathcal{D},\mathbf{s}_\mathcal{I},\mathbf{\Lambda},\sigma_f)\log \frac{p(\mathbf{y}_\mathcal{D},\mathbf{f}_\mathcal{D},\mathbf{s}_\mathcal{I},\mathbf{\Lambda},\sigma_f)}{p(\mathbf{f}_\mathcal{D},\mathbf{s}_\mathcal{I},\mathbf{\Lambda},\sigma_f|\mathbf{y}_\mathcal{D})} \mathrm{d}\mathbf{f}_\mathcal{D}\mathrm{d}\mathbf{s}_\mathcal{I}\mathrm{d}\mathbf{\Lambda} \mathrm{d}\sigma_f
	\end{array}
	\label{crab}\vspace{0mm}
	\end{equation}
	Using $\log(ab)=\log(a)+\log(b)$, 
	\begin{equation*}
		\begin{array}{rcl}
			\displaystyle	\log \frac{p(\mathbf{y}_\mathcal{D},\mathbf{f}_\mathcal{D},\mathbf{s}_\mathcal{I},\mathbf{\Lambda},\sigma_f)}{p(\mathbf{f}_\mathcal{D},\mathbf{s}_\mathcal{I},\mathbf{\Lambda},\sigma_f|\mathbf{y}_\mathcal{D})}&\hspace{-2.4mm}=&\hspace{-2.4mm}\displaystyle\log\dfrac{p(\mathbf{y}_\mathcal{D},\mathbf{f}_\mathcal{D},\mathbf{s}_\mathcal{I},\mathbf{\Lambda},\sigma_f)}{q(\mathbf{f}_\mathcal{D},\mathbf{s}_\mathcal{I},\mathbf{\Lambda},\sigma_f)}\vspace{1mm}\displaystyle\vspace{1mm}\\
			&&\hspace{-2.4mm}\displaystyle +\log\dfrac{q(\mathbf{f}_\mathcal{D},\mathbf{s}_\mathcal{I},\mathbf{\Lambda},\sigma_f)}{p(\mathbf{f}_\mathcal{D},\mathbf{s}_\mathcal{I},\mathbf{\Lambda},\sigma_f|\mathbf{y}_\mathcal{D})}
		\end{array}
		\vspace{-2mm}
	\end{equation*}
	\vspace{-1mm}
	which is substituted into~\eqref{crab} to give
	\vspace{-1mm}
	\begin{equation}
	\hspace{-1.7mm}
		\begin{array}{l}
			\log p(\mathbf{y}_\mathcal{D})=\vspace{1mm}\\
			\displaystyle\int q(\mathbf{f}_\mathcal{D},\mathbf{s}_\mathcal{I},\mathbf{\Lambda},\sigma_f)\log\dfrac{p(\mathbf{y}_\mathcal{D},\mathbf{f}_\mathcal{D},\mathbf{s}_\mathcal{I},\mathbf{\Lambda},\sigma_f)}{q(\mathbf{f}_\mathcal{D},\mathbf{s}_\mathcal{I},\mathbf{\Lambda},\sigma_f)} \mathrm{d}\mathbf{f}_\mathcal{D} \mathrm{d}\mathbf{s}_\mathcal{I} \mathrm{d}\mathbf{\Lambda} \mathrm{d}\sigma_f \vspace{1mm}\\
			\displaystyle+\hspace{-1mm}\int\hspace{-1mm} q(\mathbf{f}_\mathcal{D},\mathbf{s}_\mathcal{I},\mathbf{\Lambda},\sigma_f)\log\dfrac{q(\mathbf{f}_\mathcal{D},\mathbf{s}_\mathcal{I},\mathbf{\Lambda},\sigma_f)}{p(\mathbf{f}_\mathcal{D},\mathbf{s}_\mathcal{I},\mathbf{\Lambda},\sigma_f|\mathbf{y}_\mathcal{D})} \mathrm{d}\mathbf{f}_\mathcal{D}\mathrm{d}\mathbf{s}_\mathcal{I} \mathrm{d}\mathbf{\Lambda} \mathrm{d}\sigma_f .
		\end{array}
		\label{crab2}\vspace{1mm}
	\end{equation}\vspace{0mm}
	The first and second terms on the RHS of~\eqref{crab2} correspond to the variational lower bound $\mathcal{L}(q)$ and $\mathrm{KL}(q(\mathbf{f}_\mathcal{D},\mathbf{s}_\mathcal{I},\mathbf{\Lambda},\sigma_f)|| p(\mathbf{f}_\mathcal{D},\mathbf{s}_\mathcal{I},\mathbf{\Lambda},\sigma_f|\mathbf{y}_\mathcal{D}))$, respectively.
\vspace{-3mm}
\subsection{Proof of Theorem~\ref{thm1}}
	\label{Derivation of the Lower Bound}
	\label{Derivation of Lower Bound L}
	\vspace{-3mm}
	Given that 	\vspace{-2mm}
	\begin{equation*}
		\begin{array}{l}	\vspace{-0mm}
	p(\mathbf{y}_\mathcal{D},\mathbf{f}_\mathcal{D},\mathbf{s}_\mathcal{I},\mathbf{\Lambda},\sigma_f)=
				\\
	p(\mathbf{y}_\mathcal{D}|\mathbf{f}_\mathcal{D}) p(\mathbf{f}_\mathcal{D}|\mathbf{s}_\mathcal{I},\mathbf{\Lambda},\sigma_f)\ p(\mathbf{s}_\mathcal{I})\ p(\mathbf{\Lambda})\ p(\sigma_f)
		\end{array}
	\end{equation*}
	\vspace{-3mm}
	\begin{equation*}
		\hspace{-1.7mm}
		\begin{array}{l}
			\mathcal{L}(q)\\
			\displaystyle=\hspace{-2mm}\int \hspace{-1mm} q(\mathbf{f}_\mathcal{D},\mathbf{s}_\mathcal{I},\mathbf{\Lambda},\sigma_f)\log\dfrac{p(\mathbf{y}_\mathcal{D},\mathbf{f}_\mathcal{D},\mathbf{s}_\mathcal{I},\mathbf{\Lambda},\sigma_f)}{q(\mathbf{f}_\mathcal{D},\mathbf{s}_\mathcal{I},\mathbf{\Lambda},\sigma_f)} \mathrm{d}\mathbf{f}_\mathcal{D}\mathrm{d}\mathbf{s}_\mathcal{I} \mathrm{d}\mathbf{\Lambda}\mathrm{d}\sigma_f \vspace{-1mm}
			\\
			\displaystyle=\int q(\mathbf{f}_\mathcal{D},\mathbf{s}_\mathcal{I},\mathbf{\Lambda},\sigma_f) \\ 
			\displaystyle \hspace{-0.5mm}\log\hspace{-0.5mm}\dfrac{p(\mathbf{y}_\mathcal{D}|\mathbf{f}_\mathcal{D})p(\mathbf{f}_\mathcal{D}|\mathbf{s}_\mathcal{I},\mathbf{\Lambda},\sigma_f)p(\mathbf{s}_\mathcal{I})p(\mathbf{\Lambda})p(\sigma_f)}{p(\mathbf{f}_\mathcal{D}|\mathbf{s}_\mathcal{I},\mathbf{\Lambda},\sigma_f)q(\mathbf{s}_\mathcal{I})q(\mathbf{\Lambda})q(\sigma_f)} \mathrm{d}\mathbf{f}_\mathcal{D} \mathrm{d}\mathbf{s}_\mathcal{I} \mathrm{d}\mathbf{\Lambda} \mathrm{d}\sigma_f 
			\\
			\displaystyle=\int q(\mathbf{f}_\mathcal{D},\mathbf{s}_\mathcal{I},\mathbf{\Lambda},\sigma_f)\\
			\quad\displaystyle\log \dfrac{p(\mathbf{y}_\mathcal{D}|\mathbf{f}_\mathcal{D})p(\mathbf{s}_\mathcal{I})p(\mathbf{\Lambda})p(\sigma_f)}{q(\mathbf{s}_\mathcal{I})q(\mathbf{\Lambda})q(\sigma_f)}\ \mathrm{d}\mathbf{f}_\mathcal{D}\ \mathrm{d}\mathbf{s}_\mathcal{I}\ \mathrm{d}\mathbf{\Lambda}\  \mathrm{d}\sigma_f\\
			\displaystyle=\int p(\mathbf{f}_\mathcal{D}|\mathbf{s}_\mathcal{I},\mathbf{\Lambda},\sigma_f)\ q(\mathbf{s}_\mathcal{I})\ q(\mathbf{\Lambda})\ q(\sigma_f)\Bigg(\log p(\mathbf{y}_\mathcal{D}|\mathbf{f}_\mathcal{D})\ +\\
			\quad\displaystyle\log\dfrac{p(\mathbf{s}_\mathcal{I})}{q(\mathbf{s}_\mathcal{I})}+\log\dfrac{p(\mathbf{\Lambda})}{q(\mathbf{\Lambda})}+\log\dfrac{p(\sigma_f)}{q(\sigma_f)}\Bigg) \mathrm{d}\mathbf{f}_\mathcal{D}\ \mathrm{d}\mathbf{s}_\mathcal{I}\ \mathrm{d}\mathbf{\Lambda}\ \mathrm{d}\sigma_f\vspace{0mm}\\
			\displaystyle= \mathcal{F}(q)+\int q(\mathbf{s}_\mathcal{I})\log\dfrac{p(\mathbf{s}_\mathcal{I})}{q(\mathbf{s}_\mathcal{I})}\ \mathrm{d}\mathbf{s}_\mathcal{I}+
			\int q(\mathbf{\Lambda})\log\dfrac{p(\mathbf{\Lambda})}{q(\mathbf{\Lambda})}\ \mathrm{d}\mathbf{\Lambda}\ \vspace{-0mm}\\
			\displaystyle\quad+\int q(\sigma_f)\log\dfrac{p(\sigma_f)}{q(\sigma_f)}\ \mathrm{d}\sigma_f
		\end{array}
	\end{equation*}
	where
	\begin{equation*}
		\begin{array}{rcl}
	\mathcal{F}(q)&\hspace{-2.4mm}=&\hspace{-2.4mm}\displaystyle\int q(\mathbf{s}_\mathcal{I})\ \mathcal{G}(q,\mathbf{s}_\mathcal{I})\ \mathrm{d}\mathbf{s}_\mathcal{I}\\
	\mathcal{G}(q, \mathbf{s}_\mathcal{I})&\hspace{-2.4mm}=&\hspace{-2.4mm}\displaystyle\int q(\sigma_f)\ q(\mathbf{\Lambda})\ \mathcal{H}(\mathbf{s}_\mathcal{I},\mathbf{\Lambda},\sigma_f)\ \mathrm{d}\mathbf{\Lambda}\ \mathrm{d}\sigma_f \\
	\mathcal{H}(\mathbf{s}_\mathcal{I},\mathbf{\Lambda},\sigma_f)&\hspace{-2.4mm}=&\hspace{-2.4mm}\displaystyle\int p(\mathbf{f}_\mathcal{D}|\mathbf{s}_\mathcal{I},\mathbf{\Lambda},\sigma_f)\ \log p(\mathbf{y}_\mathcal{D}|\mathbf{f}_\mathcal{D})\ \mathrm{d}\mathbf{f}_\mathcal{D}\ .
		\end{array}
	\end{equation*}
	Let us first derive the closed-form expression of $H(\mathbf{s}_\mathcal{I},\mathbf{\Lambda},\sigma_f)$:
	\begin{equation*}
		\hspace{-1.7mm}
		\begin{array}{l}
			H(\mathbf{s}_\mathcal{I},\mathbf{\Lambda},\sigma_f)\\
			\displaystyle=\int p(\mathbf{f}_\mathcal{D}|\mathbf{s}_\mathcal{I},\mathbf{\Lambda},\sigma_f)\log p(\mathbf{y}_\mathcal{D}|\mathbf{f}_\mathcal{D})\ \mathrm{d}\mathbf{f}_\mathcal{D}  \\
			\displaystyle = \int p(\mathbf{f}_\mathcal{D}|\mathbf{s}_\mathcal{I},\mathbf{\Lambda},\sigma_f)
			\Bigg(-\frac{|\mathcal{D}|}{2}\log2\pi-\frac{1}{2}\log|\mathbf{C}_\mathcal{DD}|\\
			\displaystyle\quad\qquad\qquad\qquad\qquad-\frac{1}{2}(\mathbf{y}_\mathcal{D}-\mathbf{f}_\mathcal{D})^\top\mathbf{C}_\mathcal{DD}^{-1}(\mathbf{y}_\mathcal{D}-\mathbf{f}_\mathcal{D})\Bigg) \mathrm{d}\mathbf{f}_\mathcal{D} \vspace{1mm}\\
			\displaystyle = -\frac{|\mathcal{D}|}{2}\log2\pi-\frac{1}{2}\log|\mathbf{C}_\mathcal{DD}|\vspace{1mm}\\
			\displaystyle\quad-\mathbb{E}_{p(\mathbf{f}_\mathcal{D}|\mathbf{s}_\mathcal{I},\mathbf{\Lambda},\sigma_f)}\left[\frac{1}{2}(\mathbf{y}_\mathcal{D}-\mathbf{f}_\mathcal{D})^\top\mathbf{C}_\mathcal{DD}^{-1}(\mathbf{y}_\mathcal{D}-\mathbf{f}_\mathcal{D})\right] \vspace{1mm}\\
			\displaystyle = -\frac{|\mathcal{D}|}{2}\log2\pi-\frac{1}{2}\log|\mathbf{C}_\mathcal{DD}|\\
			\displaystyle\quad-\frac{1}{2}(\mathbf{y}_\mathcal{D}-\mathbf{K}_\mathcal{DI}\mathbf{\Sigma}_\mathcal{II}^{-1}\mathbf{s}_\mathcal{I})^\top\mathbf{C}_\mathcal{DD}^{-1}(\mathbf{y}_\mathcal{D}-\mathbf{K}_\mathcal{DI}\mathbf{\Sigma}_\mathcal{II}^{-1}\mathbf{s}_\mathcal{I})\vspace{1mm}\\
			\displaystyle\quad-\frac{1}{2}\mathrm{Tr}[\mathbf{C}_\mathcal{DD}^{-1}\mathbf{K}_\mathcal{DD}]+\frac{1}{2}\mathrm{Tr}[\mathbf{C}_\mathcal{DD}^{-1}\mathbf{K}_\mathcal{DI}\mathbf{\Sigma}_\mathcal{II}^{-1}\mathbf{K}_\mathcal{ID}]
		\end{array}			
	\end{equation*}
	where the last equality follows from eq.~$380$ of~\cite{IMM2012-03274} and~\eqref{Modified Gaussian Process Model}.
	%
	
	The closed-form expression of $\mathcal{G}(q, \mathbf{s}_\mathcal{I})$ can then be derived as follows:
	\begin{equation*}
		\hspace{-1.7mm}{}
		\begin{array}{l}
			\mathcal{G}(q, \mathbf{s}_\mathcal{I})\\
			\displaystyle=\int q(\sigma_f)\ q(\mathbf{\Lambda})\ \mathcal{H}(\mathbf{s}_\mathcal{I},\mathbf{\Lambda},\sigma_f)\ \mathrm{d}\mathbf{\Lambda}\ \mathrm{d}\sigma_f\\
			\displaystyle=-\frac{|\mathcal{D}|}{2}\log2\pi-\frac{1}{2}\log|\mathbf{C}_\mathcal{DD}|\vspace{1mm}\\
			\displaystyle\hspace{-2mm} -\frac{1}{2}\mathbb{E}_{q(\mathbf{\Lambda},\sigma_f)}\left[(\mathbf{y}_\mathcal{D}-\mathbf{K}_\mathcal{DI}\mathbf{\Sigma}_\mathcal{II}^{-1}\mathbf{s}_\mathcal{I})^\top\mathbf{C}_\mathcal{DD}^{-1}(\mathbf{y}_\mathcal{D}-\mathbf{K}_\mathcal{DI}\mathbf{\Sigma}_\mathcal{II}^{-1}\mathbf{s}_\mathcal{I})\right]\vspace{1mm}\\
			\displaystyle \hspace{-2mm}-\frac{1}{2}\mathrm{Tr}[\mathbf{C}_\mathcal{DD}^{-1}\mathbb{E}_{q(\mathbf{\Lambda},\sigma_f)}[\mathbf{K}_\mathcal{DD}]]\hspace{-0.5mm} +\hspace{-0.5mm}\frac{1}{2}\mathrm{Tr}[\mathbf{\Sigma}_\mathcal{II}^{-1}\mathbb{E}_{q(\mathbf{\Lambda},\sigma_f)}[\mathbf{K}_\mathcal{ID}\mathbf{C}_\mathcal{DD}^{-1}\mathbf{K}_\mathcal{DI}]] \vspace{1mm}\\
			\displaystyle=-\frac{|\mathcal{D}|}{2}\log2\pi-\frac{1}{2}\log|\mathbf{C}_\mathcal{DD}|-\frac{1}{2}\mathbf{y}_\mathcal{D}^\top\mathbf{C}_\mathcal{DD}^{-1}\mathbf{y}_\mathcal{D}\vspace{1mm}\\
			\displaystyle\quad +\mathbf{s}_\mathcal{I}^\top\mathbf{\Sigma}_\mathcal{II}^{-1}\mathbf{\Omega}_\mathcal{ID}\mathbf{C}_\mathcal{DD}^{-1}\mathbf{y}_\mathcal{D}-\frac{1}{2}\mathbf{s}_\mathcal{I}^\top\mathbf{\Sigma}_\mathcal{II}^{-1}\mathbf{\Psi}_\mathcal{II}\mathbf{\Sigma}_\mathcal{II}^{-1}\mathbf{s}_\mathcal{I}\vspace{1mm}\\
			\displaystyle\quad -\frac{1}{2}\mathrm{Tr}[\mathbf{C}_\mathcal{DD}^{-1}\mathbf{\Upsilon}_\mathcal{DD}] +\frac{1}{2}\mathrm{Tr}[\mathbf{\Sigma}_\mathcal{II}^{-1}\mathbf{\Psi}_\mathcal{II}] 
		\end{array}
	\end{equation*}
	such that the last equality follows from
	\begin{equation*}
		\hspace{-1.7mm}
		\begin{array}{l}
			\displaystyle\mathbb{E}_{q(\mathbf{\Lambda},\sigma_f)}\left[(\mathbf{y}_\mathcal{D}-\mathbf{K}_\mathcal{DI}\mathbf{\Sigma}_\mathcal{II}^{-1}\mathbf{s}_\mathcal{I})^\top\mathbf{C}_\mathcal{DD}^{-1}(\mathbf{y}_\mathcal{D}-\mathbf{K}_\mathcal{DI}\mathbf{\Sigma}_\mathcal{II}^{-1}\mathbf{s}_\mathcal{I})\right] \vspace{1mm}\\
			\displaystyle=\mathbb{E}_{q(\mathbf{\Lambda},\sigma_f)}\left[(\mathbf{y}_\mathcal{D}^\top-\mathbf{s}_\mathcal{I}^\top\mathbf{\Sigma}_\mathcal{II}^{-1}\mathbf{K}_\mathcal{ID})\mathbf{C}_\mathcal{DD}^{-1}(\mathbf{y}_\mathcal{D}-\mathbf{K}_\mathcal{DI}\mathbf{\Sigma}_\mathcal{II}^{-1}\mathbf{s}_\mathcal{I})\right] \vspace{1mm}\\
			\displaystyle=\mathbf{y}_\mathcal{D}^\top\mathbf{C}_\mathcal{DD}^{-1}\mathbf{y}_\mathcal{D}\hspace{-0.5mm}-\hspace{-0.5mm}2\mathbf{s}_\mathcal{I}^\top\mathbf{\Sigma}_\mathcal{II}^{-1}\mathbf{\Omega}_\mathcal{ID}\mathbf{C}_\mathcal{DD}^{-1}\mathbf{y}_\mathcal{D}\hspace{-0.5mm}+\hspace{-0.5mm}\mathbf{s}_\mathcal{I}^\top\mathbf{\Sigma}_\mathcal{II}^{-1}\mathbf{\Psi}_\mathcal{II}\mathbf{\Sigma}_\mathcal{II}^{-1}\mathbf{s}_\mathcal{I}.	
		\end{array}
	\end{equation*}
	The closed-form expression of $\mathcal{F}(q)$ is \vspace{-2mm}
	\begin{equation*}
		\vspace{-2mm}
		\begin{array}{rcl}
				\mathcal{F}(q)&\hspace{-2.4mm}=&\hspace{-2.4mm}\displaystyle\int q(\mathbf{s}_\mathcal{I})\ \mathcal{G}(q,\mathbf{s}_\mathcal{I})\ \mathrm{d}\mathbf{s}_\mathcal{I}
				=\displaystyle\mathbb{E}_{q(\mathbf{s}_\mathcal{I})}[\mathcal{G}(q,\mathbf{s}_\mathcal{I})] 
		\end{array}
	\end{equation*}
	\vspace{-1mm}
	where, using $q(\mathbf{s}_\mathcal{I})=\mathcal{N}(\mathbf{m},\mathbf{S})$,
	\begin{equation*}
		\hspace{-1.7mm}
		\begin{array}{l}
			\displaystyle\mathbb{E}_{q(\mathbf{s}_\mathcal{I})}[\mathcal{G}(q,\mathbf{s}_\mathcal{I})] \vspace{1mm}\\
			\displaystyle=-\frac{|\mathcal{D}|}{2}\log2\pi-\frac{1}{2}\log|\mathbf{C}_\mathcal{DD}|-\frac{1}{2}\mathbf{y}_\mathcal{D}^\top\mathbf{C}_\mathcal{DD}^{-1}\mathbf{y}_\mathcal{D}\vspace{1mm}\\
			\displaystyle\quad+\mathbf{m}^\top\mathbf{\Sigma}_\mathcal{II}^{-1}\mathbf{\Omega}_\mathcal{ID}\mathbf{C}_\mathcal{DD}^{-1}\mathbf{y}_\mathcal{D}-\frac{1}{2}\mathbf{m}^\top\mathbf{\Sigma}_\mathcal{II}^{-1}\mathbf{\Psi}_\mathcal{II}\mathbf{\Sigma}_\mathcal{II}^{-1}\mathbf{m}\vspace{1mm}\\
			\displaystyle\quad-\frac{1}{2}\mathrm{Tr}[\mathbf{S}\mathbf{\Sigma}_\mathcal{II}^{-1}\mathbf{\Psi}_\mathcal{II}\mathbf{\Sigma}_\mathcal{II}^{-1}]\vspace{1mm}\\
			\displaystyle\quad-\frac{1}{2}\mathrm{Tr}[\mathbf{C}_\mathcal{DD}^{-1}\mathbf{\Upsilon}_\mathcal{DD}]+\frac{1}{2}\mathrm{Tr}[\mathbf{\Sigma}_\mathcal{II}^{-1}\mathbf{\Psi}_\mathcal{II}] \vspace{1mm}\\
			
		\end{array}
	\end{equation*}
	such that $\mathbb{E}_{q(\mathbf{s}_\mathcal{I})}[\mathcal{G}(q,\mathbf{s}_\mathcal{I})]$ is derived using eqs.~$374$ and~$380$ of~\cite{IMM2012-03274}. Since \vspace{-1mm} 
	\begin{equation*}
		\int q(\mathbf{s}_\mathcal{I})\log\dfrac{p(\mathbf{s}_\mathcal{I})}{q(\mathbf{s}_\mathcal{I})}\ \mathrm{d}\mathbf{s}_\mathcal{I}=\mathbb{E}_{q(\mathbf{s}_\mathcal{I})}[\log p(\mathbf{s}_\mathcal{I})]+\mathbb{H}[q(\mathbf{s}_\mathcal{I})]
	\end{equation*}
	\vspace{-1mm}
	where
	\begin{equation*}
	\hspace{-1.7mm}
	\begin{array}{l}
		\displaystyle\mathbb{E}_{q(\mathbf{s}_\mathcal{I})}[\log p(\mathbf{s}_\mathcal{I})]\vspace{1mm}\\
		\displaystyle=-\frac{|\mathcal{I}|}{2}\log2\pi-\frac{1}{2}\log|\mathbf{\Sigma}_\mathcal{II}|-\frac{1}{2}\mathbf{m}^\top\mathbf{\Sigma}_\mathcal{II}^{-1}\mathbf{m}-\frac{1}{2}\mathrm{Tr}[\mathbf{S}\mathbf{\Sigma}_\mathcal{II}^{-1}]
	\end{array}
	\end{equation*}
and 
$$
\displaystyle\mathbb{H}[q(\mathbf{s}_\mathcal{I})]=\frac{|\mathcal{I}|}{2}\log2\pi+\frac{|\mathcal{I}|}{2}+\frac{1}{2}\log|\mathbf{S}|
$$
\vspace{-1mm}
denotes a Gaussian entropy with respect to $q(\mathbf{s}_\mathcal{I})$,	
$$
\displaystyle\int q(\mathbf{\Lambda})\log\dfrac{p(\mathbf{\Lambda})}{q(\mathbf{\Lambda})}d\mathbf{\Lambda}=-\dfrac{1}{2}\boldsymbol{\nu}^\top\boldsymbol{\nu}-\dfrac{1}{2}\mathrm{Tr}[\mathbf{\Xi}]+\dfrac{1}{2}\log|\mathbf{\Xi}|+\dfrac{d}{2}\ ,
$$
and
$$	
\displaystyle \int q(\sigma_f)\log\dfrac{p(\sigma_f)}{q(\sigma_f)}\ \mathrm{d}\sigma_f=-\dfrac{1}{2}\alpha^2-\dfrac{1}{2}\beta+\dfrac{1}{2}\log \beta+\dfrac{1}{2}\ ,
$$
	\begin{equation*}
		\hspace{-1.7mm}
		\begin{array}{l}
			\displaystyle\hspace{5mm}\mathcal{L}(q)\vspace{0mm}\\
			\displaystyle= \mathcal{F}(q)+\int q(\mathbf{s}_\mathcal{I})\log\dfrac{p(\mathbf{s}_\mathcal{I})}{q(\mathbf{s}_\mathcal{I})}\ \mathrm{d}\mathbf{s}_\mathcal{I}+
			\int q(\mathbf{\Lambda})\log\dfrac{p(\mathbf{\Lambda})}{q(\mathbf{\Lambda})}\ \mathrm{d}\mathbf{\Lambda}\ \vspace{-0.5mm}\\
			\displaystyle\quad+\int q(\sigma_f)\log\dfrac{p(\sigma_f)}{q(\sigma_f)}\ \mathrm{d}\sigma_f\vspace{0.5mm}\\
			\displaystyle=-\frac{|\mathcal{D}|}{2}\log2\pi-\frac{1}{2}\log|\mathbf{C}_\mathcal{DD}|-\frac{1}{2}\mathbf{y}_\mathcal{D}^\top\mathbf{C}_\mathcal{DD}^{-1}\mathbf{y}_\mathcal{D}\vspace{1mm}\\
			\displaystyle\quad+\mathbf{m}^\top\mathbf{\Sigma}_\mathcal{II}^{-1}\mathbf{\Omega}_\mathcal{ID}\mathbf{C}_\mathcal{DD}^{-1}\mathbf{y}_\mathcal{D}-\frac{1}{2}\mathbf{m}^\top(\mathbf{\Sigma}_\mathcal{II}^{-1}\mathbf{\Psi}_\mathcal{II}\mathbf{\Sigma}_\mathcal{II}^{-1}+\mathbf{\Sigma}_\mathcal{II}^{-1})\mathbf{m}\vspace{1mm}\\
			\displaystyle\quad-\frac{1}{2}\mathrm{Tr}[\mathbf{S}(\mathbf{\Sigma}_\mathcal{II}^{-1}\mathbf{\Psi}_\mathcal{II}\mathbf{\Sigma}_\mathcal{II}^{-1}+ \mathbf{\Sigma}_\mathcal{II}^{-1})]-\frac{1}{2}\mathrm{Tr}[\mathbf{C}_\mathcal{DD}^{-1}\mathbf{\Upsilon}_\mathcal{DD}]\vspace{1mm}\\
			\displaystyle\quad+\frac{1}{2}\mathrm{Tr}[\mathbf{\Sigma}_\mathcal{II}^{-1}\mathbf{\Psi}_\mathcal{II}]-\frac{1}{2}\log|\mathbf{\Sigma}_\mathcal{II}|+\frac{|\mathcal{I}|}{2}+\frac{1}{2}\log|\mathbf{S}|\vspace{1mm}\\
			\displaystyle \quad-\dfrac{1}{2}\boldsymbol{\nu}^\top\boldsymbol{\nu}-\dfrac{1}{2}\mathrm{Tr}[\mathbf{\Xi}]+\dfrac{1}{2}\log|\mathbf{\Xi}|+\dfrac{d}{2}-\dfrac{1}{2}\alpha^2-\dfrac{1}{2}\beta\vspace{0.5mm}\\
			\quad+\dfrac{1}{2}\log \beta+\dfrac{1}{2}\vspace{0.5mm}\\
			\displaystyle\hspace{-0.5mm}=\hspace{-0.5mm} \dfrac{1}{2}\hspace{-0.5mm}\Big(\hspace{-0.5mm}2\mathbf{m}^\top\mathbf{\Sigma}_\mathcal{II}^{-1}\mathbf{\Omega}_\mathcal{ID}\mathbf{C}_\mathcal{DD}^{-1}\mathbf{y}_\mathcal{D}\hspace{-0.5mm}-\hspace{-0.5mm}\mathbf{m}^\top(\mathbf{\Sigma}_\mathcal{II}^{-1}\mathbf{\Psi}_\mathcal{II}\mathbf{\Sigma}_\mathcal{II}^{-1}+\mathbf{\Sigma}_\mathcal{II}^{-1})\mathbf{m}\vspace{0.5mm}\\
			\displaystyle\quad-\mathrm{Tr}[\mathbf{S}(\mathbf{\Sigma}_\mathcal{II}^{-1}\mathbf{\Psi}_\mathcal{II}\mathbf{\Sigma}_\mathcal{II}^{-1}+ \mathbf{\Sigma}_\mathcal{II}^{-1})]-\mathrm{Tr}[\mathbf{C}_\mathcal{DD}^{-1}\mathbf{\Upsilon}_\mathcal{DD}]\vspace{1mm}\\
			\displaystyle\quad+\mathrm{Tr}[\mathbf{\Sigma}_\mathcal{II}^{-1}\mathbf{\Psi}_\mathcal{II}]+\log|\mathbf{S}|-\boldsymbol{\nu}^\top\boldsymbol{\nu}-\mathrm{Tr}[\mathbf{\Xi}]+\log|\mathbf{\Xi}|\vspace{1mm}\\
			\displaystyle\quad-\alpha^2-\beta
			+\log \beta\Big)+ \mathrm{const}
		\end{array}
	\end{equation*}
	where $\mathrm{const}$ absorbs all terms independent of $\mathbf{m}$, $\mathbf{S}$, $\boldsymbol{\nu}$, $\mathbf{\Xi},\alpha,\beta$.
	Then, by setting
	\vspace{-1mm}
	\begin{equation*}
		\begin{array}{rcl}
			\dfrac{\partial\mathcal{L}}{\partial\mathbf{m}}&\hspace{-2.4mm}=&\hspace{-2.4mm}\displaystyle\mathbf{\Sigma}_\mathcal{II}^{-1}\mathbf{\Omega}_\mathcal{ID}\mathbf{C}_\mathcal{DD}^{-1}\mathbf{y}_\mathcal{D}-(\mathbf{\Sigma}_\mathcal{II}^{-1}\mathbf{\Psi}_\mathcal{II}\mathbf{\Sigma}_\mathcal{II}^{-1}+\mathbf{\Sigma}_\mathcal{II}^{-1})\mathbf{m}\ ,\vspace{1mm} \\
			\dfrac{\partial\mathcal{L}}{\partial\mathbf{S}}&\hspace{-2.4mm}=&\hspace{-2.4mm}\displaystyle\dfrac{1}{2}\mathbf{S}^{-1}-\dfrac{1}{2}(\mathbf{\Sigma}_\mathcal{II}^{-1}\mathbf{\Psi}_\mathcal{II}\mathbf{\Sigma}_\mathcal{II}^{-1}+\mathbf{\Sigma}_\mathcal{II}^{-1})
		\end{array}\vspace{0mm}
	\end{equation*}
	to zero, it can be derived that $\mathcal{L}(q)$ is maximized at $q^*(\mathbf{s}_\mathcal{I})=\mathcal{N}(\mathbf{m}^*,\mathbf{S}^*)$ where
	\begin{equation}
		\begin{array}{rcl}
			\mathbf{m}^*&\hspace{-2.4mm}=&\hspace{-2.4mm}\displaystyle(\mathbf{\Sigma}_\mathcal{II}^{-1}\mathbf{\Psi}_\mathcal{II}\mathbf{\Sigma}_\mathcal{II}^{-1}+\mathbf{\Sigma}_\mathcal{II}^{-1})^{-1}\mathbf{\Sigma}_\mathcal{II}^{-1}\mathbf{\Omega}_\mathcal{ID}\mathbf{C}_\mathcal{DD}^{-1}\mathbf{y}_\mathcal{D}\ , \vspace{1mm} \\
			\mathbf{S}^*&\hspace{-2.4mm}=&\hspace{-2.4mm}\displaystyle(\mathbf{\Sigma}_\mathcal{II}^{-1}\mathbf{\Psi}_\mathcal{II}\mathbf{\Sigma}_\mathcal{II}^{-1}+\mathbf{\Sigma}_\mathcal{II}^{-1})^{-1}\ .
		\end{array}
		\label{happy}
	\end{equation}
	By substituting	
	\begin{equation*}
		\begin{array}{l}
			\displaystyle(\mathbf{\Sigma}_\mathcal{II}^{-1}\mathbf{\Psi}_\mathcal{II}\mathbf{\Sigma}_\mathcal{II}^{-1}+\mathbf{\Sigma}_\mathcal{II}^{-1})^{-1}\vspace{1mm}\\
			\displaystyle=((\mathbf{I}+\mathbf{\Sigma}_\mathcal{II}^{-1}\mathbf{\Psi}_\mathcal{II})\mathbf{\Sigma}_\mathcal{II}^{-1})^{-1}\vspace{1mm}\\
			\displaystyle=\mathbf{\Sigma}_\mathcal{II}(\mathbf{I}+\mathbf{\Sigma}_\mathcal{II}^{-1}\mathbf{\Psi}_\mathcal{II})^{-1}\vspace{1mm}\\
			\displaystyle=\mathbf{\Sigma}_\mathcal{II}(\mathbf{\Sigma}_\mathcal{II}^{-1}(\mathbf{\Sigma}_\mathcal{II}+\mathbf{\Psi}_\mathcal{II}))^{-1}\vspace{1mm}\\
			\displaystyle=\mathbf{\Sigma}_\mathcal{II}(\mathbf{\Sigma}_\mathcal{II}+\mathbf{\Psi}_\mathcal{II})^{-1}\mathbf{\Sigma}_\mathcal{II}
		\end{array}
	\end{equation*}
	into~\eqref{happy},~\eqref{q(u)} in Theorem~\ref{thm1} results.
	Using~\eqref{q(u)},
	%
	\begin{equation*}
		\begin{array}{l}
			\displaystyle\mathbf{m}^{*\top}\mathbf{\Sigma}_\mathcal{II}^{-1}\mathbf{\Omega}_\mathcal{ID}\mathbf{C}_\mathcal{DD}^{-1}\mathbf{y}_\mathcal{D}\vspace{1mm}\\
			\displaystyle=\mathbf{y}_\mathcal{D}^\top\mathbf{C}_\mathcal{DD}^{-1}\mathbf{\Omega}_\mathcal{ID}^\top(\mathbf{\Sigma}_\mathcal{II}+\mathbf{\Psi}_\mathcal{II})^{-1}\mathbf{\Omega}_\mathcal{ID}\mathbf{C}_\mathcal{DD}^{-1}\mathbf{y}_\mathcal{D}\ , \vspace{2mm}\\
			\displaystyle\mathbf{m}^{*\top}\big(\mathbf{\Sigma}_\mathcal{II}^{-1}\mathbf{\Psi}_\mathcal{II}\mathbf{\Sigma}_\mathcal{II}^{-1}+\mathbf{\Sigma}_\mathcal{II}^{-1}\big)\mathbf{m}^*\vspace{1mm}\\
			\displaystyle=\mathbf{y}_\mathcal{D}^\top\mathbf{C}_\mathcal{DD}^{-1}\mathbf{\Omega}_\mathcal{ID}^\top(\mathbf{\Sigma}_\mathcal{II}+\mathbf{\Psi}_\mathcal{II})^{-1}\mathbf{\Omega}_\mathcal{ID}\mathbf{C}_\mathcal{DD}^{-1}\mathbf{y}_\mathcal{D}\ , \vspace{2mm}\\
			\displaystyle\mathrm{Tr}(\mathbf{S}^*(\mathbf{\Sigma}_\mathcal{II}^{-1}\mathbf{\Psi}_\mathcal{II}\mathbf{\Sigma}_\mathcal{II}^{-1}+\mathbf{\Sigma}_\mathcal{II}^{-1}))=|\mathcal{I}|
		\end{array}
	\end{equation*}
	\vspace{-1mm}
	which reduce $\mathcal{L}(q)$ to\vspace{-1mm}
	\begin{equation}
		\begin{array}{l}
			\mathcal{L}(q)=
			\displaystyle\frac{1}{2}\Big(\mathbf{y}_\mathcal{D}^\top\mathbf{C}_\mathcal{DD}^{-1}\mathbf{\Omega}_\mathcal{ID}^\top(\mathbf{\Sigma}_\mathcal{II}+\mathbf{\Psi}_\mathcal{II})^{-1}\mathbf{\Omega}_\mathcal{ID}\mathbf{C}_\mathcal{DD}^{-1}\mathbf{y}_\mathcal{D}\vspace{1mm}\\
			\displaystyle-\mathrm{Tr}[\mathbf{C}_\mathcal{DD}^{-1}\mathbf{\Upsilon}_\mathcal{DD}]+\mathrm{Tr}[\mathbf{\Sigma}_\mathcal{II}^{-1}\mathbf{\Psi}_\mathcal{II}]-\log|\mathbf{\Sigma}_\mathcal{II}+\mathbf{\Psi}_\mathcal{II}|\vspace{1mm}\\
			\displaystyle-\boldsymbol{\nu}^\top\boldsymbol{\nu}-\mathrm{Tr}[\mathbf{\Xi}]+\log|\mathbf{\Xi}| -\alpha^2-\beta +\log \beta\Big)+\mathrm{const}\ .
		\end{array} \vspace{-1mm}
		\label{eq:TitsiasL(q)}
	\end{equation}

	\vspace{-5mm}
	\subsection{Derivation of $\mathbf{\Omega}_\mathcal{ID}$, $\mathbf{\Psi}_\mathcal{II}$, and $\mathbf{\Upsilon}_\mathcal{DD}$}
	\label{Derivation of Omega, Psi and Upsilon}	
	Let $\mathbf{\Omega}_\mathcal{ID}\triangleq (\omega_{\mathbf{z}\mathbf{x}})_{\mathbf{z}\in\mathcal{I},\mathbf{x}\in\mathcal{D}}$, $\mathbf{z}\triangleq (z_1,\ldots,z_d)^\top$, and $\mathbf{x}\triangleq (x_1,\ldots,x_d)^\top$. Since $\mathbf{\Omega}_\mathcal{ID}\hspace{-0mm}\triangleq\hspace{-0mm}\mathbb{E}_{q(\mathbf{\Lambda},\sigma_f)}\hspace{-0mm}(\mathbf{K}_{\mathcal{ID}})$, 
	\begin{equation*}
		\hspace{-1.7mm}
		\begin{array}{l}
			\omega_{\mathbf{z}\mathbf{x}}\\
			\displaystyle=\int q(\sigma_f)\ q(\mathbf{\Lambda})\ \mathrm{cov}[s_\mathbf{z},f_\mathbf{x}]\ \mathrm{d}\mathbf{\Lambda}\ \mathrm{d}\sigma_f\\
			\displaystyle=\int q(\sigma_f)\left(\int q(\mathbf{\Lambda})\ \sigma_f \exp\hspace{-0.7mm}\left(\hspace{-0.7mm}-\dfrac{1}{2}\sum_{k=1}^{d}(\lambda_k x_{k}-{z}_{k})^2\hspace{-0.7mm}\right) \hspace{-0.7mm}\mathrm{d}\mathbf{\Lambda}\hspace{-0.7mm}\right)\hspace{-0.7mm}\mathrm{d}\sigma_f \vspace{1mm}\\
			\displaystyle=\int q(\sigma_f)\ \sigma_f\\
			\displaystyle\quad \prod_{k=1}^{d}\int \exp\hspace{-0.7mm}\left(-\dfrac{1}{2}\sum_{k=1}^{d}(\lambda_k x_{k}-{z}_{k})^2\right)\mathcal{N}(\lambda_k|\nu_k,\xi_k)\ \mathrm{d}\lambda_k\ \mathrm{d}\sigma_f \\
			\displaystyle =\int q(\sigma_f)\ \sigma_f \prod_{k=1}^{d}{(\xi_kx_{k}^2+1)^{-\frac{1}{2}}}\exp\left(-\frac{(x_{k}\nu_k-z_{k})^2}{2(\xi_kx_{k}^2+1)}\right)\mathrm{d}\sigma_f \\
			\displaystyle =\alpha \prod_{k=1}^{d}{(\xi_kx_{k}^2+1)^{-\frac{1}{2}}}\exp\left(-\frac{(x_{k}\nu_k-z_{k})^2}{2(\xi_kx_{k}^2+1)}\right).
		\end{array}
	\end{equation*}
\noindent
	Since $\mathbf{C}_\mathcal{DD}$ is a block-diagonal matrix constructed using the $B$ blocks $\mathbf{C}_{\mathcal{D}_i\mathcal{D}_i}$ for $i=1,\ldots,B$,
	$\mathbf{C}^{-1}_\mathcal{DD}$ is also a block-diagonal matrix constructed using the $B$ blocks $\mathbf{C}^{-1}_{\mathcal{D}_i\mathcal{D}_i}$ for $i=1,\ldots,B$.
	Let $\mathbf{C}^{-1}_{\mathcal{D}_i\mathcal{D}_i}\triangleq (c^{i}_{\mathbf{x}\mathbf{x}'})_{\mathbf{x},\mathbf{x}'\in\mathcal{D}_i}$.
	Let $\mathbf{\Psi}_\mathcal{II}\triangleq (\psi_{\mathbf{z}\mathbf{z}'})_{\mathbf{z},\mathbf{z}'\in\mathcal{I}}$, $\mathbf{z}'\triangleq (z'_1,\ldots,z'_d)^\top$, and $\mathbf{x}'\triangleq (x'_1,\ldots,x'_d)^\top$.
	Since $\mathbf{\Psi}_\mathcal{II}\triangleq\mathbb{E}_{q(\mathbf{\Lambda},\sigma_f)}(\mathbf{K}_{\mathcal{ID}}\mathbf{C}^{-1}_\mathcal{DD}\mathbf{K}_{\mathcal{DI}})$,
	\vspace{-4mm}
	\begin{equation*}
		\hspace{-1.7mm}
		\begin{array}{l}
			\psi_{\mathbf{z}\mathbf{z}'} \vspace{-1mm}\\
			\displaystyle=\hspace{-1mm}\int \hspace{-1mm} q(\sigma_f)\hspace{-0.5mm} \sum_{i=1}^{B}\hspace{-0.5mm}\sum_{\mathbf{x},\mathbf{x}'\in\mathcal{D}_i}\hspace{-0.5mm}\mathbb{E}_{\mathbf{\Lambda}}\left[\mathrm{cov}[s_\mathbf{z},f_\mathbf{x}]\ c^{i}_{\mathbf{x}\mathbf{x}'}\ \mathrm{cov}[f_{\mathbf{x}'},s_{\mathbf{z}'}] \right]\hspace{-0.5mm} \mathrm{d}\sigma_f \\
			\displaystyle=\hspace{-1mm}\int\hspace{-1mm} q(\sigma_f) \sum_{i=1}^{B}\hspace{-0.5mm}\sum_{\mathbf{x},\mathbf{x}'\in\mathcal{D}_i}\hspace{-0.5mm}c^{i}_{\mathbf{x}\mathbf{x}'}\ \mathbb{E}_{\mathbf{\Lambda}}\left[\mathrm{cov}[s_\mathbf{z},f_\mathbf{x}]\ \mathrm{cov}[f_{\mathbf{x}'},s_{\mathbf{z}'}] \right]\hspace{-0.5mm} \mathrm{d}\sigma_f \\
			\displaystyle=\int q(\sigma_f)\sum_{i=1}^{B}\sum_{\mathbf{x},\mathbf{x}'\in\mathcal{D}_i}\sigma_f^2\ c^{i}_{\mathbf{x}\mathbf{x}'}\prod_{k=1}^{d}\Bigg\{(\xi_k(x_{k}^2+x_{k}^{\prime2})+1)^{-\frac{1}{2}} \vspace{-1mm}\\
			\quad\exp\left(-\frac{\xi_k(z'_{k}x_{k}-z_{k}x'_{k})^2+(x_{k}\nu_k-z_{k})^2+(x'_{k}\nu_k-z'_{k})^2}{2(\xi_k(x_{k}^2+x_{k}^{\prime 2})+1)}\right)\Bigg\}\ \mathrm{d}\sigma_f \\
			\displaystyle= \sum_{i=1}^{B}\sum_{\mathbf{x},\mathbf{x}'\in\mathcal{D}_i}(\beta+\alpha^2)\ c^{i}_{\mathbf{x}\mathbf{x}'}\prod_{k=1}^{d}\Bigg\{(\xi_k(x_{k}^2+x_{k}^{\prime2})+1)^{-\frac{1}{2}} \vspace{-1mm}\\
			\qquad\quad\exp\left(-\frac{\xi_k(z'_{k}x_{k}-z_{k}x'_{k})^2+(x_{k}\nu_k-z_{k})^2+(x'_{k}\nu_k-z'_{k})^2}{2(\xi_k(x_{k}^2+x_{k}^{\prime 2})+1)}\right)\Bigg\}\ . 
		\end{array}
	\end{equation*}
	\vspace{-0mm}
	Let $\mathbf{\Upsilon}_\mathcal{DD}\triangleq (\gamma_{\mathbf{x}\mathbf{x}'})_{\mathbf{x},\mathbf{x}'\in\mathcal{D}}$. Since $\mathbf{\Upsilon}_\mathcal{DD}\triangleq\mathbb{E}_{q(\mathbf{\Lambda},\sigma_f)}(\mathbf{K}_{\mathcal{DD}})$,
	\vspace{-2mm}
	\begin{equation*}
		\hspace{-1.7mm}
		\begin{array}{l}
			\gamma_{\mathbf{x}\mathbf{x}'}\\
			\displaystyle=\int q(\sigma_f)q(\mathbf{\Lambda})\ k_{\mathbf{x}\mathbf{x}'}\ \mathrm{d}\mathbf{\Lambda}\ \mathrm{d}\sigma_f \vspace{-1mm}\\
			\displaystyle=\hspace{-0.5mm}\int\hspace{-0.5mm} q(\sigma_f)\hspace{-0.5mm}\int \hspace{-0.5mm}q(\mathbf{\Lambda})\sigma_f^2 \exp\left(-\dfrac{1}{2}\sum_{k=1}^{d}\lambda_k^2(x_{k}-x'_{k})^2\right)\hspace{-0.5mm} \mathrm{d}\mathbf{\Lambda} \mathrm{d}\sigma_f \vspace{-1mm}\\
			\displaystyle=\int q(\sigma_f)\ \sigma_f^2\\
			\displaystyle\prod_{k=1}^{d}\int \exp\hspace{-1mm}\left(-\dfrac{1}{2}\sum_{k=1}^{d}\lambda_k^2(x_{k}-x'_{k})^2\right)\mathcal{N}(\lambda_k|\nu_k,\xi_k)\ \mathrm{d}\lambda_k\  \mathrm{d}\sigma_f   \vspace{1mm}\\
			\displaystyle = \int q(\sigma_f)\ \sigma_f^2\\
			\displaystyle\hspace{-0.5mm}\prod_{k=1}^{d}\hspace{-0.5mm}{(\xi_k(x_{k}-x'_{k})^2+1)^{-\frac{1}{2}}}\exp\hspace{-0.7mm}\left(\hspace{-0.7mm}-\frac{\nu_k^2(x_{k}-x'_{k})^2}{2(\xi_k(x_{k}-x'_{k})^2+1)}\right)\hspace{-0.5mm}\mathrm{d}\sigma_f \\
			=\displaystyle (\beta+\alpha^2)\\
			\displaystyle\prod_{k=1}^{d}{(\xi_k(x_{k}-x'_{k})^2+1)^{-\frac{1}{2}}}\exp\hspace{-0.7mm}\left(\hspace{-0.7mm}-\frac{\nu_k^2(x_{k}-x'_{k})^2}{2(\xi_k(x_{k}-x'_{k})^2+1)}\right).
		\end{array}
	\end{equation*}
\vspace{-5mm}
\subsection{Proof of Theorem~\ref{thm2}}
	\label{A.3}
	\vspace{-2mm}
	Let
	\vspace{-1mm}
		\begin{equation}
			\hspace{-2.7mm}
			\begin{array}{l}
				\displaystyle\dfrac{\partial\widetilde{\mathcal{L}}}{\partial\mathbf{m}}\hspace{-0.5mm}\triangleq\hspace{-0.5mm}\dfrac{B}{|\mathcal{S}|}\hspace{-0.5mm}\sum_{s\in\mathcal{S}}\hspace{-0.5mm}\dfrac{\partial{\mathcal{L}}_s}{\partial\mathbf{m}}\hspace{-0.5mm}-\hspace{-0.5mm}\mathbf{\Sigma}_\mathcal{II}^{-1}\mathbf{m}\ , 
				\displaystyle\dfrac{\partial\widetilde{\mathcal{L}}}{\partial\mathbf{\Xi}}\hspace{-0.5mm}\triangleq\hspace{-0.5mm}\dfrac{B}{|\mathcal{S}|}\hspace{-0.5mm}\sum_{s\in\mathcal{S}}\hspace{-0.5mm}\dfrac{\partial{\mathcal{L}}_s}{\partial\mathbf{\Xi}}\hspace{-1mm}-\hspace{-0.5mm}\dfrac{1}{2}\mathbf{I}\hspace{-0.5mm}+\hspace{-0.5mm}\dfrac{1}{2}\mathbf{\Xi}^{-1},\vspace{1mm}\\
				\displaystyle\dfrac{\partial{\widetilde{\mathcal{L}}}}{\partial\mathbf{S}}\hspace{-0.5mm}\triangleq\hspace{-0.5mm}\dfrac{B}{|\mathcal{S}|}\hspace{-0.5mm}\sum_{s\in\mathcal{S}}\hspace{-0.5mm}\dfrac{\partial{\mathcal{L}}_s}{\partial\mathbf{S}}\hspace{-0.5mm}+\hspace{-0.5mm}\dfrac{1}{2}\mathbf{S}^{-1}\hspace{-1mm}-\hspace{-0.5mm}\dfrac{1}{2}\mathbf{\Sigma}_\mathcal{II}^{-1}\ , \ 
				\dfrac{\partial\widetilde{\mathcal{L}}}{\partial\boldsymbol{\nu}}\hspace{-0.5mm}\triangleq\hspace{-0.5mm}\dfrac{B}{|\mathcal{S}|}\hspace{-0.5mm}\sum_{s\in\mathcal{S}}\hspace{-0.5mm}\dfrac{\partial{\mathcal{L}}_s}{\partial\boldsymbol{\nu}}\hspace{-0.5mm}-\hspace{-0.5mm}\boldsymbol{\nu},
				\vspace{1mm} \\
				\displaystyle\dfrac{\partial\widetilde{\mathcal{L}}}{\partial \alpha}\hspace{-0.5mm}\triangleq\hspace{-0.5mm}\dfrac{B}{|\mathcal{S}|}\hspace{-0.5mm}\sum_{s\in\mathcal{S}}\dfrac{\partial{\mathcal{L}}_s}{\partial \alpha}\hspace{-0.5mm}-\hspace{-0.5mm}\alpha\ , \ 
				\dfrac{\partial\widetilde{\mathcal{L}}}{\partial \beta}\hspace{-0.5mm}\triangleq\hspace{-0.5mm}\dfrac{B}{|\mathcal{S}|}\hspace{-0.5mm}\sum_{s\in\mathcal{S}}\hspace{-0.5mm}\dfrac{\partial{\mathcal{L}}_s}{\partial \beta}\hspace{-0.5mm}-\hspace{-0.5mm}\dfrac{\beta-1}{2\beta}
			\end{array}
			\label{zoo}
		\end{equation}
		where
		\begin{equation*}
			\hspace{-1.7mm}
			\begin{array}{l}
				\displaystyle\dfrac{\partial{\mathcal{L}}_s}{\partial\mathbf{m}}=\mathbf{\Sigma}_\mathcal{II}^{-1}\mathbf{\Omega}_{\mathcal{I}\mathcal{D}_s}\mathbf{C}_{\mathcal{D}_s\mathcal{D}_s}^{-1}\mathbf{y}_{\mathcal{D}_s}-\mathbf{\Sigma}_\mathcal{II}^{-1}\mathbf{\Psi}_\mathcal{II}^{s}\mathbf{\Sigma}_\mathcal{II}^{-1}\mathbf{m}\ ,\vspace{2mm}\\
				\displaystyle\dfrac{\partial{\mathcal{L}}_s}{\partial\mathbf{S}}=-\dfrac{1}{2}\mathbf{\Sigma}_\mathcal{II}^{-1}\mathbf{\Psi}_\mathcal{II}^{s}\mathbf{\Sigma}_\mathcal{II}^{-1}\ ,\vspace{2mm}\\
				\displaystyle\dfrac{\partial{\mathcal{L}}_s}{\partial\boldsymbol{\nu}}=\mathbf{m}^\top\mathbf{\Sigma}_\mathcal{II}^{-1}\dfrac{\partial\mathbf{\Omega}_{\mathcal{I}\mathcal{D}_s}}{\partial\boldsymbol{\nu}}\mathbf{C}_{\mathcal{D}_s\mathcal{D}_s}^{-1}\mathbf{y}_{\mathcal{D}_s}\hspace{-1mm}-\hspace{-0.5mm}\frac{1}{2}\mathbf{m}^\top\mathbf{\Sigma}_\mathcal{II}^{-1}\dfrac{\partial\mathbf{\Psi}_\mathcal{II}^{s}}{\partial\boldsymbol{\nu}}\mathbf{\Sigma}_\mathcal{II}^{-1}\mathbf{m}\vspace{1mm}\\
				\displaystyle\qquad\quad\ -\frac{1}{2}\mathrm{Tr}\Big[\mathbf{S}\mathbf{\Sigma}_\mathcal{II}^{-1}\dfrac{\partial\mathbf{\Psi}_\mathcal{II}^{s}}{\partial\boldsymbol{\nu}}\mathbf{\Sigma}_\mathcal{II}^{-1}\Big]-\frac{1}{2}\mathrm{Tr}\Big[\mathbf{C}_{\mathcal{D}_s\mathcal{D}_s}^{-1}\dfrac{\partial\mathbf{\Upsilon}_{\mathcal{D}_s\mathcal{D}_s}}{\partial\boldsymbol{\nu}}\Big]\vspace{1mm}\\
				\displaystyle\qquad\quad\ + \frac{1}{2}\mathrm{Tr}\Big[\mathbf{\Sigma}_\mathcal{II}^{-1}\dfrac{\partial\mathbf{\Psi}_\mathcal{II}^{s}}{\partial\boldsymbol{\nu}}\Big]\ ,\vspace{1mm}\\
				\displaystyle\dfrac{\partial{\mathcal{L}}_s}{\partial\mathbf{\Xi}}=\mathbf{m}^\top\mathbf{\Sigma}_\mathcal{II}^{-1}\dfrac{\partial\mathbf{\Omega}_{\mathcal{I}\mathcal{D}_s}}{\partial\mathbf{\Xi}}\mathbf{C}_{\mathcal{D}_s\mathcal{D}_s}^{-1}\mathbf{y}_{\mathcal{D}_s}\hspace{-1mm}-\hspace{-1mm}\frac{1}{2}\mathbf{m}^\top\mathbf{\Sigma}_\mathcal{II}^{-1}\dfrac{\partial\mathbf{\Psi}_\mathcal{II}^{s}}{\partial\mathbf{\Xi}}\mathbf{\Sigma}_\mathcal{II}^{-1}\mathbf{m}\vspace{1mm}\\
				\displaystyle\qquad\quad\ -\frac{1}{2}\mathrm{Tr}\Big[\mathbf{S}\mathbf{\Sigma}_\mathcal{II}^{-1}\dfrac{\partial\mathbf{\Psi}_\mathcal{II}^{s}}{\partial\mathbf{\Xi}}\mathbf{\Sigma}_\mathcal{II}^{-1}\Big]-\frac{1}{2}\mathrm{Tr}\Big[\mathbf{C}_{\mathcal{D}_s\mathcal{D}_s}^{-1}\dfrac{\partial\mathbf{\Upsilon}_{\mathcal{D}_s\mathcal{D}_s}}{\partial\mathbf{\Xi}}\Big]\vspace{1mm}\\
				\displaystyle\qquad\quad\ +\frac{1}{2}\mathrm{Tr}\Big[\mathbf{\Sigma}_\mathcal{II}^{-1}\dfrac{\partial\mathbf{\Psi}_\mathcal{II}^{s}}{\partial\mathbf{\Xi}}\Big]\ ,\vspace{2mm}\\
				\displaystyle\dfrac{\partial{\mathcal{L}}_s}{\partial \alpha}=\mathbf{m}^\top\mathbf{\Sigma}_\mathcal{II}^{-1}\dfrac{\partial\mathbf{\Omega}_{\mathcal{I}\mathcal{D}_s}}{\partial \alpha}\mathbf{C}_{\mathcal{D}_s\mathcal{D}_s}^{-1}\mathbf{y}_{\mathcal{D}_s}\hspace{-1mm}-\hspace{-1mm}\frac{1}{2}\mathbf{m}^\top\mathbf{\Sigma}_\mathcal{II}^{-1}\dfrac{\partial\mathbf{\Psi}_\mathcal{II}^{s}}{\partial \alpha}\mathbf{\Sigma}_\mathcal{II}^{-1}\mathbf{m}\vspace{1mm}\\ \displaystyle\qquad\quad\;-\frac{1}{2}\mathrm{Tr}\Big[\mathbf{S}\mathbf{\Sigma}_\mathcal{II}^{-1}\dfrac{\partial\mathbf{\Psi}_\mathcal{II}^{s}}{\partial \alpha}\mathbf{\Sigma}_\mathcal{II}^{-1}\Big] 
				-\frac{1}{2}\mathrm{Tr}\Big[\mathbf{C}_{\mathcal{D}_s\mathcal{D}_s}^{-1}\dfrac{\partial\mathbf{\Upsilon}_{\mathcal{D}_s\mathcal{D}_s}}{\partial \alpha}\Big]\vspace{1mm} \\
				\displaystyle\qquad\quad\;+\frac{1}{2}\mathrm{Tr}\Big[\mathbf{\Sigma}_\mathcal{II}^{-1}\dfrac{\partial\mathbf{\Psi}_\mathcal{II}^{s}}{\partial \alpha}\Big]\vspace{2mm}\ ,\\
				\displaystyle\dfrac{\partial{\mathcal{L}}_s}{\partial \beta}=\mathbf{m}^\top\mathbf{\Sigma}_\mathcal{II}^{-1}\dfrac{\partial\mathbf{\Omega}_{\mathcal{I}\mathcal{D}_s}}{\partial \beta}\mathbf{C}_{\mathcal{D}_s\mathcal{D}_s}^{-1}\mathbf{y}_{\mathcal{D}_s}\hspace{-1mm}-\hspace{-1mm}\frac{1}{2}\mathbf{m}^\top\mathbf{\Sigma}_\mathcal{II}^{-1}\dfrac{\partial\mathbf{\Psi}_\mathcal{II}^{s}}{\partial \beta}\mathbf{\Sigma}_\mathcal{II}^{-1}\mathbf{m}\vspace{1mm} \\
				\displaystyle\qquad\quad\;-\frac{1}{2}\mathrm{Tr}\Big[\mathbf{S}\mathbf{\Sigma}_\mathcal{II}^{-1}\dfrac{\partial\mathbf{\Psi}_\mathcal{II}^{s}}{\partial \beta}\mathbf{\Sigma}_\mathcal{II}^{-1}\Big]- \frac{1}{2}\mathrm{Tr}\Big[\mathbf{C}_{\mathcal{D}_s\mathcal{D}_s}^{-1}\dfrac{\partial\mathbf{\Upsilon}_{\mathcal{D}_s\mathcal{D}_s}}{\partial \beta}\Big]\vspace{1mm}\\ \displaystyle\qquad\quad\;+\frac{1}{2}\mathrm{Tr}\Big[\mathbf{\Sigma}_\mathcal{II}^{-1}\dfrac{\partial\mathbf{\Psi}_\mathcal{II}^{s}}{\partial \beta}\Big]\ ,
			\end{array}
		\end{equation*}	
and the closed-form expressions of $\partial\mathbf{\Omega}_{\mathcal{I}\mathcal{D}_s}/\partial\boldsymbol{\nu}$, $\partial\mathbf{\Psi}_\mathcal{II}^{s}/\partial\boldsymbol{\nu}$, $\partial\mathbf{\Upsilon}_{\mathcal{D}_s\mathcal{D}_s}/\partial\boldsymbol{\nu}$, $\partial\mathbf{\Omega}_{\mathcal{I}\mathcal{D}_s}/\partial\mathbf{\Xi}$, $\partial\mathbf{\Psi}_\mathcal{II}^{s}/\partial\mathbf{\Xi}$,  $\partial\mathbf{\Upsilon}_{\mathcal{D}_s\mathcal{D}_s}/\partial\mathbf{\Xi}$ $\partial\mathbf{\Omega}_{\mathcal{I}\mathcal{D}_s}/\partial \alpha$, $\partial\mathbf{\Psi}_\mathcal{II}^{s}/\partial \alpha$, $\partial\mathbf{\Upsilon}_{\mathcal{D}_s\mathcal{D}_s}/\partial \alpha$, $\partial\mathbf{\Omega}_{\mathcal{I}\mathcal{D}_s}/\partial \beta$, $\partial\mathbf{\Psi}_\mathcal{II}^{s}/\partial \beta$, and $\partial\mathbf{\Upsilon}_{\mathcal{D}_s\mathcal{D}_s}/\partial \beta$ are given in Appendix~\ref{argghh}.

	Then, since
	\begin{equation*}
		\begin{array}{l}
			\displaystyle\mathbb{E}[ \mathbf{\Sigma}_\mathcal{II}^{-1}\mathbf{\Omega}_{\mathcal{I}\mathcal{D}_s}\mathbf{C}_{\mathcal{D}_s\mathcal{D}_s}^{-1}\mathbf{y}_{\mathcal{D}_s}-\mathbf{\Sigma}_\mathcal{II}^{-1}\mathbf{\Psi}_\mathcal{II}^{s}\mathbf{\Sigma}_\mathcal{II}^{-1}\mathbf{m}] \\
			\displaystyle=\sum_{i=1}^{B}p(s=i)(\mathbf{\Sigma}_\mathcal{II}^{-1}\mathbf{\Omega}_{\mathcal{I}\mathcal{D}_i}\mathbf{C}_{\mathcal{D}_i\mathcal{D}_i}^{-1}\mathbf{y}_{\mathcal{D}_i}-\mathbf{\Sigma}_\mathcal{II}^{-1}\mathbf{\Psi}_\mathcal{II}^{i}\mathbf{\Sigma}_\mathcal{II}^{-1}\mathbf{m}) \\
			\displaystyle=\sum_{i=1}^{B}\dfrac{1}{B}(\mathbf{\Sigma}_\mathcal{II}^{-1}\mathbf{\Omega}_{\mathcal{I}\mathcal{D}_i}\mathbf{C}_{\mathcal{D}_i\mathcal{D}_i}^{-1}\mathbf{y}_{\mathcal{D}_i}-\mathbf{\Sigma}_\mathcal{II}^{-1}\mathbf{\Psi}_\mathcal{II}^{i}\mathbf{\Sigma}_\mathcal{II}^{-1}\mathbf{m}) \\
			\displaystyle=\dfrac{1}{B}\sum_{i=1}^{B}\mathbf{\Sigma}_\mathcal{II}^{-1}\mathbf{\Omega}_{\mathcal{I}\mathcal{D}_i}\mathbf{C}_{\mathcal{D}_i\mathcal{D}_i}^{-1}\mathbf{y}_{\mathcal{D}_i}-\mathbf{\Sigma}_\mathcal{II}^{-1}\mathbf{\Psi}_\mathcal{II}^{i}\mathbf{\Sigma}_\mathcal{II}^{-1}\mathbf{m}\ ,
		\end{array}
	\end{equation*}
	\begin{equation*}
		\begin{array}{l}
			\displaystyle\mathbb{E}\left[\sum_{s\in\mathcal{S}} \mathbf{\Sigma}_\mathcal{II}^{-1}\mathbf{\Omega}_{\mathcal{I}\mathcal{D}_s}\mathbf{C}_{\mathcal{D}_s\mathcal{D}_s}^{-1}\mathbf{y}_{\mathcal{D}_s}-\mathbf{\Sigma}_\mathcal{II}^{-1}\mathbf{\Psi}_\mathcal{II}^{s}\mathbf{\Sigma}_\mathcal{II}^{-1}\mathbf{m}\right]\\
			\displaystyle=\dfrac{|\mathcal{S}|}{B}\sum_{i=1}^{B}\mathbf{\Sigma}_\mathcal{II}^{-1}\mathbf{\Omega}_{\mathcal{I}\mathcal{D}_i}\mathbf{C}_{\mathcal{D}_i\mathcal{D}_i}^{-1}\mathbf{y}_{\mathcal{D}_i}-\mathbf{\Sigma}_\mathcal{II}^{-1}\mathbf{\Psi}_\mathcal{II}^{i}\mathbf{\Sigma}_\mathcal{II}^{-1}\mathbf{m}\ .
		\end{array}
	\end{equation*}
	It follows that $\mathbb{E}[\partial\widetilde{\mathcal{L}}/\partial\mathbf{m}]=\partial{\mathcal{L}}/\partial\mathbf{m}$. The proofs for $\mathbb{E}[\partial\widetilde{\mathcal{L}}/\partial\mathbf{S}]=\partial{\mathcal{L}}/\partial\mathbf{S}$, 
	$\mathbb{E}[\partial\widetilde{\mathcal{L}}/\partial\boldsymbol{\nu}]=\partial{\mathcal{L}}/\partial\boldsymbol{\nu}$, 
	$\mathbb{E}[\partial\widetilde{\mathcal{L}}/\partial\mathbf{\Xi}]=\partial{\mathcal{L}}/\partial\mathbf{\Xi}$, 	$\mathbb{E}[\partial\widetilde{\mathcal{L}}/\partial   \alpha]=\partial{\mathcal{L}}/\partial \alpha$, and $\mathbb{E}[\partial\widetilde{\mathcal{L}}/\partial \beta]=\partial{\mathcal{L}}/\partial \beta$ follow a similar procedure as the above.

	\subsection{Derivatives of $\mathbf{\Omega}_{\mathcal{I}\mathcal{D}_s}$, $\mathbf{\Psi}^s_\mathcal{II}$, and $\mathbf{\Upsilon}_{\mathcal{D}_s\mathcal{D}_s}$ with respect to $\boldsymbol{\nu}$, $\mathbf{\Xi}$, $\alpha$, and $\beta$}
	\label{argghh}
	\hspace{2mm}Note that $\boldsymbol{\nu}= (\nu_1,\ldots,\nu_d)^\top$ and $\mathbf{\Xi}=\mathrm{diag}[\xi_1,\ldots,\xi_d]^\top$, as defined previously in Section~\ref{Variational Inference of the Bayesian DTC}.
	
	From Appendix~\ref{Derivation of Omega, Psi and Upsilon},
	\begin{equation*}
		\omega_\mathbf{zx}=\alpha \prod_{k=1}^{d}{(\xi_kx_{k}^2+1)^{-\frac{1}{2}}}\exp\left(-\frac{(x_{k}\nu_k-z_{k})^2}{2(\xi_kx_{k}^2+1)}\right)
	\end{equation*}
	where $\mathbf{z} = (z_1,\ldots,z_d)^\top$ and $\mathbf{x} = (x_1,\ldots,x_d)^\top$.	
	The partial derivative of $\omega_\mathbf{zx}$ with respect to $\boldsymbol{\nu}$, $\mathbf{\Xi}$, $\alpha$, and $\beta$ can be derived as follows:
	\begin{equation*}
		\begin{array}{rcl}
			\displaystyle\dfrac{\partial\omega_\mathbf{zx}}{\partial\nu_i}&\hspace{-2.4mm}=&\hspace{-2.4mm} \displaystyle\alpha \prod_{k=1}^{d}{(\xi_kx_{k}^2+1)^{-\frac{1}{2}}}\exp\left(-\frac{(x_{k}\nu_k-z_{k})^2}{2(\xi_kx_{k}^2+1)}\right)\\
			&&\displaystyle\quad\times\left(-\frac{(x_{i}\nu_i-z_{i})^2}{2(\xi_ix_{i}^2+1)}\right)^\prime\\
			&\hspace{-2.4mm}=&\hspace{-2.4mm} \displaystyle\alpha \prod_{k=1}^{d}{(\xi_kx_{k}^2+1)^{-\frac{1}{2}}}\exp\left(-\frac{(x_{k}\nu_k-z_{k})^2}{2(\xi_kx_{k}^2+1)}\right)\\
			&&\displaystyle\quad\times\left(\dfrac{-\nu_ix_{i}^2+z_{i}x_{i}}{\xi_ix_{i}^2+1}\right),
		\end{array}
	\end{equation*}
	\begin{equation*}
		\begin{array}{rcl}
			\displaystyle\dfrac{\partial\omega_\mathbf{zx}}{\partial\xi_i}&\hspace{-2.4mm}=&\hspace{-2.4mm}
			\displaystyle\alpha \prod_{k\neq i}{(\xi_kx_{k}^2+1)^{-\frac{1}{2}}}\exp\left(-\frac{(x_{k}\nu_k-z_{k})^2}{2(\xi_kx_{k}^2+1)}\right)\\
			&&\displaystyle\quad\times\Bigg({(\xi_ix_{i}^2+1)^{-\frac{1}{2}}}\exp\left(-\frac{(x_{i}\nu_i-z_{i})^2}{2(\xi_ix_{i}^2+1)}\right)\Bigg)^\prime \\
			&\hspace{-2.4mm}=&\hspace{-2.4mm}\displaystyle\alpha \prod_{k\neq i}{(\xi_kx_{k}^2+1)^{-\frac{1}{2}}}\exp\left(-\frac{(x_{k}\nu_k-z_{k})^2}{2(\xi_kx_{k}^2+1)}\right)\\
			&&\displaystyle\quad\times\Bigg\{\Big({(\xi_ix_{i}^2+1)^{-\frac{1}{2}}}\Big)^\prime\exp\left(-\frac{(x_{i}\nu_i-z_{i})^2}{2(\xi_ix_{i}^2+1)}\right)\\
			&&\displaystyle\qquad+{(\xi_ix_{i}^2+1)^{-\frac{1}{2}}}\Bigg (\exp\left(-\frac{(x_{i}\nu_i-z_{i})^2}{2(\xi_ix_{i}^2+1)}\right)\Bigg)^\prime\Bigg\} \\
			&\hspace{-2.4mm}=&\hspace{-2.4mm}\displaystyle\alpha \prod^d_{k=1}{(\xi_kx_{k}^2+1)^{-\frac{1}{2}}}\exp\left(-\frac{(x_{k}\nu_k-z_{k})^2}{2(\xi_kx_{k}^2+1)}\right)\\
			&&\displaystyle\quad\times\left(-\dfrac{x_{i}^2}{2(\xi_ix_{i}^2+1)}+\dfrac{x_{i}^2(x_{i}\nu_i-z_{i})^2}{2(\xi_ix_{i}^2+1)^2}\right) ,
		\end{array}
	\end{equation*}
	\begin{equation*}
	\begin{array}{l}
	\displaystyle\dfrac{\partial\omega_\mathbf{zx}}{\partial \alpha}= \prod_{k=1}^{d}{(\xi_kx_{k}^2+1)^{-\frac{1}{2}}}\exp\left(-\frac{(x_{k}\nu_k-z_{k})^2}{2(\xi_kx_{k}^2+1)}\right) ,\\
	\displaystyle\dfrac{\partial\omega_\mathbf{zx}}{\partial \beta}=0\ .
	\end{array}
	\end{equation*}
	From Appendix~\ref{Derivation of Omega, Psi and Upsilon},
	\begin{equation*}
		\hspace{-1.7mm}
		\begin{array}{rcl}
			\psi_{\mathbf{z}\mathbf{z}^\prime}&\hspace{-2.4mm}=&\hspace{-2.4mm}
				\displaystyle\sum_{i=1}^{B}\sum_{\mathbf{x},\mathbf{x}'\in\mathcal{D}_i}(\beta+\alpha^2)\ c^{i}_{\mathbf{x}\mathbf{x}'}\prod_{k=1}^{d}\Bigg\{\hspace{-0.5mm}(\xi_k(x_{k}^2+x_{k}^{\prime2})+1)^{-\frac{1}{2}}\\
			&&\hspace{-2.4mm}\exp\left(-\frac{\xi_k(z'_{k}x_{k}-z_{k}x'_{k})^2+(x_{k}\nu_k-z_{k})^2+(x'_{k}\nu_k-z'_{k})^2}{2(\xi_k(x_{k}^2+x_{k}^{\prime 2})+1)}\right)\Bigg\}
		\end{array}
	\end{equation*}
	where $\mathbf{z}^\prime \triangleq (z^\prime_1,\ldots,z^\prime_d)^\top$ and $\mathbf{x}^\prime \triangleq (x^\prime_1,\ldots,x^\prime_d)^\top$.
	The partial derivative of $\psi_\mathbf{xx^\prime}$ with respect to $\boldsymbol{\nu}$, $\mathbf{\Xi}$, $\alpha$, and $\beta$ can be derived as follows:
	%
	\begin{equation*}
		\hspace{-1.7mm}
		\begin{array}{l}
			\displaystyle\dfrac{\partial\psi_\mathbf{zz^\prime}}{\partial\nu_i}\\
			=\displaystyle\sum_{j=1}^{B}\sum_{\mathbf{x},\mathbf{x}^\prime\in\mathcal{D}_j}\hspace{-0.5mm}\Biggl[\hspace{-0.5mm}\left(\beta+\alpha^2\right) c^{j}_{\mathbf{x}\mathbf{x}^\prime}\prod_{k=1}^{d}\Bigg\{(\xi_k(x_{k}^2+x_{k}^{\prime2})+1)^{-\frac{1}{2}}\\
			\quad\exp\left(-\frac{\xi_k(z^\prime_{k}x_{k}-z_{k}x^\prime_{k})^2+(x_{k}\nu_k-z_{k})^2+(x^\prime_{k}\nu_k-z^\prime_{k})^2}{2(\xi_k(x_{k}^2+x_{k}^{\prime 2})+1)}\right)\hspace{-0.5mm}\Bigg\}\\
			\quad\times\left(-\frac{\xi_i(z^\prime_{i}x_{i}-z_{i}x^\prime_{i})^2+(x_{i}\nu_i-z_{i})^2+(x^\prime_{i}\nu_i-z^\prime_{i})^2}{2(\xi_i(x_{i}^2+x_{i}^{\prime 2})+1)}\right)^\prime\Biggr]\\
			\displaystyle=\sum\limits_{j=1}^{B}\sum\limits_{\mathbf{x},\mathbf{x}^\prime\in\mathcal{D}_j}\hspace{-0.5mm}\Biggl[\hspace{-0.5mm}\left(\beta+\alpha^2\right) c^{j}_{\mathbf{x}\mathbf{x}^\prime}\prod_{k=1}^{d}\Bigg\{(\xi_k(x_{k}^2+x_{k}^{\prime2})+1)^{-\frac{1}{2}}\\
			\quad\exp\left(-\frac{\xi_k(z^\prime_{k}x_{k}-z_{k}x^\prime_{k})^2+(x_{k}\nu_k-z_{k})^2+(x^\prime_{k}\nu_k-z^\prime_{k})^2}{2(\xi_k(x_{k}^2+x_{k}^{\prime 2})+1)}\right)\hspace{-0.5mm}\Bigg\}\\
			\quad\times\left(-\dfrac{\nu_i(x_{i}^2+x_{i}^{\prime2})-(z_{i}x_{i}+z^\prime_{i}x_{i}^\prime)}{\xi_i(x_{i}^2+x_{i}^{\prime2})+1}\right)\Biggr],
		\end{array}
	\end{equation*}
	\begin{equation*}
		\hspace{-1.7mm}
		\begin{array}{l}
			\displaystyle\dfrac{\partial\psi_\mathbf{zz^\prime}}{\partial\xi_i}\\
			\displaystyle=\sum\limits_{j=1}^{B}\sum\limits_{\mathbf{x},\mathbf{x}^\prime\in\mathcal{D}_j}\hspace{-0.5mm}\Biggl[\hspace{-0.5mm}\left(\beta+\alpha^2\right) c^{j}_{\mathbf{x}\mathbf{x}^\prime}\prod_{k\neq i}\Bigg\{(\xi_k(x_{k}^2+x_{k}^{\prime2})+1)^{-\frac{1}{2}}\\
			\quad\exp\left(-\frac{\xi_k(z^\prime_{k}x_{k}-z_{k}x^\prime_{k})^2+(x_{k}\nu_k-z_{k})^2+(x^\prime_{k}\nu_k-z^\prime_{k})^2}{2(\xi_k(x_{k}^2+x_{k}^{\prime 2})+1)}\right)\hspace{-0.5mm}\Bigg\}\\
			\quad\times\Biggl((\xi_i(x_{i}^2+x_{i}^{\prime2})+1)^{-\frac{1}{2}}\times\\
			\quad \exp\left(-\frac{\xi_i(z^\prime_{i}x_{i}-z_{i}x^\prime_{i})^2+(x_{i}\nu_i-z_{i})^2+(x^\prime_{i}\nu_i-z^\prime_{i})^2}{2(\xi_i(x_{i}^2+x_{i}^{\prime 2})+1)}\right)\Biggr)^\prime\Biggr]\\
			\displaystyle=\sum\limits_{j=1}^{B}\sum\limits_{\mathbf{x},\mathbf{x}^\prime\in\mathcal{D}_j}\hspace{-0.5mm}\Biggl[\hspace{-0.5mm}\left(\beta+\alpha^2\right) c^{j}_{\mathbf{x}\mathbf{x}^\prime}\prod_{k=1}^{d}\Bigg\{(\xi_k(x_{k}^2+x_{k}^{\prime2})+1)^{-\frac{1}{2}}\\
			\quad\exp\left(-\frac{\xi_k(z^\prime_{k}x_{k}-z_{k}x^\prime_{k})^2+(x_{k}\nu_k-z_{k})^2+(x^\prime_{k}\nu_k-z^\prime_{k})^2}{2(\xi_k(x_{k}^2+x_{k}^{\prime 2})+1)}\right)\hspace{-0.5mm}\Bigg\}\\
			\quad\times\Bigg(-\dfrac{x_{i}^2+x_{i}^{\prime2}}{2\big(\xi_i(x_{i}^2+x_{i}^{\prime2})+1\big)}\\
			\qquad+\dfrac{\Big(z_{i}x_{i}+z_{i}^\prime x_{i}^\prime-\nu_i(x_{i}^2+x_{i}^{\prime2})\Big)^2}{2\big(\xi_i(x_{i}^2+x_{i}^{\prime2})+1\big)^2}\Bigg)\Biggr],
		\end{array}
	\end{equation*}
	\begin{equation*}
	\begin{array}{l}
	\displaystyle\dfrac{\partial\psi_\mathbf{zz^\prime}}{\partial \alpha}=\hspace{-0.5mm}\sum\limits_{i=1}^{B}\hspace{-0.5mm}\sum\limits_{\mathbf{x},\mathbf{x}^\prime\in\mathcal{D}_i}\hspace{-0.5mm}2\alpha\ c^{i}_{\mathbf{x}\mathbf{x}^\prime}\hspace{-0.5mm}\prod_{k=1}^{d}\hspace{-0.5mm}\Bigg\{\hspace{-0.5mm}(\xi_k(x_{k}^2+x_{k}^{\prime2})+1)^{-\frac{1}{2}} \\
	\displaystyle \hspace{-2mm}\exp\hspace{-1mm}\left(\hspace{-1mm}-\frac{\xi_k(z^\prime_{k}x_{k}-z_{k}x^\prime_{k})^2+(x_{k}\nu_k-z_{k})^2+(x^\prime_{k}\nu_k-z^\prime_{k})^2}{2(\xi_k(x_{k}^2+x_{k}^{\prime 2})+1)}\hspace{-1mm}\right)\hspace{-1mm}\Bigg\},\vspace{1mm}\\
	\displaystyle\dfrac{\partial\psi_\mathbf{zz^\prime}}{\partial \beta}=\hspace{-0.5mm}\sum\limits_{i=1}^{B}\hspace{-0.5mm}\sum\limits_{\mathbf{x},\mathbf{x}^\prime\in\mathcal{D}_i} c^{i}_{\mathbf{x}\mathbf{x}^\prime}\hspace{-0.5mm}\prod_{k=1}^{d}\hspace{-0.5mm}\Bigg\{\hspace{-0.5mm}(\xi_k(x_{k}^2+x_{k}^{\prime2})+1)^{-\frac{1}{2}}\\
	\displaystyle\hspace{-2mm}\exp\hspace{-1mm}\left(\hspace{-1mm}-\frac{\xi_k(z^\prime_{k}x_{k}-z_{k}x^\prime_{k})^2+(x_{k}\nu_k-z_{k})^2+(x^\prime_{k}\nu_k-z^\prime_{k})^2}{2(\xi_k(x_{k}^2+x_{k}^{\prime 2})+1)}\hspace{-1mm}\right)\hspace{-1mm}\Bigg\}.
	\end{array}
	\end{equation*}
	From Appendix~\ref{Derivation of Omega, Psi and Upsilon},
	\begin{equation*}
		\begin{array}{rcl}
			\gamma_{\mathbf{x}\mathbf{x}^\prime}&\hspace{-2.4mm}=&\hspace{-2.4mm}\displaystyle\left(\beta+\alpha^2\right)\prod_{k=1}^{d}{(\xi_k(x_{k}-x^\prime_{k})^2+1)^{-\frac{1}{2}}}\\
			&&\hspace{-2.4mm}\displaystyle\exp\left(-\frac{\nu_k^2(x_{k}-x^\prime_{k})^2}{2(\xi_k(x_{k}-x^\prime_{k})^2+1)}\right).
		\end{array}
	\end{equation*}
	%
	The partial derivative of $\gamma_{\mathbf{x}\mathbf{x}^\prime}$ with respect to $\boldsymbol{\nu}$, $\mathbf{\Xi}$, $\alpha$, and $\beta$ can be derived as follows:
	\begin{equation*}
		\hspace{-1.7mm}
		\begin{array}{l}
			\displaystyle\dfrac{\partial\gamma_{\mathbf{x}\mathbf{x}^\prime}}{\partial\nu_i}\displaystyle=\left(\beta+\alpha^2\right)\\
			\displaystyle\prod_{k=1}^{d}{(\xi_k(x_{k}-x^\prime_{k})^2+1)^{-\frac{1}{2}}}\exp\hspace{-0.5mm}\left(-\frac{\nu_k^2(x_{k}-x^\prime_{k})^2}{2(\xi_k(x_{k}-x^\prime_{k})^2+1)}\right)\\
			\displaystyle\times\left(-\dfrac{\nu_i^2(x_{i}-x^\prime_{i})^2}{2\big(\xi_i(x_{i}-x^\prime_{i})^2+1\big)}\right)^\prime\vspace{1mm}\\
			\displaystyle=\left(\beta+\alpha^2\right)\\
			\displaystyle\prod_{k=1}^{d}{(\xi_k(x_{k}-x^\prime_{k})^2+1)^{-\frac{1}{2}}}\exp\hspace{-0.5mm}\left(-\frac{\nu_k^2(x_{k}-x^\prime_{k})^2}{2(\xi_k(x_{k}-x^\prime_{k})^2+1)}\right)\hspace{-1mm}\\
			\displaystyle\times\left(-\dfrac{\nu_i(x_{i}-x^\prime_{i})^2}{\xi_i(x_{i}-x^\prime_{i})^2+1}\right),
		\end{array}
	\end{equation*}
	\begin{equation*}
		\hspace{-1.7mm}
		\begin{array}{l}
			\displaystyle\dfrac{\partial\gamma_{\mathbf{x}\mathbf{x}^\prime}}{\partial\xi_i}
			\displaystyle=\left(\beta+\alpha^2\right)\\
			\displaystyle\prod_{k\neq i}{(\xi_k(x_{k}-x^\prime_{k})^2+1)^{-\frac{1}{2}}}\exp\hspace{-0.5mm}\left(-\frac{\nu_k^2(x_{k}-x^\prime_{k})^2}{2(\xi_k(x_{k}-x^\prime_{k})^2+1)}\right)\\
			\displaystyle\times\Bigg(\left({(\xi_i(x_{i}-x^\prime_{i})^2+1)^{-\frac{1}{2}}}\right)^\prime\exp\hspace{-0.5mm}\left(-\frac{\nu_i^2(x_{i}-x^\prime_{i})^2}{2(\xi_i(x_{i}-x^\prime_{i})^2+1)}\right)\\
			\displaystyle+{(\xi_i(x_{i}-x^\prime_{i})^2+1)^{-\frac{1}{2}}}\left(\exp\hspace{-0.5mm}\left(-\frac{\nu_i^2(x_{i}-x^\prime_{i})^2}{2(\xi_i(x_{i}-x^\prime_{i})^2+1)}\right)\right)^{\hspace{-0.5mm}\prime}\Bigg)\\
			\displaystyle=\left(\beta+\alpha^2\right)\\
			\displaystyle\prod_{k=1}^{d}{(\xi_k(x_{k}-x^\prime_{k})^2+1)^{-\frac{1}{2}}}\exp\hspace{-0.5mm}\left(-\frac{\nu_k^2(x_{k}-x^\prime_{k})^2}{2(\xi_k(x_{k}-x^\prime_{k})^2+1)}\right)\\
			\displaystyle\times\left(-\dfrac{(x_{i}-x^\prime_{i})^2}{2\big(\xi_i(x_{i}-x^\prime_{i})^2+1\big)}+\dfrac{\nu_i^2(x_{i}-x^\prime_{i})^4}{2\big(\xi_i(x_{i}-x^\prime_{i})^2+1\big)^2}\right) \\
			\displaystyle=\left(\beta+\alpha^2\right)\\
			\displaystyle\prod_{k=1}^{d}{(\xi_k(x_{k}-x^\prime_{k})^2+1)^{-\frac{1}{2}}}\exp\hspace{-0.5mm}\left(-\frac{\nu_k^2(x_{k}-x^\prime_{k})^2}{2(\xi_k(x_{k}-x^\prime_{k})^2+1)}\right)\\
			\displaystyle\times\dfrac{(\nu_i^2-\xi_i)(x_{i}-x^\prime_{i})^4-(x_{i}-x^\prime_{i})^2}{2\big(\xi_i(x_{i}-x^\prime_{i})^2+1\big)^2}\ ,
		\end{array}
	\end{equation*} 
		\begin{equation*}
		\begin{array}{l}
		\displaystyle\dfrac{\partial\gamma_{\mathbf{x}\mathbf{x}^\prime}}{\partial \alpha}=\\
		\displaystyle2\alpha\prod_{k=1}^{d}{(\xi_k(x_{k}-x^\prime_{k})^2+1)^{-\frac{1}{2}}}\exp\hspace{-0.5mm}\left(-\frac{\nu_k^2(x_{k}-x^\prime_{k})^2}{2(\xi_k(x_{k}-x^\prime_{k})^2+1)}\right),\vspace{1mm}\\
		\displaystyle\dfrac{\partial\gamma_{\mathbf{x}\mathbf{x}^\prime}}{\partial \beta}=\\
		\displaystyle\prod_{k=1}^{d}{(\xi_k(x_{k}-x^\prime_{k})^2+1)^{-\frac{1}{2}}}\exp\hspace{-0.5mm}\left(-\frac{\nu_k^2(x_{k}-x^\prime_{k})^2}{2(\xi_k(x_{k}-x^\prime_{k})^2+1)}\right).
		\end{array}
		\end{equation*}
\subsection{Derivation of $\mu_{\mathbf{x}^*|\mathcal{D}}$ and $\sigma^2_{\mathbf{x}^*|\mathcal{D}}$} 
	\label{q(y^*)}
	\subsubsection{VBPITC, VBFIC, VBFITC, and VBDTC}
	VBPITC, VBFIC, VBFITC, and VBDTC share the same approximated test conditional $q(f_{\mathbf{x}^*}|\mathbf{y}_{\mathcal{D}_B}, \mathbf{s}_\mathcal{I},\mathbf{\Lambda},\sigma_f)\triangleq p(f_{\mathbf{x}^*}|\mathbf{s}_\mathcal{I},\mathbf{\Lambda},\sigma_f)$ but differ in
	$q^+(\mathbf{s}_\mathcal{I})$, $q^+(\mathbf{\Lambda})$, and $q^+(\sigma_f)$ obtained from their stochastic gradient ascent updates.
	As a result,
	\begin{equation*}
	\hspace{-1.7mm}
	\begin{array}{l}
		q(f_{\mathbf{x}^*}|\mathbf{y}_\mathcal{D})\\
		\displaystyle=\int p(f_{\mathbf{x}^*}|\mathbf{s}_\mathcal{I},\mathbf{\Lambda},\sigma_f)\ q^+(\mathbf{s}_\mathcal{I})\ q^+(\mathbf{\Lambda})\ q^+(\sigma_f)\ \mathrm{d}\mathbf{s}_\mathcal{I}\ \mathrm{d}\mathbf{\Lambda}\ \mathrm{d}\sigma_f
	\end{array}
	\end{equation*}
	where
	\begin{equation*}
		\hspace{-1.7mm}
		\begin{array}{rcl}
			p(f_{\mathbf{x}^*}|\mathbf{s}_\mathcal{I},\mathbf{\Lambda},\sigma_f)&\hspace{-2.4mm}=&\hspace{-2.4mm}\displaystyle\mathcal{N}(\mathbf{K}_{\mathbf{x}^*\mathcal{I}}\mathbf{\Sigma}_\mathcal{II}^{-1}\mathbf{s}_\mathcal{I},\\
			&&\hspace{-2.4mm}\displaystyle\quad\ \ k_{\mathbf{x}^*\mathbf{x}^*} -\mathbf{\mathbf{K}}_{\mathbf{x}^*\mathcal{I}}\mathbf{\Sigma}_\mathcal{II}^{-1}\mathbf{\mathbf{K}}_{\mathcal{I}\mathbf{x}^*}\hspace{-0.5mm}) \ , \vspace{1mm}\\
			q^+(\mathbf{s}_\mathcal{I})&\hspace{-2.4mm}=&\hspace{-2.4mm}\displaystyle\mathcal{N}(\mathbf{m}^+,\mathbf{S}^+)\ , \\
			q^+(\mathbf{\Lambda})&\hspace{-2.4mm}=&\hspace{-2.4mm}\displaystyle\prod_{i=1}^{d}\mathcal{N}(\lambda_i|\nu_i^+,\xi_i^+)\ , \vspace{1mm}\\
			q^+(\sigma_f)&\hspace{-2.4mm}=&\hspace{-2.4mm}\mathcal{N}(\alpha^+,\beta^+)\ .	
		\end{array}
	\end{equation*}
	Then,
	\begin{equation*}
		\begin{array}{l}
			\displaystyle q(f_{\mathbf{x}^*}|\mathbf{y}_\mathcal{D},\mathbf{\Lambda},\sigma_f)\\
			\displaystyle=\int p(f_{\mathbf{x}^*}|\mathbf{s}_\mathcal{I},\mathbf{\Lambda},\sigma_f)\ q^+(\mathbf{s}_\mathcal{I})\ \mathrm{d}\mathbf{s}_\mathcal{I} \\
			\displaystyle =\mathcal{N}(\mathbf{K}_{\mathbf{x}^*\mathcal{I}}\mathbf{\Sigma}_\mathcal{II}^{-1}\mathbf{m}^+,k_{\mathbf{x}^*\mathbf{x}^*}-\mathbf{K}_{\mathbf{x}^*\mathcal{I}}\mathbf{\Sigma}_\mathcal{II}^{-1}\mathbf{K}_{\mathcal{I}\mathbf{x}^*}\\
			\displaystyle\qquad\qquad\qquad\qquad\ \ \  +\mathbf{K}_{\mathbf{x}^*\mathcal{I}}\mathbf{\Sigma}_\mathcal{II}^{-1}\mathbf{S}^+\mathbf{\Sigma}_\mathcal{II}^{-1}\mathbf{K}_{\mathcal{I}\mathbf{x}^*})\ .
		\end{array}
	\end{equation*}
	Finally, 	
	\begin{equation*}
		\begin{array}{l}
			\displaystyle\mu_{\mathbf{x}^*|\mathcal{D}}\\
			\displaystyle\triangleq\mathbb{E}_{q(f_{\mathbf{x}^*}|\mathbf{y}_\mathcal{D})}[f_{\mathbf{x}^*}]\vspace{1mm}\\
			\displaystyle=\mathbb{E}_{q^+(\mathbf{\Lambda},\sigma_f)}[\mathbb{E}_{q(f_{\mathbf{x}^*}|\mathbf{y}_\mathcal{D},\mathbf{\Lambda}, \sigma_f)}[f_{\mathbf{x}^*}]] \vspace{1mm}\\
			\displaystyle=\mathbb{E}_{q^+(\mathbf{\Lambda},\sigma_f)}[ \mathbf{K}_{\mathbf{x}^*\mathcal{I}}\mathbf{\Sigma}_\mathcal{II}^{-1}\mathbf{m}^+ ] \vspace{1mm}\\
			\displaystyle=\mathbb{E}_{q^+(\mathbf{\Lambda},\sigma_f)}[\mathbf{\mathbf{K}}_{\mathbf{x}^*\mathcal{I}}]\mathbf{\Sigma}_\mathcal{II}^{-1}\mathbf{m}^+\ .
		\end{array}
	\end{equation*}		
	\begin{equation*}
		\hspace{-1.7mm}
		\begin{array}{l}
			\displaystyle\sigma^2_{\mathbf{x}^*|\mathcal{D}}\\
			\displaystyle\triangleq\mathbb{V}_{q(f_{\mathbf{x}^*}|\mathbf{y}_\mathcal{D})}[f_{\mathbf{x}^*}]\\
			\displaystyle=\mathbb{E}_{q^+(\mathbf{\Lambda},\sigma_f)}[\mathbb{V}_{q(f_{\mathbf{x}^*}|\mathbf{y}_\mathcal{D},\mathbf{\Lambda},\sigma_f)}[f_{\mathbf{x}^*}]]\\
			\displaystyle +\  \mathbb{V}_{q^+(\mathbf{\Lambda},\sigma_f)}[\mathbb{E}_{q(f_{\mathbf{x}^*}|\mathbf{y}_\mathcal{D},\mathbf{\Lambda},\sigma_f)}[f_{\mathbf{x}^*}]]\vspace{1mm}\\
			\displaystyle\hspace{-0.5mm}=\hspace{-0.5mm}\mathbb{E}_{q^+(\mathbf{\Lambda},\sigma_f)}[k_{\mathbf{x}^*\mathbf{x}^*}\hspace{-1mm}-\hspace{-0.5mm}\mathbf{\mathbf{K}}_{\mathbf{x}^*\mathcal{I}}\mathbf{\Sigma}_\mathcal{II}^{-1}\mathbf{K}_{\mathcal{I}\mathbf{x}^*}\hspace{-1mm}+\hspace{-0.5mm}\mathbf{\mathbf{K}}_{\mathbf{x}^*\mathcal{I}}\mathbf{\Sigma}_\mathcal{II}^{-1}\mathbf{S}^+\mathbf{\Sigma}_\mathcal{II}^{-1}\mathbf{\mathbf{K}}_{\mathcal{I}\mathbf{x}^*}]\vspace{1mm}\\
			\displaystyle\quad+\mathbb{V}_{q^+(\mathbf{\Lambda},\sigma_f)}[\mathbf{m}^{+\top} \mathbf{\Sigma}_\mathcal{II}^{-1}\mathbf{K}_{\mathcal{I}\mathbf{x}^*} ]
		\end{array}
	\end{equation*}
	where
	\begin{equation*}
		\begin{array}{l}
			\displaystyle\mathbb{V}_{q^+(\mathbf{\Lambda},\sigma_f)}[\mathbf{m}^{+\top} \mathbf{\Sigma}_\mathcal{II}^{-1}\mathbf{K}_{\mathcal{I}\mathbf{x}^*} ]\vspace{1mm}\\
			\displaystyle= \mathbf{m}^{+\top} \mathbf{\Sigma}_\mathcal{II}^{-1}\mathbb{V}_{q^+(\mathbf{\Lambda},\sigma_f)}[\mathbf{K}_{\mathcal{I}\mathbf{x}^*} ]\mathbf{\Sigma}_\mathcal{II}^{-1}\mathbf{m}^{+}\vspace{1mm}\\
			\displaystyle= \mathbf{m}^{+\top} \mathbf{\Sigma}_\mathcal{II}^{-1}\Big(\mathbb{E}_{q^+(\mathbf{\Lambda},\sigma_f)}[\mathbf{K}_{\mathcal{I}\mathbf{x}^*}\mathbf{K}_{\mathbf{x}^*\mathcal{I}} ]\vspace{1mm}\\
			\displaystyle\qquad\qquad -  \mathbb{E}_{q^+(\mathbf{\Lambda},\sigma_f)}[\mathbf{K}_{\mathcal{I}\mathbf{x}^*} ]\mathbb{E}_{q^+(\mathbf{\Lambda},\sigma_f)}[\mathbf{K}_{\mathbf{x}^*\mathcal{I}} ]\Big)\mathbf{\Sigma}_\mathcal{II}^{-1}\mathbf{m}^{+}.
		\end{array}
	\end{equation*}
	Note that the closed-form expressions of all the above expectation terms  with respect to 
$q^+(\mathbf{\Lambda},\sigma_f)\triangleq q^+(\mathbf{\Lambda})q^+(\sigma_f)$ can be derived in a similar manner as that of $\mathbf{\Psi}_\mathcal{II}\triangleq\mathbb{E}_{q(\mathbf{\Lambda},\sigma_f)}[\mathbf{K}_{\mathcal{ID}}\mathbf{C}^{-1}_\mathcal{DD}\mathbf{K}_{\mathcal{DI}}]$, $\mathbf{\Omega}_\mathcal{ID}\triangleq\mathbb{E}_{q(\mathbf{\Lambda},\sigma_f)}[\mathbf{K}_{\mathcal{ID}}]$, and $\mathbf{\Upsilon}_\mathcal{DD}\triangleq\mathbb{E}_{q(\mathbf{\Lambda},\sigma_f)}[\mathbf{K}_{\mathcal{DD}}]$. Hence, $\mu_{\mathbf{x}^*|\mathcal{D}}$ and $\sigma^2_{\mathbf{x}^*|\mathcal{D}}$ can be derived in closed form.
	\subsubsection{VBPIC}
	VBPIC uses the exact test conditional $q(f_{\mathbf{x}^*}|\mathbf{y}_{\mathcal{D}_B}, \mathbf{s}_\mathcal{I},\mathbf{\Lambda},\sigma_f)\triangleq p(f_{\mathbf{x}^*}|\mathbf{y}_{\mathcal{D}_B}, \mathbf{s}_\mathcal{I},\mathbf{\Lambda},\sigma_f)$.
	To derive $p(f_{\mathbf{x}^*}|\mathbf{y}_{\mathcal{D}_B}, \mathbf{s}_\mathcal{I},\mathbf{\Lambda}, \sigma_f)$, we use the fundamental definition of GP to give the following expression for the Gaussian joint distribution $p(f_{\mathbf{x}^*},\mathbf{s}_\mathcal{I},\mathbf{y}_{\mathcal{D}_B}|\mathbf{\Lambda},\sigma_f)$:
	\begin{equation*}
		\mathcal{N}\Biggm(\mathbf{0},
		\begin{pmatrix}
			k_{\mathbf{x}^*\mathbf{x}^*} & \mathbf{K}_{\mathbf{x}^*\mathcal{I}} & \mathbf{K}_{\mathbf{x}^*\mathcal{D}_B}\\
			\mathbf{K}_{\mathcal{I}\mathbf{x}^*} & \mathbf{\Sigma}_{\mathcal{I}\mathcal{I}} & \mathbf{K}_{\mathcal{I}\mathcal{D}_B} \\
			\mathbf{K}_{\mathcal{D}_B\mathbf{x}^*} & \mathbf{K}_{\mathcal{D}_B\mathcal{I}} & \mathbf{K}_{\mathcal{D}_B\mathcal{D}_B}\hspace{-1mm} + \hspace{-0.5mm} \mathbf{C}_{\mathcal{D}_B\mathcal{D}_B}
		\end{pmatrix}\Bigg).
	\end{equation*} 
	Then, $p(f_{\mathbf{x}^*}|\mathbf{s}_\mathcal{I},\mathbf{y}_{\mathcal{D}_B},\mathbf{\Lambda},\sigma_f)
	=\mathcal{N}(\mathbb{E}_{p(f_{\mathbf{x}^*}|\mathbf{s}_\mathcal{I},\mathbf{y}_{\mathcal{D}_B},\mathbf{\Lambda},\sigma_f)}[f_{\mathbf{x}^*}],\mathbb{V}_{p(f_{\mathbf{x}^*}|\mathbf{s}_\mathcal{I},\mathbf{y}_{\mathcal{D}_B},\mathbf{\Lambda},\sigma_f)}[f_{\mathbf{x}^*}])$ 
	where
	\begin{equation*}
		\hspace{-1.7mm}
		\begin{array}{l}
			\displaystyle\mathbb{E}_{p(f_{\mathbf{x}^*}|\mathbf{s}_\mathcal{I},\mathbf{y}_{\mathcal{D}_B},\mathbf{\Lambda},\sigma_f)}[f_{\mathbf{x}^*}] \vspace{1mm}\\
			\hspace{-.5mm}=\hspace{-.5mm}\begin{pmatrix}
				\mathbf{K}_{\mathbf{x}^*\mathcal{I}} & \hspace{-2mm}\mathbf{K}_{\mathbf{x}^*\mathcal{D}_B}
			\end{pmatrix}
			\hspace{-1mm}
			\begin{pmatrix}
				\mathbf{\Sigma}_\mathcal{II} & \mathbf{K}_{\mathcal{I}\mathcal{D}_B} \\
				\mathbf{K}_{\mathcal{D}_B\mathcal{I}} & \mathbf{K}_{\mathcal{D}_B\mathcal{D}_B}\hspace{-3mm} + \hspace{-0.5mm} \mathbf{C}_{\mathcal{D}_B\mathcal{D}_B}
			\end{pmatrix}^{-1}
			\hspace{-1mm}
			\begin{pmatrix}
				\mathbf{s}_\mathcal{I} \\
				\mathbf{y}_{\mathcal{D}_B}
			\end{pmatrix}  ,\vspace{2mm}\\
			\displaystyle\mathbb{V}_{p(f_{\mathbf{x}^*}|\mathbf{s}_\mathcal{I},\mathbf{y}_{\mathcal{D}_B},\mathbf{\Lambda},\sigma_f)}[f_{\mathbf{x}^*}] =k_{\mathbf{x}^*\mathbf{x}^*}-\vspace{1mm}\\
			\hspace{-1mm}
			\begin{pmatrix}
				\mathbf{K}_{\mathbf{x}^*\mathcal{I}} & \hspace{-2mm}\mathbf{K}_{\mathbf{x}^*\mathcal{D}_B}
			\end{pmatrix}
			\hspace{-1mm}
			\begin{pmatrix}
				\mathbf{\Sigma}_\mathcal{II} & \mathbf{K}_{\mathcal{I}\mathcal{D}_B} \\
				\mathbf{K}_{\mathcal{D}_B\mathcal{I}} & \mathbf{K}_{\mathcal{D}_B\mathcal{D}_B}\hspace{-3mm} + \hspace{-0.5mm} \mathbf{C}_{\mathcal{D}_B\mathcal{D}_B}
			\end{pmatrix}^{-1}
			\hspace{-1mm}
			\begin{pmatrix}
				\mathbf{K}_{\mathcal{I}\mathbf{x}^*}\\
				\mathbf{K}_{\mathcal{D}_B\mathbf{x}^*}
			\end{pmatrix}.
		\end{array}
	\end{equation*}
	To simplify the above expressions, let
	\begin{equation*}
		\mathbf{J}\triangleq
		\begin{pmatrix}
			\mathbf{\Sigma}_{\mathcal{I}\mathcal{I}} & \mathbf{K}_{\mathcal{I}\mathcal{D}_B} \\
			\mathbf{K}_{\mathcal{D}_B\mathcal{I}} & \mathbf{K}_{\mathcal{D}_B\mathcal{D}_B}\hspace{-3mm} + \hspace{-0.5mm} \mathbf{C}_{\mathcal{D}_B\mathcal{D}_B}
		\end{pmatrix}^{-1}=
		\begin{pmatrix}
			\mathbf{J}_{\mathcal{I}\mathcal{I}} & \hspace{-1mm}\mathbf{J}_{\mathcal{I}\mathcal{D}_B} \\
			\mathbf{J}_{\mathcal{D}_B\mathcal{I}} & \mathbf{J}_{\mathcal{D}_B\mathcal{D}_B}
		\end{pmatrix}
	\end{equation*}
	where $\mathbf{J}_{\mathcal{I}\mathcal{I}}$, $\mathbf{J}_{\mathcal{I}\mathcal{D}_B}$,
	$\mathbf{J}_{\mathcal{D}_B\mathcal{I}}$, and $\mathbf{J}_{\mathcal{D}_B\mathcal{D}_B}$ can be derived by applying the matrix inversion lemma for partitioned matrices directly.
	Then,
	\begin{equation*}
		\hspace{-1.7mm}
		\begin{array}{l}
			\displaystyle\mathbb{E}_{p(f_{\mathbf{x}^*}|\mathbf{s}_\mathcal{I},\mathbf{y}_{\mathcal{D}_B},\mathbf{\Lambda},\sigma_f)}[f_{\mathbf{x}^*}] \vspace{1mm}\\
			=\begin{pmatrix}
				\mathbf{K}_{\mathbf{x}^*\mathcal{I}} & \mathbf{K}_{\mathbf{x}^*\mathcal{D}_B}
			\end{pmatrix}
			\begin{pmatrix}
				\mathbf{J}_{\mathcal{I}\mathcal{I}} & \mathbf{J}_{\mathcal{I}\mathcal{D}_B} \\
				\mathbf{J}_{\mathcal{D}_B\mathcal{I}} & \mathbf{J}_{\mathcal{D}_B\mathcal{D}_B}
			\end{pmatrix}
			\begin{pmatrix}
				\mathbf{s}_\mathcal{I} \\
				\mathbf{y}_{\mathcal{D}_B}
			\end{pmatrix} \vspace{1mm}\\
			\displaystyle=\left(\mathbf{K}_{\mathbf{x}^*\mathcal{I}}\mathbf{J}_{\mathcal{I}\mathcal{I}}+\mathbf{K}_{\mathbf{x}^*\mathcal{D}_B}\mathbf{J}_{\mathcal{D}_B\mathcal{I}}\right)\mathbf{s}_\mathcal{I}\vspace{1mm}\\
			\displaystyle\quad+\left(\mathbf{K}_{\mathbf{x}^*\mathcal{I}}\mathbf{J}_{\mathcal{I}\mathcal{D}_B}+\mathbf{K}_{\mathbf{x}^*\mathcal{D}_B}\mathbf{J}_{\mathcal{D}_B\mathcal{D}_B}\right)\mathbf{y}_{\mathcal{D}_B}\ ,\vspace{2mm}\\
			\displaystyle\mathbb{V}_{p(f_{\mathbf{x}^*}|\mathbf{s}_\mathcal{I},\mathbf{y}_{\mathcal{D}_B},\mathbf{\Lambda},\sigma_f)}[f_{\mathbf{x}^*}] \vspace{1mm}\\
			=k_{\mathbf{x}^*\mathbf{x}^*}-
			\mathbf{K}_{\mathbf{x}^*(\mathcal{I}\cup\mathcal{D}_B)} 
			\mathbf{J}
			\mathbf{K}_{(\mathcal{I}\cup\mathcal{D}_B)\mathbf{x}^*}\ .
		\end{array}
	\end{equation*}
	Now,
	\begin{equation*}
	\hspace{-1.7mm}
	\begin{array}{l}
		q(f_{\mathbf{x}^*}|\mathbf{y}_\mathcal{D})\\
		\displaystyle=\hspace{-1mm}\int\hspace{-0.5mm} p(f_{\mathbf{x}^*}|\mathbf{y}_{\mathcal{D}_B}, \mathbf{s}_\mathcal{I},\mathbf{\Lambda},\sigma_f)\ q^+(\mathbf{s}_\mathcal{I})\ q^+(\mathbf{\Lambda})\ q^+(\sigma_f)\mathrm{d}\mathbf{s}_\mathcal{I}\mathrm{d}\mathbf{\Lambda}\mathrm{d}\sigma_f
	\end{array}
	\end{equation*}
	where
	\begin{equation*}
		\begin{array}{rcl}
			p(f_{\mathbf{x}^*}|\mathbf{y}_{\mathcal{D}_B}, \mathbf{s}_\mathcal{I},\mathbf{\Lambda},\sigma_f)&\hspace{-2.4mm}=&\hspace{-2.4mm}\displaystyle\mathcal{N}(f_{\mathbf{x}^*}|\mathbb{E}_{p(f_{\mathbf{x}^*}|\mathbf{s}_\mathcal{I},\mathbf{y}_{\mathcal{D}_B},\mathbf{\Lambda},\sigma_f)}[f_{\mathbf{x}^*}],\\
			&&\hspace{-2.4mm}\qquad\quad\mathbb{V}_{p(f_{\mathbf{x}^*}|\mathbf{s}_\mathcal{I},\mathbf{y}_{\mathcal{D}_B},\mathbf{\Lambda},\sigma_f)}[f_{\mathbf{x}^*}]) \ ,\vspace{1mm}\\
			q^+(\mathbf{s}_\mathcal{I})&\hspace{-2.4mm}=&\hspace{-2.4mm}\displaystyle\mathcal{N}(\mathbf{m}^+,\mathbf{S}^+)\ , \\
			q^+(\mathbf{\Lambda})&\hspace{-2.4mm}=&\hspace{-2.4mm}\displaystyle\prod_{i=1}^{d}\mathcal{N}(\lambda_i|\nu_i^+,\xi_i^+)\ , \vspace{1mm}\\
			q^+(\sigma_f)&\hspace{-2.4mm}=&\hspace{-2.4mm}\mathcal{N}(\alpha^+,\beta^+)\ .	
		\end{array}
	\end{equation*}
	Then,
	\begin{equation*}
		\begin{array}{l}
			\displaystyle q(f_{\mathbf{x}^*}|\mathbf{y}_\mathcal{D},\mathbf{\Lambda},\sigma_f)\\
			\displaystyle=\int p(f_{\mathbf{x}^*}|\mathbf{y}_{\mathcal{D}_B},\mathbf{s}_\mathcal{I},\mathbf{\Lambda},\sigma_f)\ q^+(\mathbf{s}_\mathcal{I})\ \mathrm{d}\mathbf{s}_\mathcal{I} \\
			\displaystyle =\mathcal{N}(\left(\mathbf{K}_{\mathbf{x}^*\mathcal{I}}\mathbf{J}_{\mathcal{I}\mathcal{I}}+\mathbf{K}_{\mathbf{x}^*\mathcal{D}_B}\mathbf{J}_{\mathcal{D}_B\mathcal{I}}\right)\mathbf{m}^+\\
			\displaystyle\qquad\ \ +\left(\mathbf{K}_{\mathbf{x}^*\mathcal{I}}\mathbf{J}_{\mathcal{I}\mathcal{D}_B}+\mathbf{K}_{\mathbf{x}^*\mathcal{D}_B}\mathbf{J}_{\mathcal{D}_B\mathcal{D}_B}\right)\mathbf{y}_{\mathcal{D}_B}\ ,\vspace{1mm}\\
			\displaystyle\qquad\ \ k_{\mathbf{x}^*\mathbf{x}^*}-
			\mathbf{K}_{\mathbf{x}^*(\mathcal{I}\cup\mathcal{D}_B)} 
			\mathbf{J}
			\mathbf{K}_{(\mathcal{I}\cup\mathcal{D}_B)\mathbf{x}^*}\\
			\displaystyle\qquad\ \ + \left(\mathbf{K}_{\mathbf{x}^*\mathcal{I}}\mathbf{J}_{\mathcal{I}\mathcal{I}}+\mathbf{K}_{\mathbf{x}^*\mathcal{D}_B}\mathbf{J}_{\mathcal{D}_B\mathcal{I}}\right)\\
			\displaystyle\qquad\qquad \mathbf{S}^+\left(\mathbf{K}_{\mathbf{x}^*\mathcal{I}}\mathbf{J}_{\mathcal{I}\mathcal{I}}+\mathbf{K}_{\mathbf{x}^*\mathcal{D}_B}\mathbf{J}_{\mathcal{D}_B\mathcal{I}}\right)^{\top} )\ .
		\end{array}
	\end{equation*}
	Finally,
	\begin{equation*}
		\begin{array}{l}
			\displaystyle\mu_{\mathbf{x}^*|\mathcal{D}}\\
			\displaystyle\triangleq\mathbb{E}_{q(f_{\mathbf{x}^*}|\mathbf{y}_\mathcal{D})}[f_{\mathbf{x}^*}]\vspace{1mm}\\
			\displaystyle=\mathbb{E}_{q^+(\mathbf{\Lambda},\sigma_f)}[\mathbb{E}_{q(f_{\mathbf{x}^*}|\mathbf{y}_\mathcal{D},\mathbf{\Lambda},\sigma_f)}[f_{\mathbf{x}^*}]] \vspace{1mm}\\
			\displaystyle=\mathbb{E}_{q^+(\mathbf{\Lambda},\sigma_f)}[ \left(\mathbf{K}_{\mathbf{x}^*\mathcal{I}}\mathbf{J}_{\mathcal{I}\mathcal{I}}+\mathbf{K}_{\mathbf{x}^*\mathcal{D}_B}\mathbf{J}_{\mathcal{D}_B\mathcal{I}}\right)\mathbf{m}^+\\
			\displaystyle\qquad\qquad\quad\ \ +\left(\mathbf{K}_{\mathbf{x}^*\mathcal{I}}\mathbf{J}_{\mathcal{I}\mathcal{D}_B}+\mathbf{K}_{\mathbf{x}^*\mathcal{D}_B}\mathbf{J}_{\mathcal{D}_B\mathcal{D}_B}\right)\mathbf{y}_{\mathcal{D}_B} ] \vspace{1mm}\\
			\displaystyle=\mathbb{E}_{q^+(\mathbf{\Lambda},\sigma_f)}\left[\mathbf{K}_{\mathbf{x}^*\mathcal{I}}\mathbf{J}_{\mathcal{I}\mathcal{I}}+\mathbf{K}_{\mathbf{x}^*\mathcal{D}_B}\mathbf{J}_{\mathcal{D}_B\mathcal{I}}\right]\mathbf{m}^+\\
			\displaystyle\quad +\ \mathbb{E}_{q^+(\mathbf{\Lambda},\sigma_f)}\left[\mathbf{K}_{\mathbf{x}^*\mathcal{I}}\mathbf{J}_{\mathcal{I}\mathcal{D}_B}+\mathbf{K}_{\mathbf{x}^*\mathcal{D}_B}\mathbf{J}_{\mathcal{D}_B\mathcal{D}_B}\right]\mathbf{y}_{\mathcal{D}_B}\ .
		\end{array}
	\end{equation*}		
	\begin{equation*}
		\hspace{-5.7mm}
		\begin{array}{l}
			\displaystyle\sigma^2_{\mathbf{x}^*|\mathcal{D}}
			\displaystyle\triangleq\mathbb{V}_{q(f_{\mathbf{x}^*}|\mathbf{y}_\mathcal{D})}[f_{\mathbf{x}^*}] \vspace{1mm}\\
			\displaystyle=\mathbb{E}_{q^+(\mathbf{\Lambda},\sigma_f)}[\mathbb{V}_{q(f_{\mathbf{x}^*}|\mathbf{y}_\mathcal{D},\mathbf{\Lambda},\sigma_f)}[f_{\mathbf{x}^*}]] \\
			\ +\hspace{0.5mm} \mathbb{V}_{q^+(\mathbf{\Lambda},\sigma_f)}[\mathbb{E}_{q(f_{\mathbf{x}^*}|\mathbf{y}_\mathcal{D},\mathbf{\Lambda},\sigma_f)}[f_{\mathbf{x}^*}]]\vspace{1mm}\\
			\displaystyle=\mathbb{E}_{q^+(\mathbf{\Lambda},\sigma_f)}[k_{\mathbf{x}^*\mathbf{x}^*}-
			\mathbf{K}_{\mathbf{x}^*(\mathcal{I}\cup\mathcal{D}_B)} 
			\mathbf{J}
			\mathbf{K}_{(\mathcal{I}\cup\mathcal{D}_B)\mathbf{x}^*}\\
			\displaystyle\qquad\qquad\quad\ \  + \left(\mathbf{K}_{\mathbf{x}^*\mathcal{I}}\mathbf{J}_{\mathcal{I}\mathcal{I}}+\mathbf{K}_{\mathbf{x}^*\mathcal{D}_B}\mathbf{J}_{\mathcal{D}_B\mathcal{I}}\right)\\
			\displaystyle\qquad\qquad\quad\ \ \mathbf{S}^+\left(\mathbf{K}_{\mathbf{x}^*\mathcal{I}}\mathbf{J}_{\mathcal{I}\mathcal{I}}+\mathbf{K}_{\mathbf{x}^*\mathcal{D}_B}\mathbf{J}_{\mathcal{D}_B\mathcal{I}}\right)^{\top}]\vspace{1mm}\\
			\displaystyle\quad+\mathbb{V}_{q^+(\mathbf{\Lambda},\sigma_f)}\left[
			(\mathbf{m}^{+\top}\ \mathbf{y}^{\top}_{\mathcal{D}_B})
			\mathbf{J}
			\mathbf{K}_{(\mathcal{I}\cup\mathcal{D}_B)\mathbf{x}^*} \right]
		\end{array}
	\end{equation*}
	where
	\begin{equation*}
		\begin{array}{l}
			\mathbb{V}_{q^+(\mathbf{\Lambda},\sigma_f)}\left[
			(\mathbf{m}^{+\top}\ \mathbf{y}^{\top}_{\mathcal{D}_B})
			\mathbf{J}
			\mathbf{K}_{(\mathcal{I}\cup\mathcal{D}_B)\mathbf{x}^*} \right]\\
			=(\mathbf{m}^{+\top}\ \mathbf{y}^{\top}_{\mathcal{D}_B})
			\mathbb{V}_{q^+(\mathbf{\Lambda},\sigma_f)}\left[
			\mathbf{J}
			\mathbf{K}_{(\mathcal{I}\cup\mathcal{D}_B)\mathbf{x}^*} \right]
			\begin{pmatrix}
				\mathbf{m}^{+} \\
				\mathbf{y}_{\mathcal{D}_B}
			\end{pmatrix}
			\vspace{1mm}\\
			=(\mathbf{m}^{+\top}\ \mathbf{y}^{\top}_{\mathcal{D}_B})
			\Big(\mathbb{E}_{q^+(\mathbf{\Lambda},\sigma_f)}\left[\mathbf{J}
			\mathbf{K}_{(\mathcal{I}\cup\mathcal{D}_B)\mathbf{x}^*}
			\mathbf{K}_{\mathbf{x}^*(\mathcal{I}\cup\mathcal{D}_B)}\mathbf{J} \right]\vspace{1mm}\\
			\displaystyle  -  \mathbb{E}_{q^+(\mathbf{\Lambda},\sigma_f)}[\mathbf{J}
			\mathbf{K}_{(\mathcal{I}\cup\mathcal{D}_B)\mathbf{x}^*} ]\mathbb{E}_{q^+(\mathbf{\Lambda},\sigma_f)}[\mathbf{K}_{\mathbf{x}^*(\mathcal{I}\cup\mathcal{D}_B)}\mathbf{J} ]\Big)\begin{pmatrix}
				\mathbf{m}^{+} \\
				\mathbf{y}_{\mathcal{D}_B}
			\end{pmatrix}\ .
		\end{array}
	\end{equation*}
	Unfortunately, the closed-form expressions of all the above expectation terms  with respect to $q^+(\mathbf{\Lambda},\sigma_f)\triangleq q^+(\mathbf{\Lambda})q^+(\sigma_f)$ cannot be obtained because it involves integrating, over $\mathbf{\Lambda}$, terms containing $\mathbf{J}$ that depends on $\mathbf{\Lambda}$ but without an analytical form with respect to $\mathbf{\Lambda}$.
	So, we approximate them via sampling.

\end{document}